\PassOptionsToPackage{table}{xcolor}
\documentclass[11pt,a4paper]{article}
\usepackage{technical_report}

\usepackage{amsmath,amsfonts,bm}

\def\eqref#1{equation~\ref{#1}}

\def\1{\bm{1}}

\DeclareMathAlphabet{\mathsfit}{\encodingdefault}{\sfdefault}{m}{sl}
\SetMathAlphabet{\mathsfit}{bold}{\encodingdefault}{\sfdefault}{bx}{n}

\usepackage{hyperref}
\hypersetup{colorlinks=true,allcolors=reportaccent,bookmarksnumbered=true}
\usepackage{url}
\usepackage{graphicx}
\usepackage{wrapfig}
\usepackage{array}
\usepackage{tabularx}
\usepackage{multirow}
\usepackage{booktabs}
\usepackage{capt-of}
\usepackage{placeins}
\usepackage{flafter}
\usepackage{algorithm}
\usepackage{algpseudocode}

\newcolumntype{Y}{>{\centering\arraybackslash}X}
\definecolor{highlight}{RGB}{255,215,0}

\title{MoGe-4D: Efficient, Flexible, and Unified\\Dense Motion and Geometry}

\author{%
	Ruicheng Wang\textsuperscript{1,2,*}, Linlin Xin\textsuperscript{3,2,*}, Sicheng Xu\textsuperscript{2}, Jianfeng Xiang\textsuperscript{4,2,*}, Zelong Lv\textsuperscript{1,2,*},\\
	Yu Deng\textsuperscript{2}, Yue Dong\textsuperscript{2}, Guangzhong Sun\textsuperscript{1}, Tong Zhang\textsuperscript{3}, Jiaolong Yang\textsuperscript{2}\\[5pt]
	{\small\color{reportmuted}\textsuperscript{1}University of Science and Technology of China \quad \textsuperscript{2}Microsoft Research}\\
	{\small\color{reportmuted}\textsuperscript{3}University of Chinese Academy of Sciences \quad \textsuperscript{4}Tsinghua University}%
}

\date{}

\begin{document}

\maketitle
\thispagestyle{empty}
\begingroup
\renewcommand\thefootnote{}%
\footnote{\textsuperscript{*}Work done during internship at MSRA.}%
\footnote{This work is the result of an open-source project.}%
\addtocounter{footnote}{-2}%
\endgroup

\noindent\begin{minipage}{\linewidth}
    \centering
    \captionsetup{hypcap=false}
    \includegraphics[width=\linewidth]{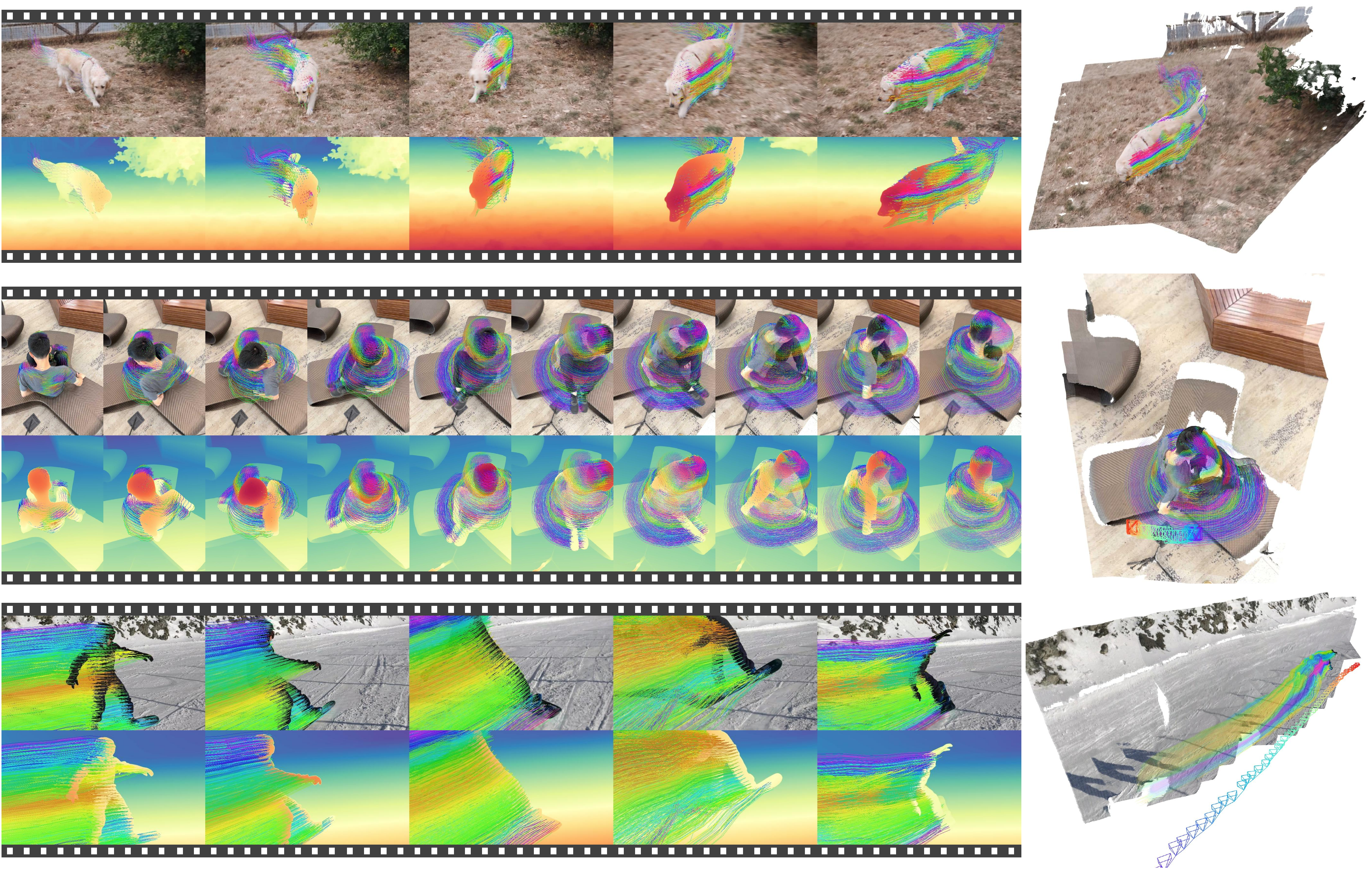}
    \captionof{figure}{\textbf{Dense 4D reconstruction and tracking with MoGe-4D.} Given an unposed RGB video, MoGe-4D predicts metric surface geometry, camera poses, and long-range dense 3D trajectories across dynamic scenes.}
    \label{fig:teaser}
\end{minipage}
\vspace{-0.2em}

\begin{abstract}
Reconstructing dynamic scenes from RGB video requires reasoning about both 3D geometry and motion across viewpoints, yet existing methods either split reconstruction and tracking into fragmented pipelines or face prohibitive dense spatiotemporal decoding costs.
We introduce \textbf{MoGe-4D}, a feed-forward framework that unifies geometry and motion prediction via ordered queries of paired allocentric and egocentric dynamic point maps.
Each query maps source pixels to a target time simultaneously in both camera frames, resolving relative camera pose in closed form while flexibly serving depth, dynamic registration, and dense 3D tracking.
For efficient dense querying, MoGe-4D decouples scene modeling from query decoding: a high-capacity context encoder caches coarse global spatiotemporal features once, while a lightweight query encoder rapidly decodes dense point maps at flexible resolutions.
Furthermore, as supervising dense 4D queries demands continuous 3D correspondences whose raw trajectory storage is prohibitive, we introduce \textbf{RigIDTrack}, a synthetic data pipeline and dataset that compresses surface dynamics into rigidity-regularized affine bodies, recovering exact arbitrary-frame 3D trajectories with $\sim$15\% storage overhead over RGB-D video.
Across diverse benchmarks, MoGe-4D achieves state-of-the-art dense 3D tracking, matches dedicated 3D foundation models in reconstruction, and runs substantially faster than prior 4D baselines. \textbf{Project Page:} \href{https://microsoft.github.io/MoGe-4D}{https://microsoft.github.io/MoGe-4D}
\end{abstract}

\clearpage
\begin{figure}[!ht]
	\centering
	\captionsetup{hypcap=false}
	\includegraphics[width=\linewidth]{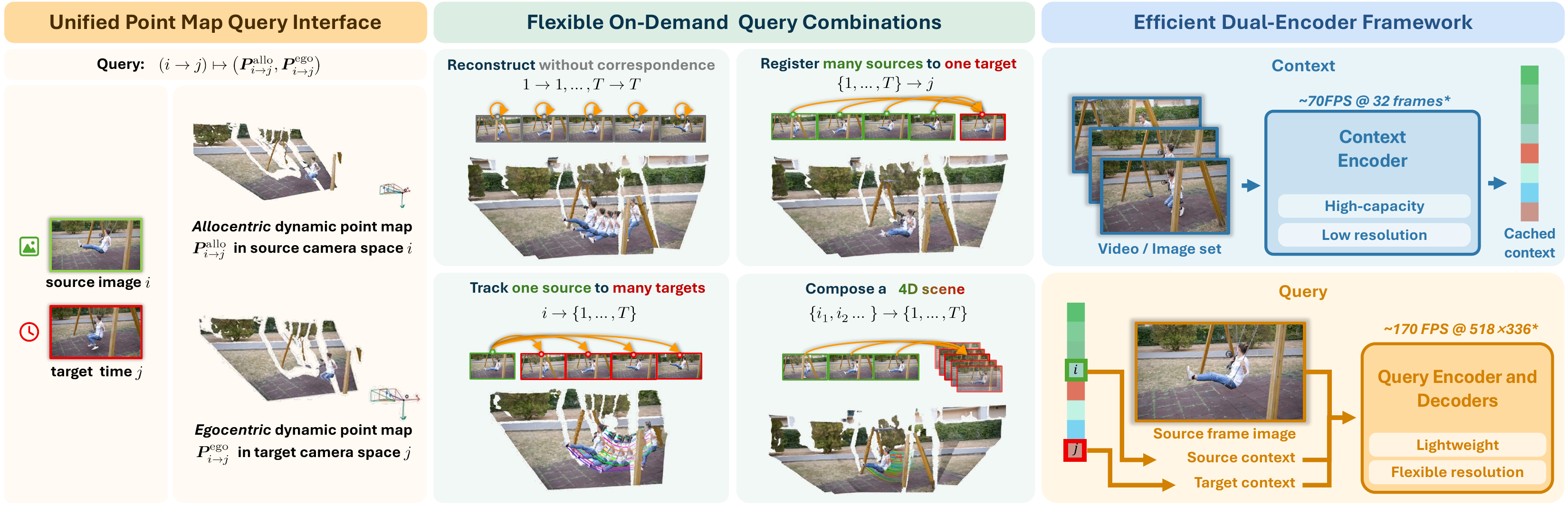}
	\captionof{figure}{\textbf{Overview of the MoGe-4D framework.}
	\emph{Left (unification):} querying paired allocentric and egocentric point maps unifies 4D geometry, motion, and camera pose.
	\emph{Middle (flexibility):} on-demand query compositions adapt to diverse reconstruction and tracking tasks.
	\emph{Right (efficiency):} a dual-encoder architecture decouples global spatiotemporal reasoning from lightweight query decoding for high-throughput inference.
	$^{*}$Measured on an NVIDIA A100 GPU with 32 frames at $518\times336$ resolution (Fig.~\ref{fig:runtime_scaling}).}
	\label{fig:framework_overview}
	\vspace{-1.5em}
\end{figure}

\section{Introduction}
\label{sec:introduction}

Reconstructing a dynamic world from RGB video requires reasoning about 3D scene structure and motion across changing viewpoints.
While feed-forward 3D foundation models~\citep{dust3r,vggt,pi3,depthanything3} excel at static multiview reconstruction, extending them to dynamic 4D scenes has largely relied on appending motion-specific branches or separate tracking modules~\citep{st4rtrack,vdpm,4rc,track4world}, which fragments geometry and motion into decoupled tasks.
Alternatively, D4RT~\citep{d4rt} unifies both tasks by formulating 4D prediction as spatiotemporal queries over an encoded scene representation.
Yet because its decoder operates point by point, scaling to dense prediction across space and time incurs prohibitive computational costs.
This tension exposes a central challenge for 4D foundation models: retaining the generality of query-based prediction while enabling efficient, dense evaluation over video sequences.

Toward a practical and scalable 4D foundation model, we pursue three complementary design goals.
First, \emph{unification}: because 3D geometry and motion both reduce to locating scene points across space and time, a single prediction interface can learn and service both coherently without fragmented task-specific branches.
Second, \emph{flexibility}: reconstruction aggregates observations across views, whereas tracking follows surface points over time; the framework should therefore support arbitrary source--target query combinations at task-adaptive output resolutions.
Third, \emph{efficiency}: as video analysis requires hundreds of dense spatiotemporal queries, computation and memory must scale tractably with sequence length and query count.

We introduce \textbf{MoGe-4D}, a feed-forward framework that realizes these goals through a unified query interface and a dual-encoder architecture (Fig.~\ref{fig:framework_overview}).
Given an unposed video, an ordered source--target frame query predicts paired \emph{allocentric} and \emph{egocentric} point maps, locating source-frame pixels at the target time in the source and target camera coordinate systems, respectively (Fig.~\ref{fig:framework_overview}, left).
Pairing both maps within a query directly captures dynamic geometry and relative camera pose, while varying source--target query combinations covers video depth estimation, multiview dynamic registration, and dense 3D tracking (Fig.~\ref{fig:framework_overview}, middle).
To keep dense evaluation tractable, we decouple scene modeling from query decoding via a \emph{context--query dual encoder} (Fig.~\ref{fig:framework_overview}, right): a high-capacity context encoder runs once at coarse resolution to cache global spatiotemporal features, while a lightweight query encoder re-encodes source images at flexible resolutions to decode dense point maps.
This design amortizes global video computation across queries and decouples output resolution from the context backbone at high throughput.

Supervising dense 4D prediction requires continuous 3D correspondences across arbitrary frames alongside surface geometry, posing a severe storage bottleneck for synthetic data pipelines.
Exploiting the local coherence of surface motion even under non-rigid deformation, we introduce \textbf{RigIDTrack}, a synthetic data pipeline and data format that clusters dense surface elements into rigidity-regularized affine bodies, compressing millions of dynamic mesh faces into a few thousand shared transformation trajectories.
This format recovers exact 3D trajectories for any pixel across arbitrary times and camera frames with only ${\sim}15\%$ storage overhead beyond standard RGB-D video, establishing a scalable foundation for dense 4D pretraining.

Across diverse multiview reconstruction and 3D tracking benchmarks, MoGe-4D outperforms existing tracking methods on all aggregate metrics, reducing world-space relative point error by 28\%--44\% across model variants.
At the same time, its unified predictions match dedicated state-of-the-art 3D reconstruction models such as $\pi^3$, DA3, and VGGT-$\Omega$ while surpassing prior unified 4D methods.
Meanwhile, our dual-encoder design substantially accelerates dense 4D reconstruction and tracking over existing baselines.

\section{Related Work}
\label{sec:related_work}

\begin{table}[!t]
    \centering
    \fontsize{8.5pt}{10.5pt}\selectfont
    \setlength{\tabcolsep}{4pt}
    \renewcommand{\arraystretch}{1.15}
    \caption{\textbf{Taxonomy of feed-forward 3D and 4D dynamic prediction paradigms.}
    We denote by $\boldsymbol{P}_{i\rightarrow j,k}$ the dense 3D point map of source frame $i$ evaluated at target time $j$ in camera $k$'s coordinate frame.
    Scene motion across timestamps can be represented as \emph{egocentric flow} $\boldsymbol{F}^{\mathrm{ego}}_{i\rightarrow j} = \boldsymbol{P}_{i\rightarrow j,j} - \boldsymbol{P}_{i\rightarrow i,i}$ (displacement relative to the observer) or \emph{allocentric flow} $\boldsymbol{F}^{\mathrm{allo}}_{i\rightarrow j} = \boldsymbol{P}_{i\rightarrow j,i} - \boldsymbol{P}_{i\rightarrow i,i}$ (motion mapped back to source camera coordinates).}
    \label{tab:paradigms}
    \resizebox{\textwidth}{!}{%
    \begin{tabular}{@{}llcccc@{}}
        \toprule
        \textbf{Paradigm} & \textbf{Method} & \textbf{Input Scope} & \textbf{Geometry Target} & \textbf{Motion Target} & \textbf{Camera Pose Recovery} \\
        \midrule
        \multirow{3}{*}{Feed-forward 3D} & VGGT~\citep{vggt} / VGGT-$\Omega$~\citep{vggtomega} & Sequence & $(\boldsymbol{P}_{i\rightarrow i,i}, \boldsymbol{P}_{i\rightarrow i,0})$ & --- & Direct regression \\
        & $\pi^3$~\citep{pi3} & Sequence & $\boldsymbol{P}_{i\rightarrow i,i}$ & --- & Direct regression \\
        & Depth Anything 3~\citep{depthanything3} & Sequence & $\boldsymbol{P}_{i\rightarrow i,0}$ & --- & Direct regression / Ray solve \\
        \midrule
        \multirow{2}{*}{Pairwise point maps} & DPM~\citep{dpm} & 2-view & $(\boldsymbol{P}_{0\rightarrow 0,0}, \boldsymbol{P}_{1\rightarrow 1,0})$ & $(\boldsymbol{P}_{0\rightarrow 1,0}, \boldsymbol{P}_{1\rightarrow 0,0})$ & RANSAC PnP \\
        & St4RTrack~\citep{st4rtrack} & 2-view & $\boldsymbol{P}_{1\rightarrow 1,0}$ & $\boldsymbol{P}_{0\rightarrow 1,0}$ & RANSAC PnP \\
        \midrule
        \multirow{4}{*}{Decoupled flow \& geometry} & Flow4R~\citep{flow4r} & 2-view & $\boldsymbol{P}_{0\rightarrow 0,0}$ & $\boldsymbol{F}^{\mathrm{ego}}_{0\rightarrow 1}$ & RANSAC Umeyama \\
        & 4RC~\citep{4rc} & Sequence & $\boldsymbol{P}_{i\rightarrow i,i}$ & $\boldsymbol{F}^{\mathrm{allo}}_{i\rightarrow j}$ & Direct regression \\
        & Any4D~\citep{any4d} & Sequence & $\boldsymbol{P}_{i\rightarrow i,i}$ & $\boldsymbol{F}^{\mathrm{allo}}_{0\rightarrow j}$ & Direct regression \\
        & Track4World~\citep{track4world} & Sequence & $\boldsymbol{P}_{i\rightarrow i,i}$ & $\boldsymbol{F}^{\mathrm{ego}}_{i\rightarrow j}$ or $(\boldsymbol{F}^{\mathrm{ego}}_{0\rightarrow 0}, \dots, \boldsymbol{F}^{\mathrm{ego}}_{0\rightarrow T-1})$ & Direct regression \\
        \midrule
        Sequence point maps & V-DPM~\citep{vdpm} & Sequence & $\boldsymbol{P}_{i\rightarrow i,0}$ & $(\boldsymbol{P}_{0\rightarrow j,0}, \dots, \boldsymbol{P}_{T-1\rightarrow j,0})$ & Direct regression \\
        \midrule
        Unified sparse point query & D4RT~\citep{d4rt} & Sequence & \multicolumn{2}{c}{Unified query: $p(u,v,i,j,k) \in \mathbb{R}^3$} & Closed-form Umeyama \\
        \midrule
        \textbf{Unified point map query} & \textbf{MoGe-4D} & \textbf{Sequence} & \multicolumn{2}{c}{\textbf{Unified query: } $(\boldsymbol{P}_{i\rightarrow j,i}, \boldsymbol{P}_{i\rightarrow j,j})$} & \textbf{Closed-form Umeyama} \\
        \bottomrule
    \end{tabular}%
    }
\end{table}

\subsection{Feed-forward 3D reconstruction}

Feed-forward 4D models build on advances in 3D reconstruction.
Classical methods estimate geometry and cameras via iterative optimization~\citep{colmap,mvsnet}, which DROID-SLAM~\citep{droidslam} and MegaSaM~\citep{megasam} augment with learned correspondence, depth, or flow priors, while monocular models~\citep{marigold,depthanything,moge,pixelperfectdepth,infinidepth} scale learned priors for direct single-image geometry prediction.
For multiview reconstruction, DUSt3R~\citep{dust3r} directly regresses pixel-aligned point maps from unposed image pairs, with extensions improving matching~\citep{mast3r}, scaling to multiple views~\citep{mvdust3r,fast3r,dens3r,more}, maintaining persistent state~\citep{spann3r,cut3r,stream3r}, accepting geometric inputs~\citep{mapanything}, or handling dynamic scenes~\citep{monst3r,easi3r}.
VGGT~\citep{vggt} further scales joint camera and geometry prediction, followed by $\pi^3$, Depth Anything 3, and VGGT-$\Omega$ with reference-free point maps, depth--ray prediction, and larger-scale training~\citep{pi3,depthanything3,vggtomega}.
Most related in architecture, DAGE~\citep{dage} pairs $\pi^3$ with MoGe-2~\citep{moge2} in a dual-stream design for high-resolution 3D geometry.

\subsection{Dynamic reconstruction and tracking}

Optimization-based methods fit scene-specific dynamic NeRFs~\citep{dnerf,nerfies,hypernerf,nsff}, 3D Gaussians~\citep{dynamic3dgaussians,gaussian4d}, or low-rank motion bases~\citep{shapeofmotion} using depth and track priors.
For feed-forward 3D tracking, SpatialTracker~\citep{spatialtracker,spatialtrackerv2} and TAPIP3D~\citep{tapip3d} track queried points via correspondence and geometric reasoning, whereas Trace Anything~\citep{traceanything} regresses continuous-time B-spline trajectories.
Joint 4D models extend point-map regression to motion: DPM~\citep{dpm} and St4RTrack~\citep{st4rtrack} predict pairwise dynamic point maps, Flow4R~\citep{flow4r} predicts points and scene flow, V-DPM~\citep{vdpm} extends VGGT to video point maps, 4RC~\citep{4rc} and Any4D~\citep{any4d} decode geometry and allocentric motion from shared video features, and Track4World~\citep{track4world} uses 3D correlation for dense flow.

\subsection{Unified 4D representations}

Table~\ref{tab:paradigms} compares feed-forward 4D prediction paradigms across input scope, motion representation, and query granularity.
Pairwise point-map methods~\citep{dpm,st4rtrack} operate on two-view inputs without sequence-level context, coupling distinct tasks across two source images (cross-time motion for one view and cross-camera geometry for the other) in a single pass with separate decoders and inter-map redundancy.
Decoupled flow-and-geometry methods~\citep{flow4r,4rc,any4d,track4world} separate static geometry from scene flow, leveraging flow inductive biases at the cost of a coordinate trade-off: allocentric flow removes camera motion yet suffers from relative-motion ambiguity without static references, whereas egocentric flow entangles camera and object motion.
V-DPM~\citep{vdpm} scales dynamic point maps to sequences, but anchors all predictions to a single privileged frame, accumulating coordinate drift away from the reference while requiring repeated evaluation of a heavy conditional decoder across target times.
As a conceptual turning point, D4RT~\citep{d4rt} formalizes a unified spatiotemporal query space, yet its unconstrained query axes and point-wise granularity make dense per-frame decoding computationally expensive.

MoGe-4D instead adopts a concise unified point map query interface (Table~\ref{tab:paradigms}, bottom), offering:
(i) \emph{Both allocentric and egocentric dynamics:} combining the advantages of world-referenced motion and observer-relative coordinates.
(ii) \emph{Shared intra-query computation:} because both maps in a query depict the same geometry up to a rigid coordinate change, they share feature computation and diverge only at final output heads.
(iii) \emph{Relative pose from dense geometry:} rather than regressing poses in a black-box manner, pairing the two maps terminates network prediction at dense 3D coordinates, preserving interpretable geometric residuals while yielding relative poses in closed form from a single query.
(iv) \emph{Privileged-reference-free:} operating in each query's local camera space avoids anchoring to a privileged frame, preventing coordinate drift from degrading global predictions.


\section{MoGe-4D}
\label{sec:method}

\begin{figure*}[t]
    \centering
    \includegraphics[width=\textwidth]{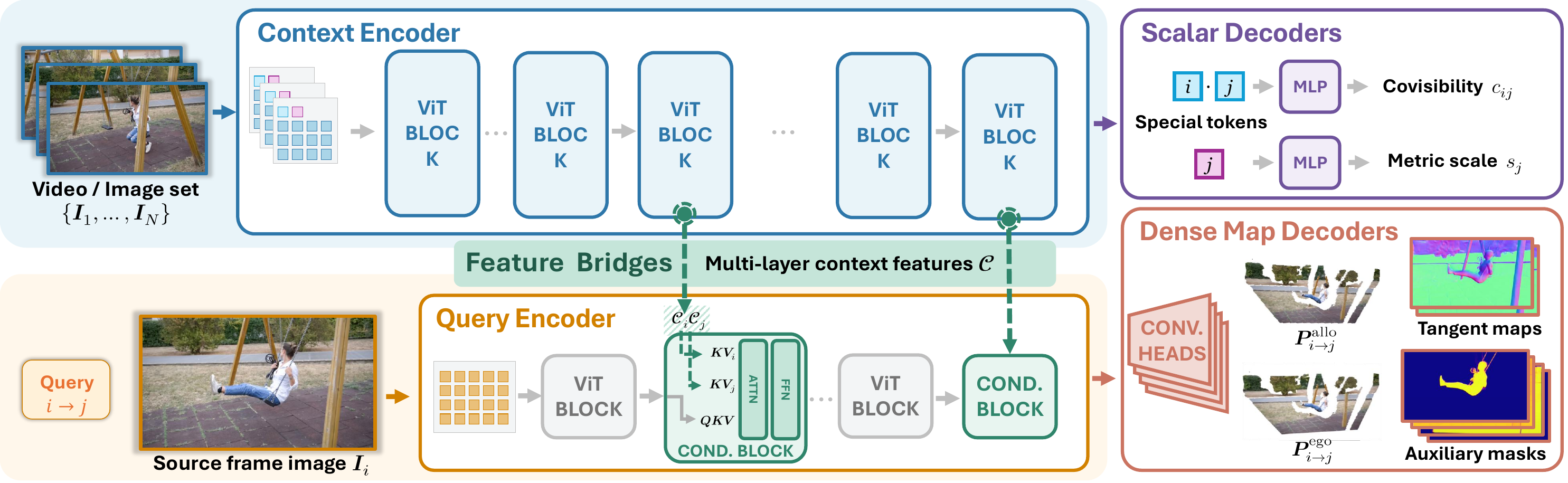}
    \caption{\textbf{MoGe-4D architecture.} A high-capacity context encoder processes the input sequence once at coarse resolution and caches multi-layer features. For each query $(i,j)$, a lightweight query encoder re-encodes source image $i$ at flexible resolution, attending to bridged source and target context tokens via conditioned attention blocks. Convolutional heads then decode point maps and auxiliary maps, while MLP heads predict scalars from special context tokens.}
    \label{fig:model_overview}
\end{figure*}

\subsection{Formulation}
\label{sec:formulation}

Given an unposed sequence of RGB frames $\mathcal{I}=\{\boldsymbol{I}_t\}_{t=1}^{T}$, let $\boldsymbol{P}_{i\rightarrow j,k}\in\mathbb{R}^{H_i\times W_i\times 3}$ denote the dense point map aligned with the pixel grid of \emph{source} frame $i$ that locates those pixels' 3D scene points at \emph{target} time $j$ (including points occluded or outside the field of view in frame $j$), expressed in the coordinate system of camera $k$.
For an ordered query $(i,j)$, MoGe-4D predicts paired point maps in the source and target camera coordinate systems:
\begin{equation}
    \left(\boldsymbol{P}^{\mathrm{allo}}_{i\rightarrow j},\boldsymbol{P}^{\mathrm{ego}}_{i\rightarrow j}\right)
    =
    \left(\boldsymbol{P}_{i\rightarrow j,i},\boldsymbol{P}_{i\rightarrow j,j}\right).
    \label{eq:point_map_query}
\end{equation}
While ideally related by a rigid camera transformation, the two predictions exhibit complementary inductive biases in practice: the \emph{allocentric} map $\boldsymbol{P}^{\mathrm{allo}}_{i\rightarrow j}$ anchors motion to source view $i$, naturally preserving static background and near-stationary structures even after they leave the target view, whereas the \emph{egocentric} map $\boldsymbol{P}^{\mathrm{ego}}_{i\rightarrow j}$ operates relative to target view $j$, more accurately tracking dynamic foreground objects that remain observed by the moving camera.
Coupling both maps in a single query thus combines stability on static structure with precision on large-displacement foregrounds, while directly constraining the relative camera pose between frames $i$ and $j$.

To decouple geometry prediction from camera focal-length and scale ambiguities, we follow MoGe~\citep{moge} and predict affine-invariant point maps $\left(\widehat{\boldsymbol{P}}^{\mathrm{allo}}_{i\rightarrow j},\widehat{\boldsymbol{P}}^{\mathrm{ego}}_{i\rightarrow j}\right)$, which relate to camera-space coordinates via a per-camera scale $\alpha_k>0$ and optical-axis shift $\delta_k$ (recovered at inference in Sec.~\ref{sec:inference}):
\begin{equation}
    \boldsymbol{P}^{\mathrm{allo}}_{i\rightarrow j}=\alpha_i\left(\widehat{\boldsymbol{P}}^{\mathrm{allo}}_{i\rightarrow j}+\delta_i\boldsymbol{e}_z\right)
    ,\qquad
    \boldsymbol{P}^{\mathrm{ego}}_{i\rightarrow j}=\alpha_j\left(\widehat{\boldsymbol{P}}^{\mathrm{ego}}_{i\rightarrow j}+\delta_j\boldsymbol{e}_z\right),
    \label{eq:point_map_scale_shift}
\end{equation}
where $\boldsymbol{e}_z=[0,0,1]^\top$ is the optical-axis unit vector.

\subsection{Architecture}
\label{sec:architecture}

MoGe-4D pairs a high-capacity context encoder with a lightweight query encoder (Fig.~\ref{fig:model_overview}).
The context encoder processes the input sequence once, while the query encoder re-encodes each source image conditioned on the cached source and target context:
\begin{equation}
    \mathcal{C}=E_c(\mathcal{I}),
    \qquad
    \boldsymbol{Z}_{i\rightarrow j}
    =E_q\!\left(\boldsymbol{I}_i;\mathcal{C}_i,\mathcal{C}_j\right),
    \label{eq:dual_encoder}
\end{equation}
where $\mathcal{C}_t$ denotes the cached multi-layer context features of frame $t$, and $\boldsymbol{Z}_{i\rightarrow j}$ denotes the source-aligned query features for pair $(i,j)$.

\paragraph{Context encoder.}
To relate observations across viewpoints and time, the context encoder applies a high-capacity ViT with cross-frame attention over the full sequence, concentrating over 90\% of the model parameters in this stage while operating on a coarse spatial token grid to curb the quadratic cost of global attention on long videos.
Multi-layer spatial features from each frame are computed once and cached, amortizing scene-wide spatiotemporal reasoning across all subsequent reconstruction and tracking queries.
To explore different pretrained representations for this role, we train three model variants initialized from DINOv2, V-JEPA 2.1, and VGGT-$\Omega$.

\paragraph{Feature bridges.}
Feature bridges project selected context-layer features into the representation spaces of corresponding query blocks via LayerNorm and a linear layer, and are initialized with a standalone feature-alignment stage before end-to-end training to bridge both encoders' pretrained representations.

\paragraph{Query encoder.}
For each query $(i,j)$, a pretrained ViT-Base query encoder re-encodes source image $i$ to recover fine spatial detail omitted by the coarse context grid.
At conditioned blocks, bridged context features from frames $i$ and $j$ supply additional keys and values alongside the query tokens, fusing local visual detail with global scene context while keeping tokens aligned with the source image grid.
Because this conditioning requires no spatial grid matching between streams, query token budgets can be adjusted freely at inference to trade off spatial detail and speed without recomputing the cached context.

\paragraph{Decoders.}
From the source-aligned features $\boldsymbol{Z}_{i\rightarrow j}$, a shared convolutional neck branches into separate heads that decode the paired point maps $(\widehat{\boldsymbol{P}}^{\mathrm{allo}}_{i\rightarrow j}, \widehat{\boldsymbol{P}}^{\mathrm{ego}}_{i\rightarrow j})$, surface tangents, point-wise uncertainty, and masks for visibility, sky, and dynamic regions.
Meanwhile, lightweight MLP heads operate directly on context tokens to predict metric scale and pairwise covisibility.
Architectural specifications and auxiliary output definitions are detailed in Appendix~\ref{sec:supp_d4d_architecture}.

\begin{figure}[!t]
    \centering
    \includegraphics[width=\linewidth]{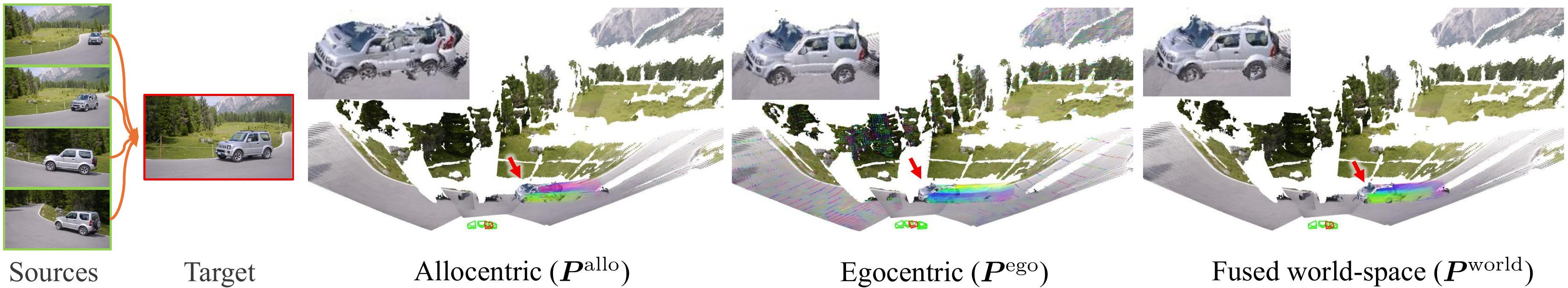}
    \caption{\textbf{Allocentric, egocentric, and fused world-space predictions.}
    Under large viewpoint change, $\boldsymbol{P}^{\mathrm{allo}}_{i\rightarrow j}$ preserves static backgrounds but can ghost on moving objects, whereas $\boldsymbol{P}^{\mathrm{ego}}_{i\rightarrow j}$ accurately tracks dynamic foregrounds but can jitter on backgrounds; fusing both into $\boldsymbol{P}^{\mathrm{world}}_{i\rightarrow j}$ combines their strengths.}
    \label{fig:allo_ego_world}
\end{figure}

\subsection{Training}
\label{sec:training}

We train MoGe-4D end to end on a mixture of synthetic and real, dynamic and static multiview datasets, sampling $O(T)$ source--target queries per clip to keep decoding costs manageable (full loss definitions and schedules in Appendices~\ref{sec:supp_d4d_training} and~\ref{sec:supp_d4d_training_setup}).
After aligning the affine-invariant predictions to ground truth in each camera frame via ROE scale--shift alignment~\citep{moge}, we supervise the model with three groups of objectives (using inverse-distance-weighted $L_1$ errors for both geometry and motion):
\begin{itemize}\setlength{\itemsep}{3pt}\setlength{\parskip}{0pt}\setlength{\topsep}{4pt}
    \item \textbf{Geometry losses.} A \emph{global point-map loss} supervises overall scene layout, while a \emph{multi-scale local point-map loss}~\citep{moge3} fits local translations to supervise regional shape, complemented by an edge loss on neighboring-point differences for surface continuity.
    \item \textbf{Motion losses.} To further reinforce temporal coherence, we penalize errors on point-map differences between adjacent sampled target times for each source frame, fitting translation offsets to absorb bulk displacement and camera-origin residuals. Applied at two granularities, a \emph{global motion loss} constrains scene-wide dynamics, while a \emph{local motion loss} isolates intra-region rotation, articulation, and deformation.
    \item \textbf{Auxiliary losses.} Dedicated objectives supervise surface tangents, point-wise uncertainty, visibility/sky/dynamic masks, metric scale, and pairwise covisibility.
\end{itemize}

\paragraph{Training curriculum.}
Training begins with shorter clips and smaller query-token budgets on dynamic tracking datasets to establish cross-time prediction at low cost, then progressively scales both budgets while expanding to the full mixture of dynamic and static multiview data.

\subsection{Inference}
\label{sec:inference}

At inference, MoGe-4D first recovers camera intrinsics and poses, then evaluates task-specific query combinations for 4D reconstruction and tracking.

\paragraph{Camera focal length and shift recovery.}
For each self-query $(i,i)$, $\widehat{\boldsymbol{P}}^{\mathrm{allo}}_{i\rightarrow i}$ and $\widehat{\boldsymbol{P}}^{\mathrm{ego}}_{i\rightarrow i}$ are identical by definition and directly satisfy 2D--3D perspective projection constraints~\citep{moge}.
Assuming a centered principal point, we jointly recover the optical-axis shift $\delta_i$ (Eq.~\ref{eq:point_map_scale_shift}) and focal length (either per-frame or sequence-shared) by minimizing weighted reprojection error, applying the recovered $\delta_i$ to all predictions expressed in camera $i$'s coordinate system.

\paragraph{Camera pose recovery.}
For any query $(i,j)$, the paired point maps $(\boldsymbol{P}^{\mathrm{allo}}_{i\rightarrow j}, \boldsymbol{P}^{\mathrm{ego}}_{i\rightarrow j})$ provide direct 3D--3D correspondences of the same physical surface across the two views, yielding the relative pose between cameras $i$ and $j$ in closed form via rigid alignment weighted by predicted uncertainty and dynamic/sky masks.
Across long sequences, rather than evaluating all $T^2$ pairs, we connect views with high predicted covisibility $c_{ij}$ into a sparse pose graph to synchronize global camera poses $(\boldsymbol{R}_j,\boldsymbol{t}_j)$ and optional relative scales $\alpha_j$.

\paragraph{Reconstruction and tracking in world space.}
Using the estimated camera poses, we transform both predictions of query $(i,j)$ into world coordinates and fuse them into a single world-space map $\boldsymbol{P}^{\mathrm{world}}_{i\rightarrow j}$:
\begin{equation}
    \boldsymbol{P}^{\mathrm{world}}_{i\rightarrow j}=(1-M^{\mathrm{dyn}}_{i\rightarrow j})\boldsymbol{R}_i^\top\left(\boldsymbol{P}^{\mathrm{allo}}_{i\rightarrow j}-\boldsymbol{t}_i\right)+M^{\mathrm{dyn}}_{i\rightarrow j}\boldsymbol{R}_j^\top\left(\boldsymbol{P}^{\mathrm{ego}}_{i\rightarrow j}-\boldsymbol{t}_j\right),
    \label{eq:world_point_map}
\end{equation}
where $M^{\mathrm{dyn}}_{i\rightarrow j}\in[0,1]$ smoothly blends the two maps (Fig.~\ref{fig:allo_ego_world}), leveraging $\boldsymbol{P}^{\mathrm{allo}}_{i\rightarrow j}$'s stability on static backgrounds and $\boldsymbol{P}^{\mathrm{ego}}_{i\rightarrow j}$'s accuracy on moving foregrounds.

These fused world-space maps directly service downstream tasks through query selection.
For \emph{dynamic reconstruction} at target time $j$, we aggregate maps $\{\boldsymbol{P}^{\mathrm{world}}_{i\rightarrow j}\}$ across multiple source views, optionally selecting a compact subset of visibility-guided \emph{support frames} to reduce decoded queries while preserving scene coverage.
For \emph{dense 3D tracking}, holding source frame $i$ fixed while sweeping target time $j\in\{1,\dots,T\}$ traces continuous 3D world trajectories for all source pixels, from which camera-free scene flow follows directly by temporal differencing.
Camera recovery and support-frame selection algorithms are detailed in Appendix~\ref{sec:supp_d4d_inference}.


\begin{figure}[!t]
	\centering
	\includegraphics[width=\linewidth]{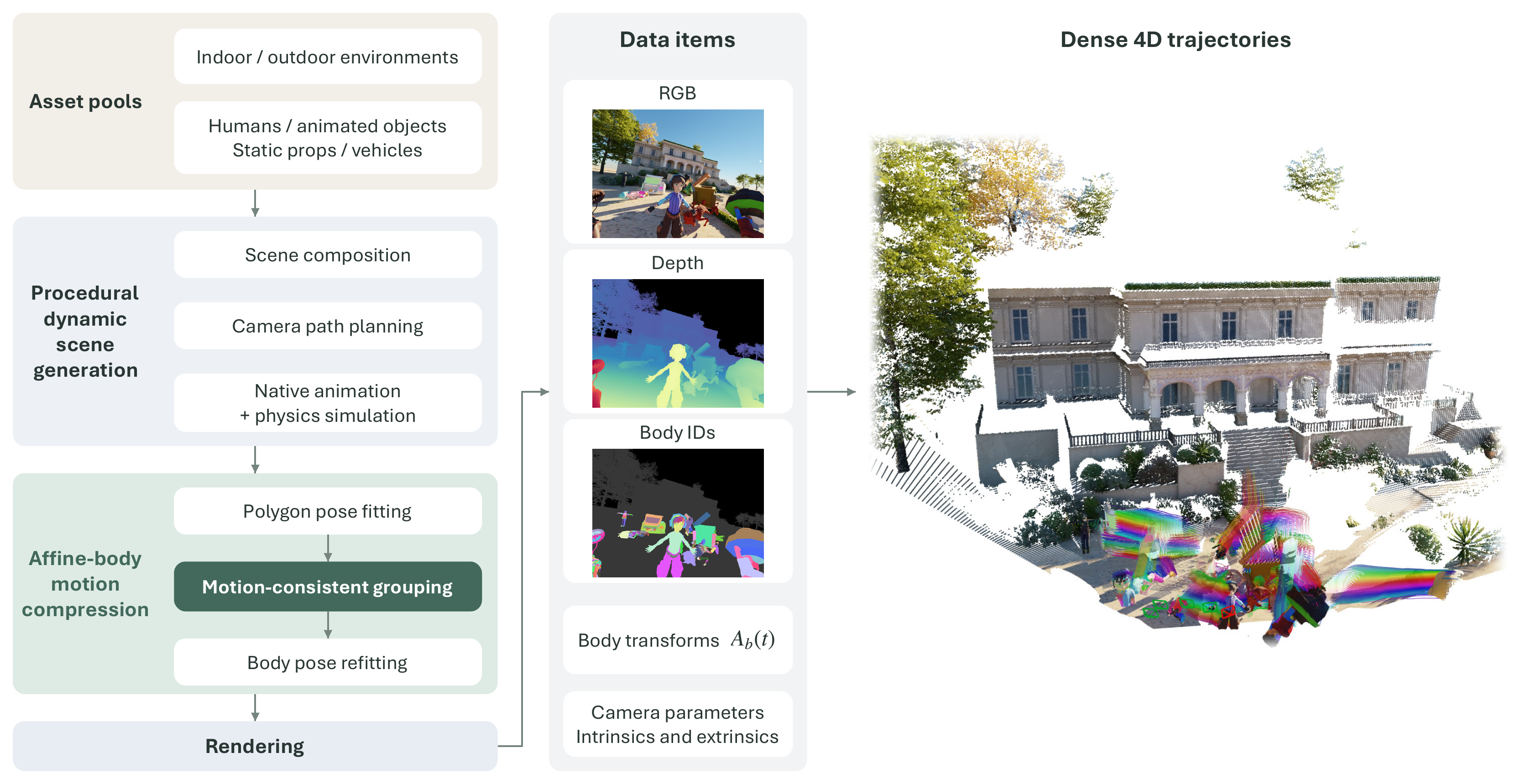}
	\caption{\textbf{RigIDTrack data pipeline.} Dynamic scenes are procedurally assembled from 3D asset pools with animation, physics simulation, and collision-aware camera paths. Affine-body motion compression groups motion-consistent mesh faces into shared transformation trajectories, from which dense 4D supervision is recovered via per-frame depth, body-ID maps, and body transforms.}
	\label{fig:rigidtrack_pipeline}
\end{figure}

\begin{figure}[!t]
	\centering
	\begin{minipage}[t]{0.245\linewidth}
		\centering
		{\footnotesize\strut RGB\par}
		\includegraphics[width=\linewidth]{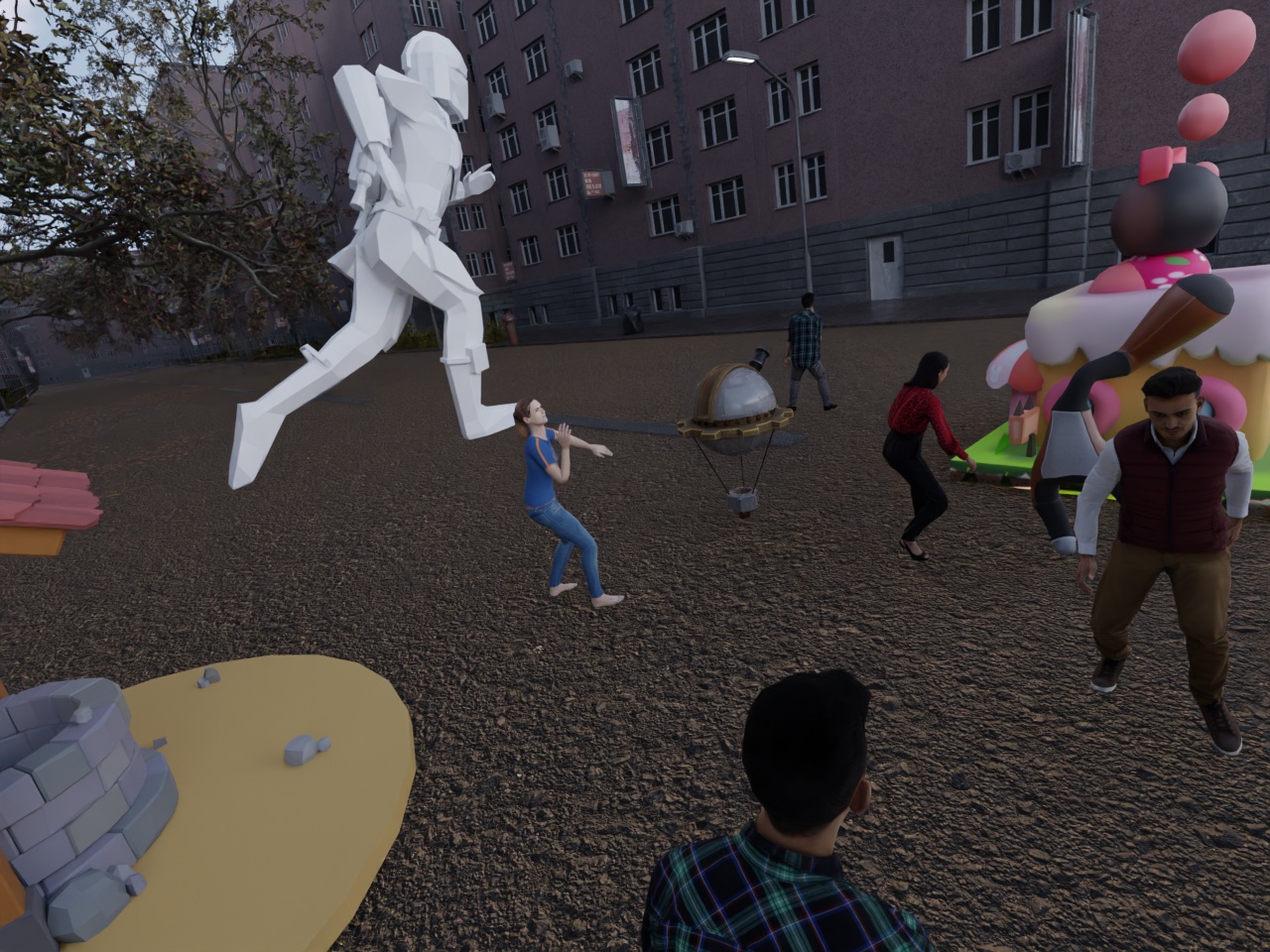}
	\end{minipage}\hfill
	\begin{minipage}[t]{0.245\linewidth}
		\centering
		{\footnotesize\strut All faces (6.7M)\par}
		\includegraphics[width=\linewidth]{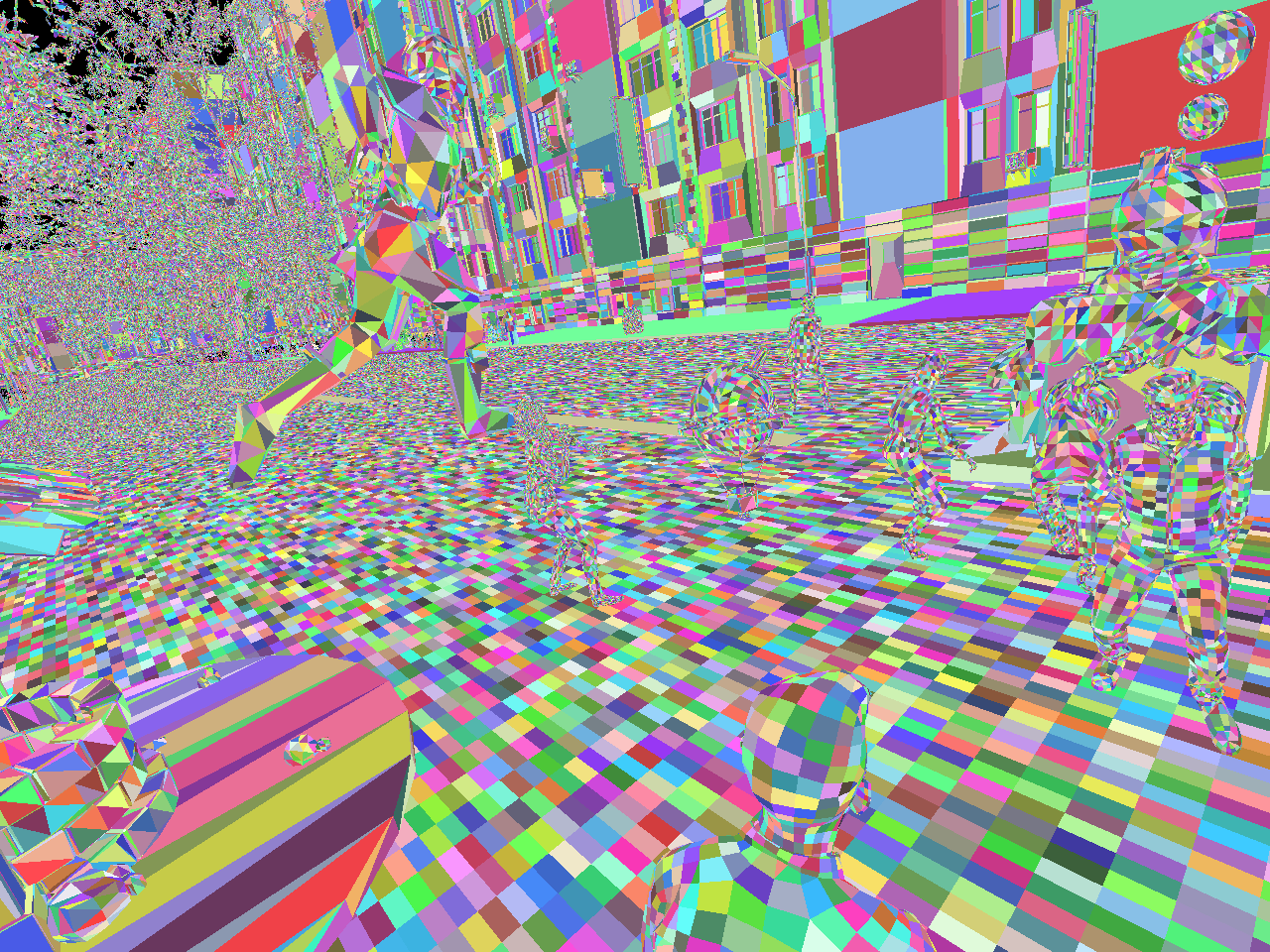}
	\end{minipage}\hfill
	\begin{minipage}[t]{0.245\linewidth}
		\centering
		{\footnotesize\strut Dynamic faces (2.3M)\par}
		\includegraphics[width=\linewidth]{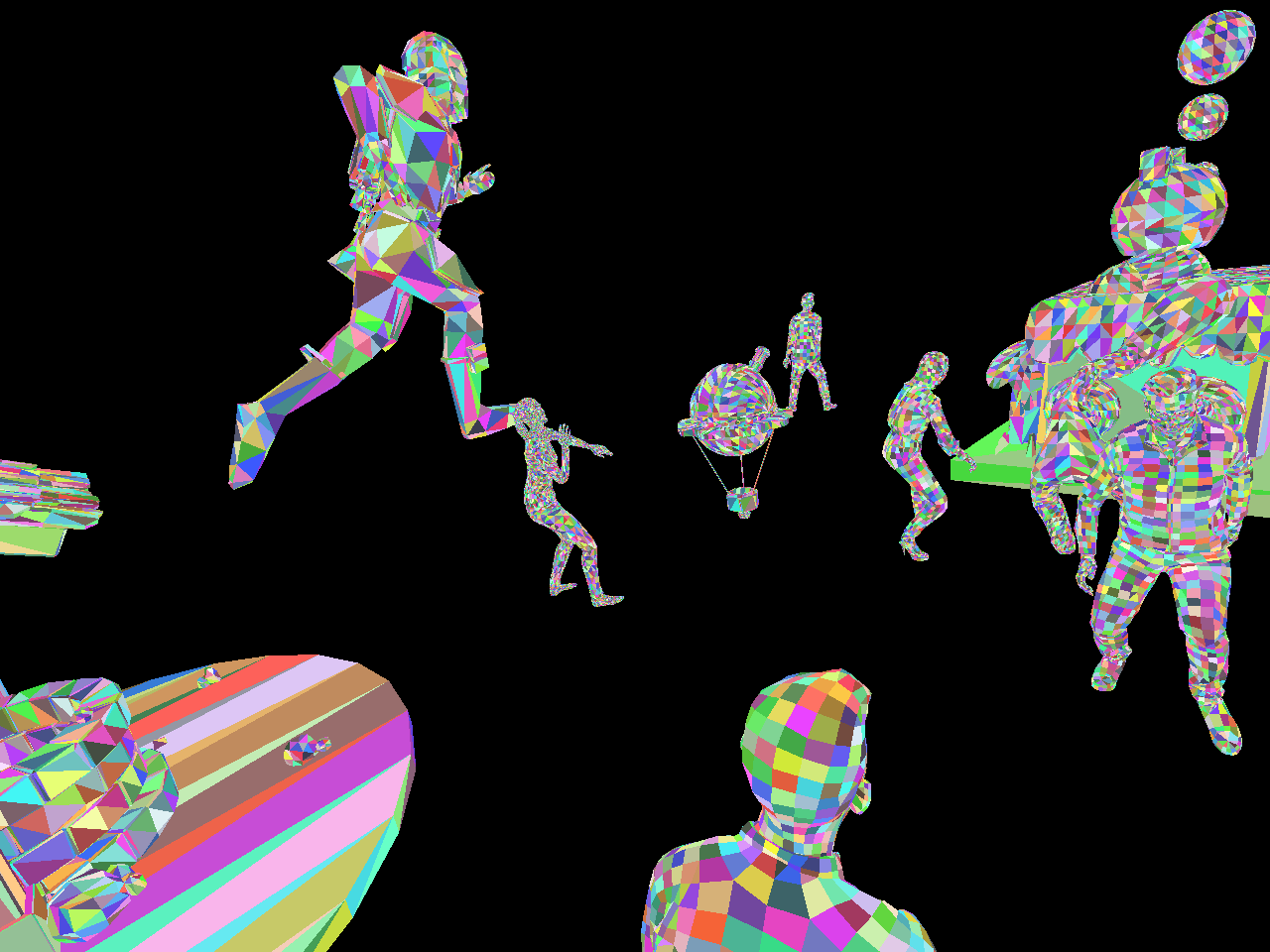}
	\end{minipage}\hfill
	\begin{minipage}[t]{0.245\linewidth}
		\centering
		{\footnotesize\strut Compressed bodies (3.2K)\par}
		\includegraphics[width=\linewidth]{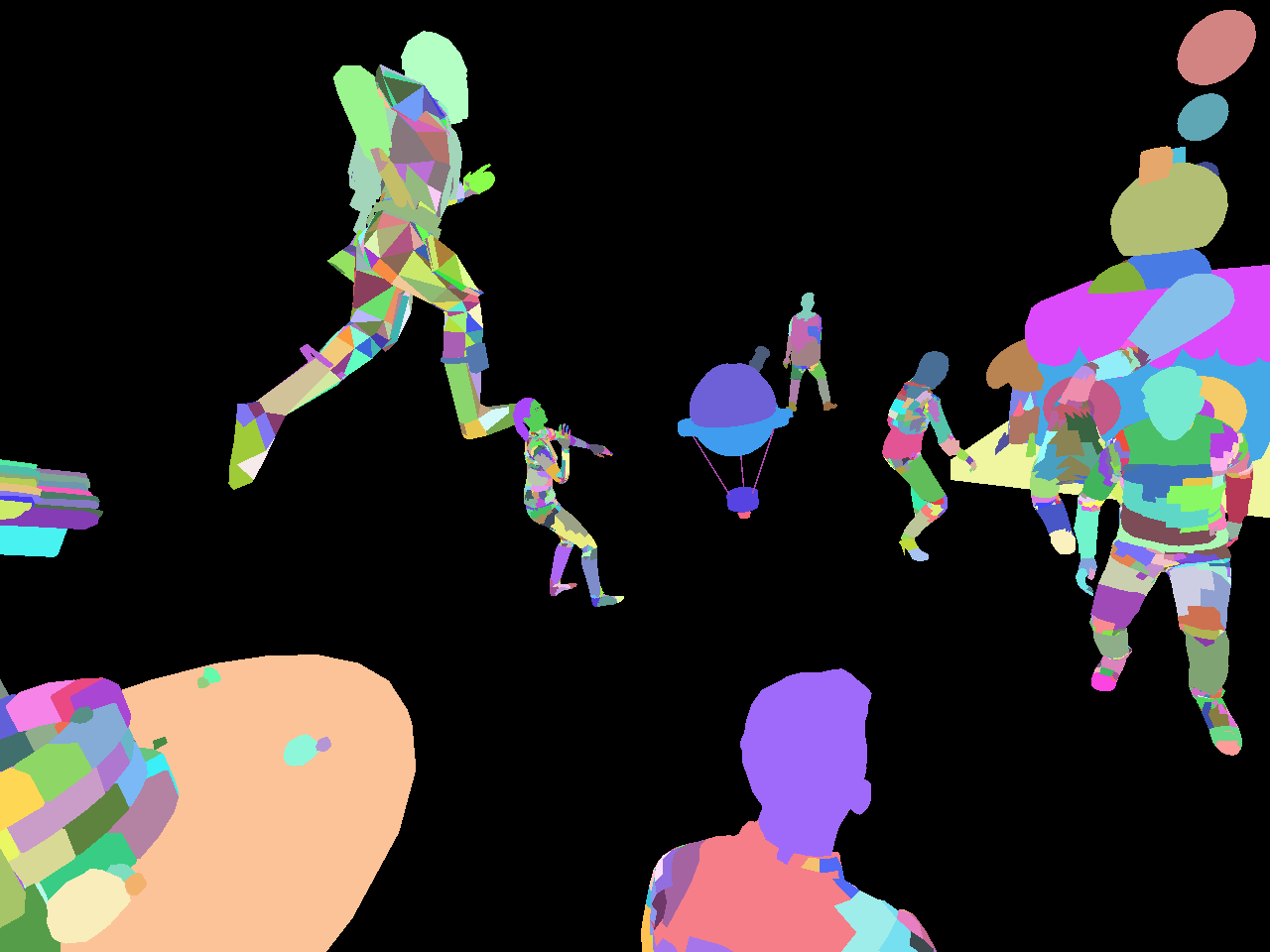}
	\end{minipage}
	\caption{\textbf{Affine-body motion compression.} Dynamic mesh faces are clustered into motion-consistent affine bodies (shown as contiguous regions of shared color), reducing stored transformation trajectories by over $700\times$.}
	\label{fig:rigidtrack_motion_compression}
\end{figure}

\section{RigIDTrack}
\label{sec:rigidtrack}

Training dense 4D models on synthetic data requires persistent cross-frame correspondences through object motion, camera motion, and occlusion.
Yet storing explicit trajectories for every pixel or mesh face is prohibitively expensive and locally redundant, as nearby surfaces move coherently even on articulated or deforming objects.
We introduce \textbf{RigIDTrack}, a synthetic data pipeline and representation format that compresses dense surface motion into rigidity-regularized affine bodies, i.e., surface clusters sharing transformation trajectories (Fig.~\ref{fig:rigidtrack_pipeline}).
This compact format supports dense 4D supervision across arbitrary source pixels, target times, and camera frames (see Appendix~\ref{sec:supp_data} for full pipeline and algorithm details).

\paragraph{Dynamic scene generation.}
We procedurally compose indoor and outdoor environments in Blender from diverse 3D scene and object assets, combining skeletal animation, non-rigid deformation, and rigid-body physics.
Independently moving cameras follow smooth, collision-aware paths that alternate between free exploration and tracking nearby objects, producing varied object dynamics, viewpoint changes, and occlusions.

\paragraph{Affine-body motion compression.}
To compress surface motion, we first fit rigidity-regularized per-face affine trajectories to capture local stretch and shear beyond pure rigid motion.
We then cluster faces with consistent relative motion under a prescribed error tolerance and refit a shared transformation trajectory for each body (with static scene geometry grouped into a single background body).
This formulation operates directly on observed mesh motion without requiring predefined rigid parts, skeletons, or skinning weights, naturally accommodating rigid, articulated, and general deforming assets alike.
In practice, it compresses millions of dynamic faces into a few thousand trajectories with sub-pixel level reprojection error (Fig.~\ref{fig:rigidtrack_motion_compression}).

\paragraph{Dense tracking ground truth recovery.}
For each rendered frame $t$, we store the RGB image, depth $D_t(\boldsymbol{u})$, intrinsics $K_t$, world-to-camera extrinsics $E_t$, pixelwise body ID $B_t(\boldsymbol{u})$, and per-body world affine transforms $\{A_{b,t}\}$.
Unlike mesh-based formats that require per-pixel triangle IDs and barycentric interpolation on full meshes~\citep{syn4d}, associating each pixel with a single body ID enables direct coordinate transport without storing or querying mesh geometry.
For a valid source pixel $\boldsymbol{u}$ belonging to body $b=B_i(\boldsymbol{u})$, its 3D position at target time $j$ in camera frame $k$ is recovered as
\begin{equation}
		\begin{bmatrix}
			\boldsymbol{P}^{\mathrm{gt}}_{i\rightarrow j,k}(\boldsymbol{u}) \\
			1
		\end{bmatrix}
		= E_k A_{b,j}A_{b,i}^{-1}E_i^{-1}
			\begin{bmatrix}
				D_i(\boldsymbol{u})K_i^{-1}[\boldsymbol{u}, 1]^\top \\
				1
			\end{bmatrix}.
		\label{eq:rigidtrack_transport}
\end{equation}
This transport depends only on frames $(i,j,k)$ via a single $4\times4$ matrix chain per body, taking $O(1)$ time regardless of the temporal separation between source and target frames.

\section{Experiments}
\label{sec:experiments}

We evaluate MoGe-4D on 3D tracking and multiview reconstruction across three context-encoder variants initialized from DINOv2, V-JEPA 2.1, and VGGT-$\Omega$, using default budgets of 500 context tokens and 900 map-query tokens per frame.

\subsection{3D Tracking}
\label{sec:tracking_3d}

\begin{table}[!t]
    \captionsetup{hypcap=false}
    \centering
    \fontsize{8.5pt}{10.2pt}\selectfont
    \setlength{\aboverulesep}{0pt}
    \setlength{\belowrulesep}{0pt}
    \renewcommand{\arraystretch}{1.1}
    \captionof{table}{\textbf{Quantitative comparison on (a) 3D tracking and (b) multiview reconstruction.}
    Results are averaged across the respective benchmarks (see Appendix~\ref{sec:supp_per_benchmark} for individual benchmark results); darker shading indicates a better rank within each metric.}
    \label{tab:tracking_3d}
    \label{tab:multiview_reconstruction}
    \setlength{\tabcolsep}{3pt}
    \begin{tabularx}{\textwidth}{@{}l*{9}{Y}@{}}
    \multicolumn{10}{c}{\textbf{(a) 3D tracking}} \\
    \hline
    \multirow[c]{3}{*}{Method}
    & \multirow{3}{*}{OA$\uparrow$}
    & \multicolumn{4}{c}{Camera space}
    & \multicolumn{4}{c}{World space} \\
    \cmidrule(lr){3-6}\cmidrule(lr){7-10}
    & & \multicolumn{2}{c}{All} & \multicolumn{2}{c}{Visible}
    & \multicolumn{2}{c}{All} & \multicolumn{2}{c}{Visible} \\
    \cmidrule(lr){3-4}\cmidrule(lr){5-6}\cmidrule(lr){7-8}\cmidrule(lr){9-10}
    & & AUC$\uparrow$ & RelEPE$\downarrow$
    & AUC$\uparrow$ & RelEPE$\downarrow$
    & AUC$\uparrow$ & RelEPE$\downarrow$
    & AUC$\uparrow$ & RelEPE$\downarrow$ \\
    \hline
    St4RTrack & - & \cellcolor{highlight!0}17.38 & \cellcolor{highlight!0}0.659 & \cellcolor{highlight!0}20.53 & \cellcolor{highlight!0}0.557 & \cellcolor{highlight!0}29.52 & \cellcolor{highlight!0}0.394 & \cellcolor{highlight!0}34.74 & \cellcolor{highlight!0}0.343 \\
    Open-D4RT & \cellcolor{highlight!0}84.16 & \cellcolor{highlight!7}56.79 & \cellcolor{highlight!7}0.178 & \cellcolor{highlight!7}60.07 & \cellcolor{highlight!3}0.153 & \cellcolor{highlight!7}65.61 & \cellcolor{highlight!7}0.158 & \cellcolor{highlight!7}69.03 & \cellcolor{highlight!3}0.142 \\
    Any4D & - & \cellcolor{highlight!1}45.94 & \cellcolor{highlight!1}0.218 & \cellcolor{highlight!1}49.98 & \cellcolor{highlight!1}0.180 & \cellcolor{highlight!3}61.96 & \cellcolor{highlight!1}0.178 & \cellcolor{highlight!1}66.54 & \cellcolor{highlight!1}0.151 \\
    TraceAnything & - & \cellcolor{highlight!0}28.66 & \cellcolor{highlight!0}0.370 & \cellcolor{highlight!0}31.33 & \cellcolor{highlight!0}0.305 & \cellcolor{highlight!0}52.78 & \cellcolor{highlight!0}0.244 & \cellcolor{highlight!0}56.25 & \cellcolor{highlight!0}0.224 \\
    V-DPM & - & \cellcolor{highlight!14}66.11 & \cellcolor{highlight!24}0.160 & \cellcolor{highlight!14}69.76 & \cellcolor{highlight!24}0.133 & \cellcolor{highlight!24}72.65 & \cellcolor{highlight!24}0.138 & \cellcolor{highlight!24}76.92 & \cellcolor{highlight!24}0.121 \\
    Track4World & \cellcolor{highlight!1}84.96 & \cellcolor{highlight!3}53.91 & \cellcolor{highlight!3}0.192 & \cellcolor{highlight!3}59.94 & \cellcolor{highlight!7}0.146 & \cellcolor{highlight!1}59.48 & \cellcolor{highlight!3}0.169 & \cellcolor{highlight!3}66.96 & \cellcolor{highlight!14}0.125 \\
    4RC & - & \cellcolor{highlight!24}67.22 & \cellcolor{highlight!14}0.172 & \cellcolor{highlight!24}70.70 & \cellcolor{highlight!14}0.145 & \cellcolor{highlight!14}71.86 & \cellcolor{highlight!14}0.152 & \cellcolor{highlight!14}76.43 & \cellcolor{highlight!7}0.131 \\
    \hline
    \textbf{Ours} \scriptsize(DINO) & \cellcolor{highlight!80}88.82 & \cellcolor{highlight!38}73.92 & \cellcolor{highlight!38}0.098 & \cellcolor{highlight!38}78.36 & \cellcolor{highlight!56}0.072 & \cellcolor{highlight!38}77.22 & \cellcolor{highlight!38}0.099 & \cellcolor{highlight!38}81.91 & \cellcolor{highlight!38}0.077 \\
    \textbf{Ours} \scriptsize(VJEPA) & \cellcolor{highlight!34}88.75 & \cellcolor{highlight!56}75.83 & \cellcolor{highlight!80}0.084 & \cellcolor{highlight!56}79.39 & \cellcolor{highlight!80}0.066 & \cellcolor{highlight!80}78.89 & \cellcolor{highlight!80}0.077 & \cellcolor{highlight!80}82.84 & \cellcolor{highlight!80}0.060 \\
    \textbf{Ours} \scriptsize(VGGT-$\Omega$) & \cellcolor{highlight!10}87.80 & \cellcolor{highlight!80}76.14 & \cellcolor{highlight!56}0.098 & \cellcolor{highlight!80}79.82 & \cellcolor{highlight!38}0.076 & \cellcolor{highlight!56}77.96 & \cellcolor{highlight!56}0.095 & \cellcolor{highlight!56}82.29 & \cellcolor{highlight!56}0.075 \\
    \hline
\end{tabularx}

    \vspace{0.3em}
    \setlength{\tabcolsep}{0pt}
    \begin{tabularx}{\textwidth}{@{}l*{13}{Y}@{}}
    \multicolumn{14}{c}{\textbf{(b) Multiview reconstruction}} \\
    \hline
    \multirow[c]{3}{*}{Method}
    & \multicolumn{4}{c}{Depth}
    & \multicolumn{3}{c}{Camera}
    & \multicolumn{3}{c}{Matching}
    & \multicolumn{3}{c}{Global points} \\
    \cmidrule(lr){2-5}\cmidrule(lr){6-8}\cmidrule(lr){9-11}\cmidrule(lr){12-14}
    & \multicolumn{2}{c}{Affine} & \multicolumn{2}{c}{Scale}
    & \multirow{2}{*}{{\scriptsize ATE$\downarrow$}}
    & \multirow{2}{*}{{\scriptsize RPE-R$\downarrow$}}
    & \multirow{2}{*}{{\scriptsize RPE-t$\downarrow$}}
    & \multirow{2}{*}{{\scriptsize $\mathrm{AUC}_{5}\uparrow$}}
    & \multirow{2}{*}{{\scriptsize $\delta_{5}\uparrow$}}
    & \multirow{2}{*}{{\scriptsize RD$\downarrow$}}
    & \multirow{2}{*}{{\scriptsize $\mathrm{AUC}_{25}\uparrow$}}
    & \multirow{2}{*}{{\scriptsize $\delta_{25}\uparrow$}}
    & \multirow{2}{*}{{\scriptsize RD$\downarrow$}} \\
    \cmidrule(lr){2-3}\cmidrule(lr){4-5}
    & $\delta_{25}\uparrow$ & Rel$\downarrow$
    & $\delta_{25}\uparrow$ & Rel$\downarrow$
    & & & & & & & & & \\
    \hline
    MonST3R & \cellcolor{highlight!0}86.12 & \cellcolor{highlight!0}0.124 & \cellcolor{highlight!0}82.42 & \cellcolor{highlight!0}0.144 & \cellcolor{highlight!0}0.144 & \cellcolor{highlight!0}14.89 & \cellcolor{highlight!0}0.315 & \cellcolor{highlight!0}33.61 & \cellcolor{highlight!0}56.48 & \cellcolor{highlight!0}0.125 & \cellcolor{highlight!0}50.92 & \cellcolor{highlight!0}76.55 & \cellcolor{highlight!0}0.217 \\
    MegaSAM & \cellcolor{highlight!0}85.83 & \cellcolor{highlight!0}0.122 & \cellcolor{highlight!0}82.03 & \cellcolor{highlight!0}0.142 & \cellcolor{highlight!0}0.125 & \cellcolor{highlight!0}18.33 & \cellcolor{highlight!0}0.221 & \cellcolor{highlight!2}53.30 & \cellcolor{highlight!1}73.95 & \cellcolor{highlight!0}0.159 & \cellcolor{highlight!0}51.84 & \cellcolor{highlight!0}72.06 & \cellcolor{highlight!0}0.275 \\
    VGGT & \cellcolor{highlight!4}95.24 & \cellcolor{highlight!10}0.060 & \cellcolor{highlight!7}94.13 & \cellcolor{highlight!7}0.068 & \cellcolor{highlight!10}0.026 & \cellcolor{highlight!43}1.609 & \cellcolor{highlight!10}0.056 & \cellcolor{highlight!14}63.15 & \cellcolor{highlight!7}88.43 & \cellcolor{highlight!20}0.029 & \cellcolor{highlight!26}77.44 & \cellcolor{highlight!14}94.14 & \cellcolor{highlight!26}0.073 \\
    MapAnything & \cellcolor{highlight!1}93.36 & \cellcolor{highlight!1}0.083 & \cellcolor{highlight!1}91.92 & \cellcolor{highlight!1}0.092 & \cellcolor{highlight!2}0.034 & \cellcolor{highlight!1}2.464 & \cellcolor{highlight!2}0.072 & \cellcolor{highlight!4}54.96 & \cellcolor{highlight!4}85.65 & \cellcolor{highlight!2}0.043 & \cellcolor{highlight!1}72.56 & \cellcolor{highlight!1}91.76 & \cellcolor{highlight!1}0.106 \\
    $\pi^3$ & \cellcolor{highlight!43}96.64 & \cellcolor{highlight!43}0.044 & \cellcolor{highlight!26}94.99 & \cellcolor{highlight!43}0.055 & \cellcolor{highlight!26}0.019 & \cellcolor{highlight!54}1.469 & \cellcolor{highlight!34}0.042 & \cellcolor{highlight!66}69.57 & \cellcolor{highlight!80}94.44 & \cellcolor{highlight!66}0.022 & \cellcolor{highlight!54}81.09 & \cellcolor{highlight!43}95.70 & \cellcolor{highlight!43}0.058 \\
    Depth Anything 3 & \cellcolor{highlight!20}95.48 & \cellcolor{highlight!20}0.055 & \cellcolor{highlight!14}94.31 & \cellcolor{highlight!20}0.062 & \cellcolor{highlight!54}0.016 & \cellcolor{highlight!34}1.651 & \cellcolor{highlight!54}0.041 & \cellcolor{highlight!20}64.03 & \cellcolor{highlight!26}90.37 & \cellcolor{highlight!10}0.029 & \cellcolor{highlight!7}74.00 & \cellcolor{highlight!7}93.42 & \cellcolor{highlight!4}0.083 \\
    LingBot-Map & \cellcolor{highlight!1}94.58 & \cellcolor{highlight!2}0.071 & \cellcolor{highlight!1}93.12 & \cellcolor{highlight!2}0.078 & \cellcolor{highlight!4}0.029 & \cellcolor{highlight!10}1.794 & \cellcolor{highlight!4}0.060 & \cellcolor{highlight!1}46.88 & \cellcolor{highlight!1}79.97 & \cellcolor{highlight!1}0.043 & \cellcolor{highlight!4}73.82 & \cellcolor{highlight!2}93.22 & \cellcolor{highlight!2}0.087 \\
    VGGT-$\Omega$ & \cellcolor{highlight!80}98.46 & \cellcolor{highlight!66}0.037 & \cellcolor{highlight!80}98.07 & \cellcolor{highlight!80}0.040 & \cellcolor{highlight!80}0.012 & \cellcolor{highlight!80}1.068 & \cellcolor{highlight!80}0.030 & \cellcolor{highlight!54}69.04 & \cellcolor{highlight!66}93.93 & \cellcolor{highlight!54}0.022 & \cellcolor{highlight!80}85.81 & \cellcolor{highlight!80}98.06 & \cellcolor{highlight!66}0.048 \\
    $\pi^3$ \scriptsize(DINO, our data) & \cellcolor{highlight!26}96.19 & \cellcolor{highlight!14}0.056 & \cellcolor{highlight!43}95.32 & \cellcolor{highlight!14}0.063 & \cellcolor{highlight!1}0.040 & \cellcolor{highlight!2}2.388 & \cellcolor{highlight!1}0.076 & \cellcolor{highlight!1}49.55 & \cellcolor{highlight!2}80.67 & \cellcolor{highlight!4}0.041 & \cellcolor{highlight!2}73.41 & \cellcolor{highlight!20}94.27 & \cellcolor{highlight!14}0.079 \\
    \hline
    Any4D & \cellcolor{highlight!2}94.58 & \cellcolor{highlight!1}0.076 & \cellcolor{highlight!2}93.45 & \cellcolor{highlight!1}0.085 & \cellcolor{highlight!1}0.075 & \cellcolor{highlight!1}3.498 & \cellcolor{highlight!1}0.155 & \cellcolor{highlight!0}40.97 & \cellcolor{highlight!0}72.24 & \cellcolor{highlight!1}0.056 & \cellcolor{highlight!1}68.47 & \cellcolor{highlight!1}91.63 & \cellcolor{highlight!1}0.114 \\
    Open-D4RT & \cellcolor{highlight!0}91.18 & \cellcolor{highlight!0}0.099 & \cellcolor{highlight!0}88.15 & \cellcolor{highlight!0}0.116 & \cellcolor{highlight!0}0.110 & \cellcolor{highlight!0}11.94 & \cellcolor{highlight!0}0.214 & \cellcolor{highlight!0}37.36 & \cellcolor{highlight!0}65.48 & \cellcolor{highlight!0}0.126 & \cellcolor{highlight!0}56.27 & \cellcolor{highlight!0}80.68 & \cellcolor{highlight!0}0.209 \\
    V-DPM & \cellcolor{highlight!10}95.39 & \cellcolor{highlight!4}0.061 & \cellcolor{highlight!4}93.77 & \cellcolor{highlight!4}0.071 & \cellcolor{highlight!7}0.027 & \cellcolor{highlight!7}1.865 & \cellcolor{highlight!7}0.058 & \cellcolor{highlight!10}62.96 & \cellcolor{highlight!10}88.92 & \cellcolor{highlight!14}0.029 & \cellcolor{highlight!20}77.24 & \cellcolor{highlight!10}94.11 & \cellcolor{highlight!20}0.075 \\
    Track4World & \cellcolor{highlight!7}95.35 & \cellcolor{highlight!7}0.061 & \cellcolor{highlight!10}94.19 & \cellcolor{highlight!10}0.068 & \cellcolor{highlight!34}0.017 & \cellcolor{highlight!14}1.775 & \cellcolor{highlight!43}0.042 & \cellcolor{highlight!7}61.45 & \cellcolor{highlight!14}89.15 & \cellcolor{highlight!7}0.031 & \cellcolor{highlight!14}76.18 & \cellcolor{highlight!26}94.43 & \cellcolor{highlight!10}0.080 \\
    4RC & \cellcolor{highlight!14}95.44 & \cellcolor{highlight!26}0.054 & \cellcolor{highlight!20}94.69 & \cellcolor{highlight!26}0.060 & \cellcolor{highlight!43}0.017 & \cellcolor{highlight!26}1.670 & \cellcolor{highlight!26}0.043 & \cellcolor{highlight!34}65.43 & \cellcolor{highlight!20}90.21 & \cellcolor{highlight!26}0.029 & \cellcolor{highlight!10}75.31 & \cellcolor{highlight!4}93.24 & \cellcolor{highlight!7}0.083 \\
    \hline
    \textbf{Ours} \scriptsize(DINO) & \cellcolor{highlight!34}96.49 & \cellcolor{highlight!34}0.049 & \cellcolor{highlight!34}95.02 & \cellcolor{highlight!34}0.060 & \cellcolor{highlight!20}0.024 & \cellcolor{highlight!20}1.720 & \cellcolor{highlight!14}0.052 & \cellcolor{highlight!43}66.73 & \cellcolor{highlight!43}93.32 & \cellcolor{highlight!43}0.023 & \cellcolor{highlight!34}79.26 & \cellcolor{highlight!34}95.59 & \cellcolor{highlight!34}0.063 \\
    \textbf{Ours} \scriptsize(VJEPA) & \cellcolor{highlight!54}97.40 & \cellcolor{highlight!54}0.042 & \cellcolor{highlight!54}96.25 & \cellcolor{highlight!54}0.051 & \cellcolor{highlight!14}0.024 & \cellcolor{highlight!4}2.031 & \cellcolor{highlight!20}0.051 & \cellcolor{highlight!26}65.06 & \cellcolor{highlight!34}91.38 & \cellcolor{highlight!34}0.025 & \cellcolor{highlight!43}80.86 & \cellcolor{highlight!54}96.33 & \cellcolor{highlight!54}0.057 \\
    \textbf{Ours} \scriptsize(VGGT-$\Omega$) & \cellcolor{highlight!66}97.91 & \cellcolor{highlight!80}0.035 & \cellcolor{highlight!66}96.82 & \cellcolor{highlight!66}0.042 & \cellcolor{highlight!66}0.015 & \cellcolor{highlight!66}1.379 & \cellcolor{highlight!66}0.035 & \cellcolor{highlight!80}70.93 & \cellcolor{highlight!54}93.58 & \cellcolor{highlight!80}0.021 & \cellcolor{highlight!66}84.13 & \cellcolor{highlight!66}97.34 & \cellcolor{highlight!80}0.045 \\
    \hline
\end{tabularx}

\end{table}

\paragraph{Setup and baselines.}
We evaluate 32-frame clips (5\,fps) seeded from the first frame across six synthetic and real-world benchmarks: DynamicReplica~\citep{dynamicstereo}, PointOdyssey~\citep{pointodyssey}, Syn4D~\citep{syn4d}, Aria Digital Twin (ADT)~\citep{pan2023adt}, Panoptic Studio (PStudio)~\citep{joo2015panoptic}, and the held-out test split of RigIDTrack.
Trajectories are evaluated in both camera and world space (after sequence-level scale and $\mathrm{Sim}(3)$ alignment, respectively) over \emph{all} valid points and the \emph{visible} subset, using relative end-point error (RelEPE, normalized by ground-truth camera distance), $\mathrm{AUC}_{25}$, and occlusion accuracy (OA).
Baselines include St4RTrack~\citep{st4rtrack}, Open-D4RT\footnote{Unofficial open-source reproduction of D4RT~\citep{d4rt}: \url{https://github.com/Lijiaxin0111/Open-d4rt}.}, Any4D~\citep{any4d}, Trace Anything~\citep{traceanything}, V-DPM~\citep{vdpm}, Track4World~\citep{track4world}, and 4RC~\citep{4rc}.

\paragraph{Results.}
All three MoGe-4D variants outperform prior methods across every averaged tracking metric (Table~\ref{tab:tracking_3d}(a)).
The V-JEPA 2.1 variant performs best overall, reflecting the benefit of spatiotemporal video pretraining for motion prediction.
Compared with the strongest tracking baselines (V-DPM and 4RC), it reduces all-point RelEPE by 47.5\% in camera space (0.084 vs.\ 0.160) and 44.2\% in world space (0.077 vs.\ 0.138), while raising world-space AUC from 72.65\% to 78.89\%.

\subsection{Multiview Reconstruction}
\label{sec:multiview_reconstruction}

\begin{figure*}[!t]
    \captionsetup{hypcap=false}
    \centering
    \includegraphics[width=\textwidth]{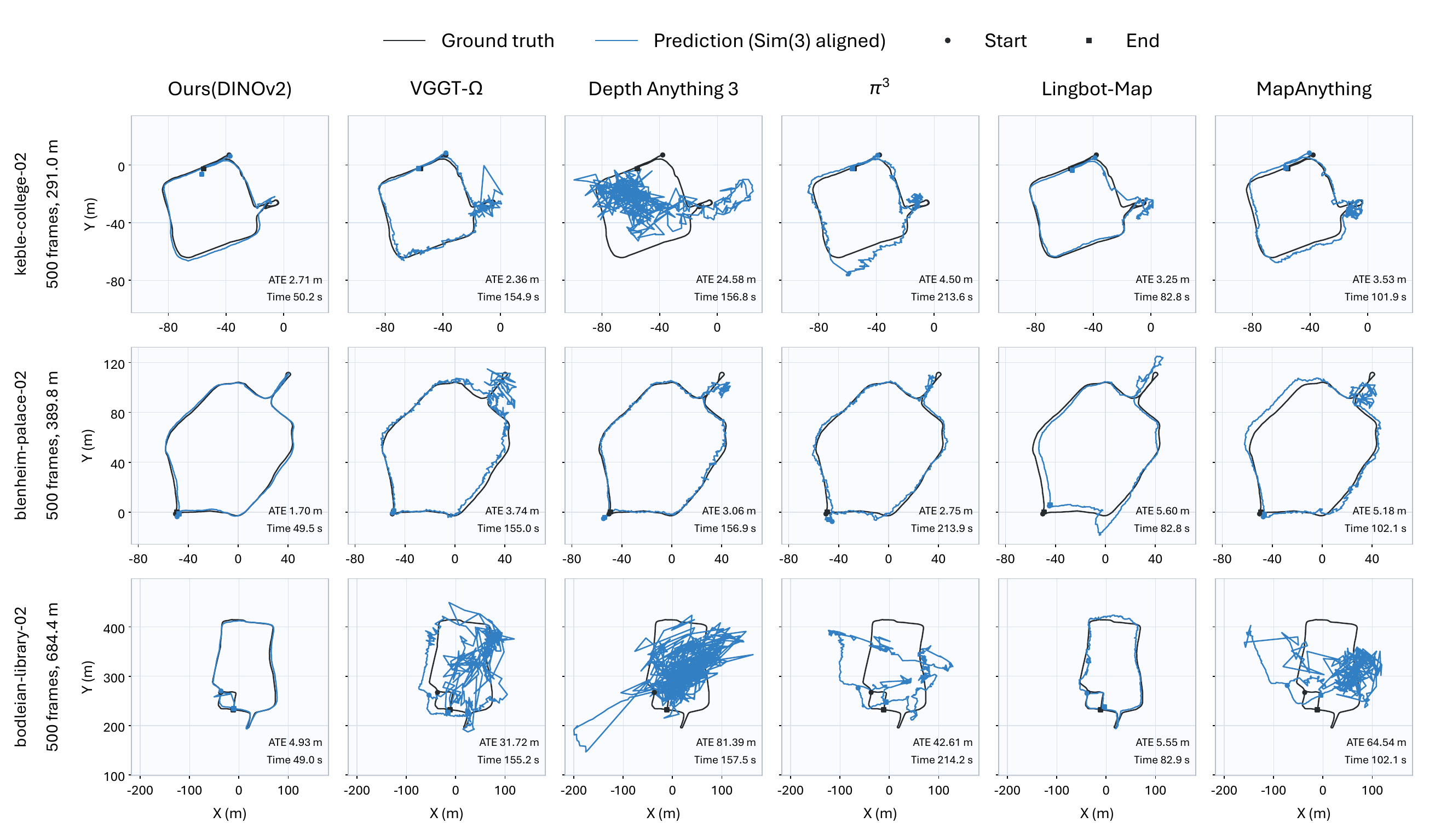}
    \vspace{-2em}
    \captionof{figure}{\textbf{Long-horizon feed-forward camera trajectory comparisons on Oxford-Spires.}
    Each method processes 500 frames in a single feed-forward pass without sliding windows.
    Trajectories are shown on the horizontal ($XY$) plane after $\mathrm{Sim}(3)$ alignment to ground truth, annotated with absolute trajectory error (ATE) and total inference time.}
    \label{fig:trajectories_selected}
\end{figure*}

\begin{figure*}[!t]
    \centering
    \includegraphics[width=\textwidth,height=0.86\textheight,keepaspectratio]{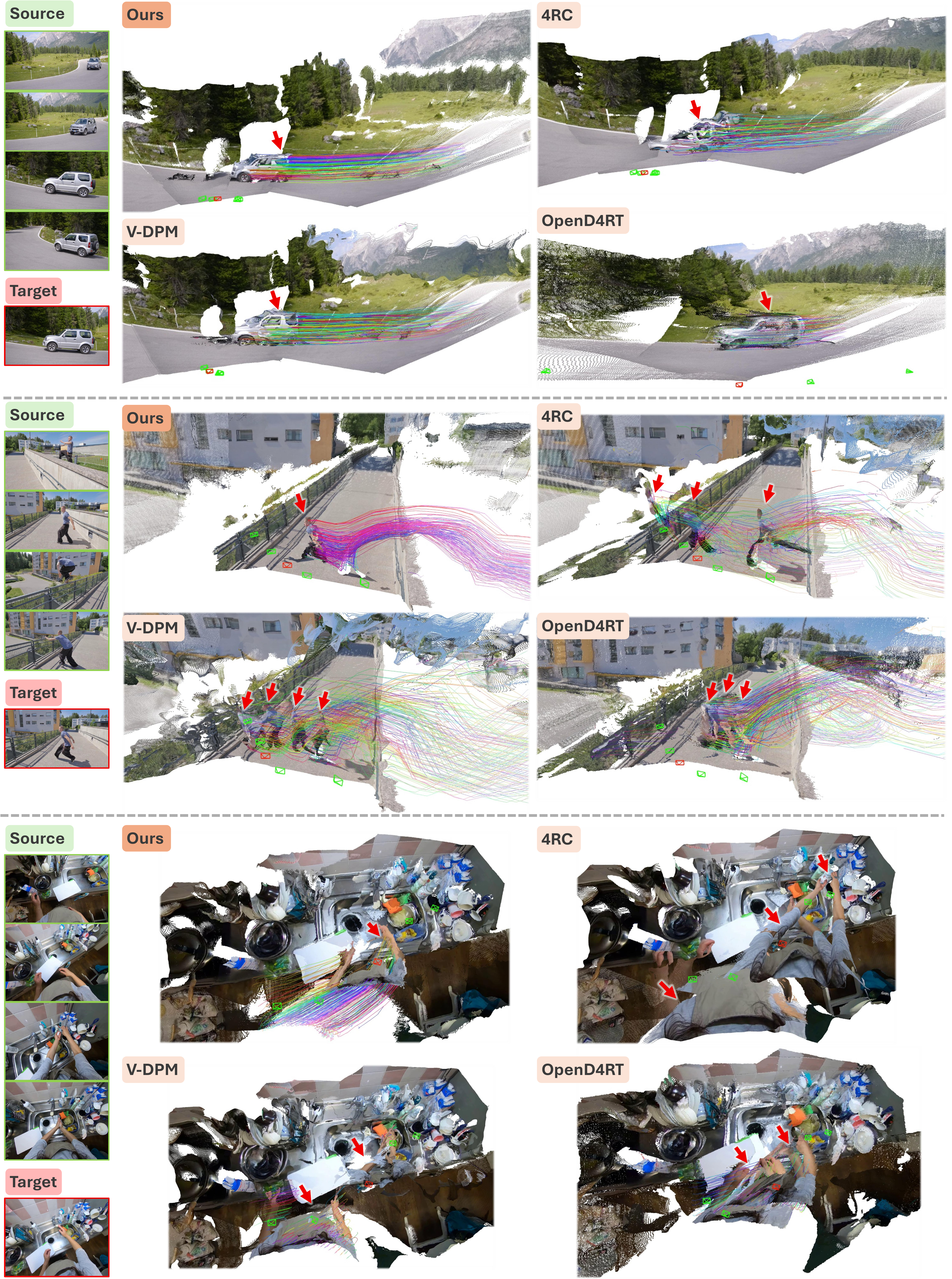}
    \caption{\textbf{Qualitative comparison of unified 4D reconstruction and tracking.}
    For each sequence, four source frames (green) are reconstructed at the timestamp of a target frame (red), accompanied by 3D point trajectories.
    Compared with 4RC, V-DPM, and Open-D4RT, MoGe-4D recovers cleaner scene geometry while tracking dynamic foreground motion accurately across indoor and outdoor scenes (see Appendix~\ref{sec:supp_qualitative_comparisons} for additional examples).}
    \label{fig:qualitative_results}
\end{figure*}

\newcommand{\ablationresultblock}{\begingroup
\fontsize{6.5pt}{7.8pt}\selectfont
\setlength{\aboverulesep}{0pt}
\setlength{\belowrulesep}{0pt}
\renewcommand{\arraystretch}{1.1}
\setlength{\tabcolsep}{1.2pt}
\resizebox{\textwidth}{!}{%
\begin{tabular}{@{}l*{22}{c}@{}}
    \hline
    \multirow[c]{4}{*}{\fontsize{5.8pt}{7pt}\selectfont Variant}
    & \multicolumn{13}{c}{\fontsize{5.8pt}{7pt}\selectfont\textbf{Multiview reconstruction}}
    & \multicolumn{9}{c}{\fontsize{5.8pt}{7pt}\selectfont\textbf{3D tracking}} \\
    \cmidrule(lr){2-14}\cmidrule(lr){15-23}
    & \multicolumn{4}{c}{\fontsize{5.8pt}{7pt}\selectfont Depth}
    & \multicolumn{3}{c}{\fontsize{5.8pt}{7pt}\selectfont Camera}
    & \multicolumn{3}{c}{\fontsize{5.8pt}{7pt}\selectfont Matching}
    & \multicolumn{3}{c}{\fontsize{5.8pt}{7pt}\selectfont Global points}
    & \multirow{3}{*}{\fontsize{5.8pt}{7pt}\selectfont OA$\uparrow$}
    & \multicolumn{4}{c}{\fontsize{5.8pt}{7pt}\selectfont Camera space}
    & \multicolumn{4}{c}{\fontsize{5.8pt}{7pt}\selectfont World space} \\
    \cmidrule(lr){2-5}\cmidrule(lr){6-8}\cmidrule(lr){9-11}\cmidrule(lr){12-14}\cmidrule(lr){16-19}\cmidrule(lr){20-23}
    & \multicolumn{2}{c}{\fontsize{5.8pt}{7pt}\selectfont Affine} & \multicolumn{2}{c}{\fontsize{5.8pt}{7pt}\selectfont Scale}
    & \multirow{2}{*}{\fontsize{5.8pt}{7pt}\selectfont ATE$\downarrow$}
    & \multirow{2}{*}{\fontsize{5.8pt}{7pt}\selectfont RPE-R$\downarrow$}
    & \multirow{2}{*}{\fontsize{5.8pt}{7pt}\selectfont RPE-t$\downarrow$}
    & \multirow{2}{*}{\fontsize{5.8pt}{7pt}\selectfont $\mathrm{AUC}_{5}\uparrow$}
    & \multirow{2}{*}{\fontsize{5.8pt}{7pt}\selectfont $\delta_{5}\uparrow$}
    & \multirow{2}{*}{\fontsize{5.8pt}{7pt}\selectfont RD$\downarrow$}
    & \multirow{2}{*}{\fontsize{5.8pt}{7pt}\selectfont $\mathrm{AUC}_{25}\uparrow$}
    & \multirow{2}{*}{\fontsize{5.8pt}{7pt}\selectfont $\delta_{25}\uparrow$}
    & \multirow{2}{*}{\fontsize{5.8pt}{7pt}\selectfont RD$\downarrow$}
    & & \multicolumn{2}{c}{\fontsize{5.8pt}{7pt}\selectfont All} & \multicolumn{2}{c}{\fontsize{5.8pt}{7pt}\selectfont Visible}
    & \multicolumn{2}{c}{\fontsize{5.8pt}{7pt}\selectfont All} & \multicolumn{2}{c}{\fontsize{5.8pt}{7pt}\selectfont Visible} \\
    \cmidrule(lr){2-3}\cmidrule(lr){4-5}\cmidrule(lr){16-17}\cmidrule(lr){18-19}\cmidrule(lr){20-21}\cmidrule(lr){22-23}
    & \fontsize{5.8pt}{7pt}\selectfont $\delta_{25}\uparrow$ & \fontsize{5.8pt}{7pt}\selectfont Rel$\downarrow$
    & \fontsize{5.8pt}{7pt}\selectfont $\delta_{25}\uparrow$ & \fontsize{5.8pt}{7pt}\selectfont Rel$\downarrow$
    & & & & & & & & & &
    & \fontsize{5.8pt}{7pt}\selectfont AUC$\uparrow$ & \fontsize{5.8pt}{7pt}\selectfont RelEPE$\downarrow$ & \fontsize{5.8pt}{7pt}\selectfont AUC$\uparrow$ & \fontsize{5.8pt}{7pt}\selectfont RelEPE$\downarrow$
    & \fontsize{5.8pt}{7pt}\selectfont AUC$\uparrow$ & \fontsize{5.8pt}{7pt}\selectfont RelEPE$\downarrow$ & \fontsize{5.8pt}{7pt}\selectfont AUC$\uparrow$ & \fontsize{5.8pt}{7pt}\selectfont RelEPE$\downarrow$ \\
    \hline

    \textbf{full model} & \cellcolor{highlight!0}95.36 & \cellcolor{highlight!80}0.066 & \cellcolor{highlight!80}94.34 & \cellcolor{highlight!80}0.076 & \cellcolor{highlight!80}0.057 & \cellcolor{highlight!80}2.965 & \cellcolor{highlight!0}0.127 & \cellcolor{highlight!80}52.96 & \cellcolor{highlight!80}83.58 & \cellcolor{highlight!80}0.037 & \cellcolor{highlight!80}71.17 & \cellcolor{highlight!80}92.84 & \cellcolor{highlight!80}0.103 & \cellcolor{highlight!80}86.98 & \cellcolor{highlight!80}62.51 & \cellcolor{highlight!0}0.144 & \cellcolor{highlight!80}67.70 & \cellcolor{highlight!0}0.106 & \cellcolor{highlight!80}68.12 & \cellcolor{highlight!0}0.126 & \cellcolor{highlight!80}74.66 & \cellcolor{highlight!0}0.093 \\
    w/o bridge pretrain & \cellcolor{highlight!0}94.58 & \cellcolor{highlight!0}0.073 & \cellcolor{highlight!0}93.39 & \cellcolor{highlight!0}0.084 & \cellcolor{highlight!0}0.079 & \cellcolor{highlight!0}9.651 & \cellcolor{highlight!0}0.167 & \cellcolor{highlight!0}43.89 & \cellcolor{highlight!0}73.86 & \cellcolor{highlight!0}0.059 & \cellcolor{highlight!0}62.65 & \cellcolor{highlight!0}86.50 & \cellcolor{highlight!0}0.139 & \cellcolor{highlight!0}85.78 & \cellcolor{highlight!0}60.27 & \cellcolor{highlight!0}0.153 & \cellcolor{highlight!0}65.02 & \cellcolor{highlight!0}0.118 & \cellcolor{highlight!0}66.64 & \cellcolor{highlight!0}0.130 & \cellcolor{highlight!0}72.33 & \cellcolor{highlight!0}0.102 \\
    w/o self KV & \cellcolor{highlight!80}95.41 & \cellcolor{highlight!0}0.068 & \cellcolor{highlight!0}94.29 & \cellcolor{highlight!0}0.076 & \cellcolor{highlight!0}0.062 & \cellcolor{highlight!0}2.997 & \cellcolor{highlight!80}0.126 & \cellcolor{highlight!0}52.14 & \cellcolor{highlight!0}82.22 & \cellcolor{highlight!0}0.039 & \cellcolor{highlight!0}69.28 & \cellcolor{highlight!0}92.26 & \cellcolor{highlight!0}0.107 & \cellcolor{highlight!0}86.88 & \cellcolor{highlight!0}62.46 & \cellcolor{highlight!80}0.143 & \cellcolor{highlight!0}67.64 & \cellcolor{highlight!80}0.106 & \cellcolor{highlight!0}68.07 & \cellcolor{highlight!80}0.125 & \cellcolor{highlight!0}74.58 & \cellcolor{highlight!80}0.092 \\
    camera token & \cellcolor{highlight!0}95.35 & \cellcolor{highlight!0}0.068 & \cellcolor{highlight!0}94.04 & \cellcolor{highlight!0}0.078 & \cellcolor{highlight!0}0.125 & \cellcolor{highlight!0}9.803 & \cellcolor{highlight!0}0.221 & \cellcolor{highlight!0}21.24 & \cellcolor{highlight!0}45.19 & \cellcolor{highlight!0}0.125 & \cellcolor{highlight!0}54.11 & \cellcolor{highlight!0}81.38 & \cellcolor{highlight!0}0.180 & - & - & - & - & - & - & - & - & - \\
    \hline
\end{tabular}%
}
\endgroup
}

\paragraph{Setup and baselines.}
We evaluate up to 32 unposed frames per clip (subsampled to keep ${\ge}50\%$ adjacent-view overlap) across six benchmarks: 7-Scenes~\citep{shotton2013sevenscenes}, Bonn~\citep{palazzolo2019refusion}, ETH3D~\citep{schops2017eth3d}, KITTI~\citep{geiger2012kitti}, Sintel~\citep{butler2012sintel}, and ScanNet~\citep{dai2017scannet}.
We measure scale-/affine-aligned depth, camera pose error (ATE, RPE), and correspondence matching distance.
Following VGGT-$\Omega$~\citep{vggtomega}, we emphasize global-point accuracy (RD, $\delta_{25}$, and $\mathrm{AUC}_{25}$ normalized by ground-truth camera distance after sequence-wide $\mathrm{Sim}(3)$ alignment) as a joint measure of depth, intrinsic, and pose quality.
We compare against dedicated 3D models (MonST3R~\citep{monst3r}, MegaSaM~\citep{megasam}, VGGT~\citep{vggt}, MapAnything~\citep{mapanything}, $\pi^3$~\citep{pi3}, Depth Anything 3 (DA3)~\citep{depthanything3}, LingBot-Map~\citep{lingbotmap}, and VGGT-$\Omega$~\citep{vggtomega}), and the unified 4D models from Sec.~\ref{sec:tracking_3d} (Any4D, Open-D4RT, V-DPM, 4RC, Track4World).
Additionally, we evaluate $\pi^3$ trained from DINOv2 initialization under our exact data, batch size, and step budget.

\paragraph{Results.}
In Table~\ref{tab:multiview_reconstruction}(b), all three MoGe-4D variants outperform prior unified 4D methods on depth, matching, and global-point metrics while matching strong dedicated 3D models.
Among our variants, VGGT-$\Omega$ initialization achieves the strongest geometry: it beats all unified 4D baselines on every metric, attains the best overall global-point RD and matching $\mathrm{AUC}_{5}$, and ranks second only to VGGT-$\Omega$ on global-point AUC and $\delta_{25}$.
The V-JEPA 2.1 variant also matches the original $\pi^3$ on global points (80.86\% vs.\ 81.09\% AUC; 0.057 vs.\ 0.058 RD).
Across individual benchmarks, it improves over our DINOv2 variant on video sequences but trails on the unordered ETH3D sets, consistent with its video pretraining and temporal RoPE~\citep{su2024roformer}.
In the controlled DINOv2 comparison, MoGe-4D outperforms $\pi^3$ on 12 of 13 metrics, improving global-point AUC from 73.41\% to 79.26\% and RD from 0.079 to 0.063.

\begin{wraptable}{r}{0.48\textwidth}
    \vspace{-1.2em}
    \setlength{\textwidth}{\linewidth}
    \captionsetup{aboveskip=0pt,belowskip=4pt}
    \caption{\textbf{Metric-scale reconstruction comparison.}
    Averaged across five real-world benchmarks (KITTI, Bonn, 7-Scenes, ETH3D, ScanNet) without scale alignment.}
    \label{tab:metric_scale}
    \fontsize{8pt}{9.6pt}\selectfont
    \setlength{\aboverulesep}{0pt}
    \setlength{\belowrulesep}{0pt}
    \renewcommand{\arraystretch}{1.1}
    \setlength{\tabcolsep}{3pt}
\begin{tabularx}{\textwidth}{@{}l*{4}{Y}@{}}
    \hline
    \multirow[c]{2}{*}{Method}
    & \multicolumn{1}{c}{Scale Factor}
    & \multicolumn{3}{c}{Depth} \\
    \cmidrule(lr){2-2}\cmidrule(lr){3-5}
    & $|\log s|\downarrow$
    & Rel$\downarrow$
    & $\delta_{25}\uparrow$
    & $\mathrm{AUC}_{25}\uparrow$ \\
    \hline
    MapAnything & \cellcolor{highlight!0}0.177 & \cellcolor{highlight!0}0.262 & \cellcolor{highlight!0}61.79 & \cellcolor{highlight!0}32.20 \\
    Depth Anything 3 & \cellcolor{highlight!46}0.072 & \cellcolor{highlight!10}0.103 & \cellcolor{highlight!3}89.34 & \cellcolor{highlight!10}64.20 \\
    Any4D & \cellcolor{highlight!0}0.232 & \cellcolor{highlight!0}0.382 & \cellcolor{highlight!0}29.69 & \cellcolor{highlight!0}17.65 \\
    Track4World & \cellcolor{highlight!3}0.101 & \cellcolor{highlight!3}0.105 & \cellcolor{highlight!10}91.25 & \cellcolor{highlight!3}62.50 \\
    \hline
    \textbf{Ours} \scriptsize(DINO) & \cellcolor{highlight!80}0.060 & \cellcolor{highlight!80}0.079 & \cellcolor{highlight!80}97.00 & \cellcolor{highlight!46}69.78 \\
    \textbf{Ours} \scriptsize(V-JEPA) & \cellcolor{highlight!24}0.073 & \cellcolor{highlight!46}0.080 & \cellcolor{highlight!46}95.33 & \cellcolor{highlight!80}70.77 \\
    \textbf{Ours} \scriptsize(VGGT-$\Omega$) & \cellcolor{highlight!10}0.077 & \cellcolor{highlight!24}0.090 & \cellcolor{highlight!24}94.77 & \cellcolor{highlight!24}66.02 \\
    \hline
\end{tabularx}

    \vspace{-1.0em}
\end{wraptable}
\paragraph{Metric-scale reconstruction.}
We also evaluate unaligned metric depth and scene scale error ($|\log s|$, with $\mathrm{Sim}(3)$ scale factor $s$) across the five real-world reconstruction benchmarks (excluding synthetic Sintel) against metric-scale baselines (MapAnything, DA3, Any4D, and Track4World).
In Table~\ref{tab:metric_scale}, all three MoGe-4D variants outperform prior methods across all unaligned depth metrics (Rel, $\delta_{25}$, and $\mathrm{AUC}_{25}$).
Our DINOv2 variant achieves the best overall scale and depth accuracy ($|\log s|$ of 0.060 vs.\ 0.072 for DA3; 0.079 Rel and 97.00\% $\delta_{25}$), while V-JEPA 2.1 reaches the highest $\mathrm{AUC}_{25}$ (70.77\%).

\begin{figure*}[!t]
    \captionsetup{hypcap=false}
    \centering
    \begin{minipage}[t]{0.765\textwidth}
        \vspace{0pt}
        \centering
        \begin{minipage}[t]{0.262\linewidth}
            \centering
            \includegraphics[width=\linewidth]{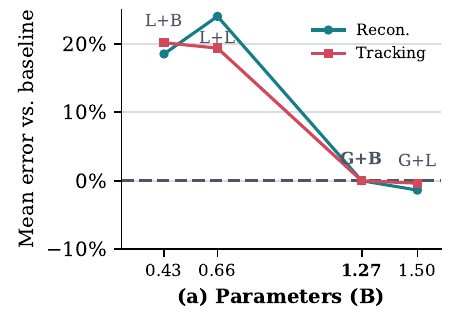}
        \end{minipage}\hfill
        \begin{minipage}[t]{0.238\linewidth}
            \centering
            \includegraphics[width=\linewidth]{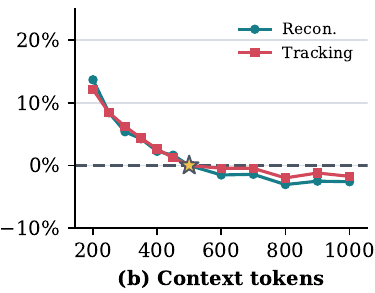}
        \end{minipage}\hfill
        \begin{minipage}[t]{0.238\linewidth}
            \centering
            \includegraphics[width=\linewidth]{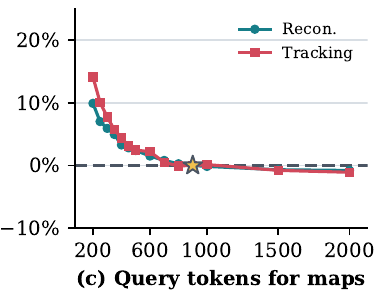}
        \end{minipage}\hfill
        \begin{minipage}[t]{0.238\linewidth}
            \centering
            \includegraphics[width=\linewidth]{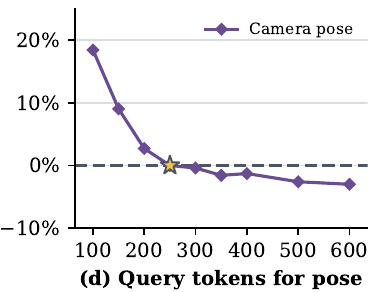}
        \end{minipage}
        \vspace{-0.4em}
        \captionof{figure}{\textbf{Scaling with model capacity (a) and inference token budgets (b--d).}
        Curves show mean relative error changes from the default settings (bold ticks and stars; lower is better).}
        \label{fig:ablation_tradeoffs}
    \end{minipage}\hfill
    \begin{minipage}[t]{0.215\textwidth}
        \vspace{0pt}
        \centering
        \includegraphics[width=\linewidth]{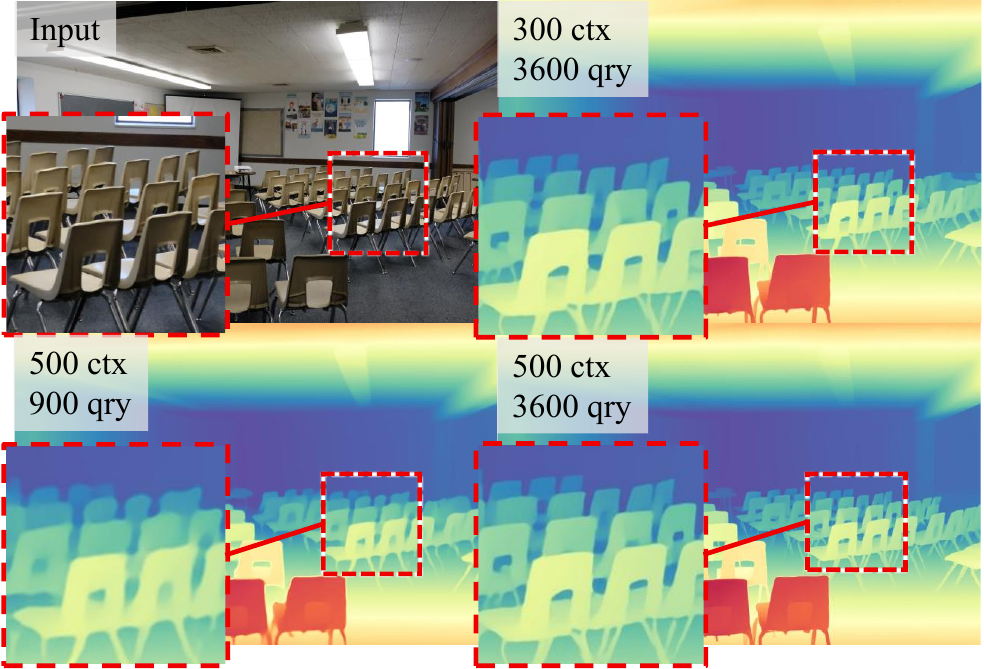}
        \vspace{-1.79em}
        \captionof{figure}{\textbf{Depth detail across query-token budgets.}}
        \label{fig:tokens_for_depth}
    \end{minipage}

    \vspace{0.6em}
    \begin{minipage}{\linewidth}
    \captionsetup{aboveskip=4pt,belowskip=0pt}
    \captionof{table}{\textbf{Ablations of bridge alignment, query self-attention, and camera pose recovery.}
    All variants use DINOv2-initialized ViT-Large context and ViT-Base query encoders; metrics follow Table~\ref{tab:multiview_reconstruction}.}
    \label{tab:camera_recovery_ablation}
    \vspace{2pt}
    \begingroup
\fontsize{6.5pt}{7.8pt}\selectfont
\setlength{\aboverulesep}{0pt}
\setlength{\belowrulesep}{0pt}
\renewcommand{\arraystretch}{1.1}
\setlength{\tabcolsep}{1.2pt}
\resizebox{\textwidth}{!}{%
\begin{tabular}{@{}l*{22}{c}@{}}
    \hline
    \multirow[c]{4}{*}{\fontsize{5.8pt}{7pt}\selectfont Variant}
    & \multicolumn{13}{c}{\fontsize{5.8pt}{7pt}\selectfont\textbf{Multiview reconstruction}}
    & \multicolumn{9}{c}{\fontsize{5.8pt}{7pt}\selectfont\textbf{3D tracking}} \\
    \cmidrule(lr){2-14}\cmidrule(lr){15-23}
    & \multicolumn{4}{c}{\fontsize{5.8pt}{7pt}\selectfont Depth}
    & \multicolumn{3}{c}{\fontsize{5.8pt}{7pt}\selectfont Camera}
    & \multicolumn{3}{c}{\fontsize{5.8pt}{7pt}\selectfont Matching}
    & \multicolumn{3}{c}{\fontsize{5.8pt}{7pt}\selectfont Global points}
    & \multirow{3}{*}{\fontsize{5.8pt}{7pt}\selectfont OA$\uparrow$}
    & \multicolumn{4}{c}{\fontsize{5.8pt}{7pt}\selectfont Camera space}
    & \multicolumn{4}{c}{\fontsize{5.8pt}{7pt}\selectfont World space} \\
    \cmidrule(lr){2-5}\cmidrule(lr){6-8}\cmidrule(lr){9-11}\cmidrule(lr){12-14}\cmidrule(lr){16-19}\cmidrule(lr){20-23}
    & \multicolumn{2}{c}{\fontsize{5.8pt}{7pt}\selectfont Affine} & \multicolumn{2}{c}{\fontsize{5.8pt}{7pt}\selectfont Scale}
    & \multirow{2}{*}{\fontsize{5.8pt}{7pt}\selectfont ATE$\downarrow$}
    & \multirow{2}{*}{\fontsize{5.8pt}{7pt}\selectfont RPE-R$\downarrow$}
    & \multirow{2}{*}{\fontsize{5.8pt}{7pt}\selectfont RPE-t$\downarrow$}
    & \multirow{2}{*}{\fontsize{5.8pt}{7pt}\selectfont $\mathrm{AUC}_{5}\uparrow$}
    & \multirow{2}{*}{\fontsize{5.8pt}{7pt}\selectfont $\delta_{5}\uparrow$}
    & \multirow{2}{*}{\fontsize{5.8pt}{7pt}\selectfont RD$\downarrow$}
    & \multirow{2}{*}{\fontsize{5.8pt}{7pt}\selectfont $\mathrm{AUC}_{25}\uparrow$}
    & \multirow{2}{*}{\fontsize{5.8pt}{7pt}\selectfont $\delta_{25}\uparrow$}
    & \multirow{2}{*}{\fontsize{5.8pt}{7pt}\selectfont RD$\downarrow$}
    & & \multicolumn{2}{c}{\fontsize{5.8pt}{7pt}\selectfont All} & \multicolumn{2}{c}{\fontsize{5.8pt}{7pt}\selectfont Visible}
    & \multicolumn{2}{c}{\fontsize{5.8pt}{7pt}\selectfont All} & \multicolumn{2}{c}{\fontsize{5.8pt}{7pt}\selectfont Visible} \\
    \cmidrule(lr){2-3}\cmidrule(lr){4-5}\cmidrule(lr){16-17}\cmidrule(lr){18-19}\cmidrule(lr){20-21}\cmidrule(lr){22-23}
    & \fontsize{5.8pt}{7pt}\selectfont $\delta_{25}\uparrow$ & \fontsize{5.8pt}{7pt}\selectfont Rel$\downarrow$
    & \fontsize{5.8pt}{7pt}\selectfont $\delta_{25}\uparrow$ & \fontsize{5.8pt}{7pt}\selectfont Rel$\downarrow$
    & & & & & & & & & &
    & \fontsize{5.8pt}{7pt}\selectfont AUC$\uparrow$ & \fontsize{5.8pt}{7pt}\selectfont RelEPE$\downarrow$ & \fontsize{5.8pt}{7pt}\selectfont AUC$\uparrow$ & \fontsize{5.8pt}{7pt}\selectfont RelEPE$\downarrow$
    & \fontsize{5.8pt}{7pt}\selectfont AUC$\uparrow$ & \fontsize{5.8pt}{7pt}\selectfont RelEPE$\downarrow$ & \fontsize{5.8pt}{7pt}\selectfont AUC$\uparrow$ & \fontsize{5.8pt}{7pt}\selectfont RelEPE$\downarrow$ \\
    \hline

    \textbf{full model} & \cellcolor{highlight!0}95.36 & \cellcolor{highlight!80}0.066 & \cellcolor{highlight!80}94.34 & \cellcolor{highlight!80}0.076 & \cellcolor{highlight!80}0.057 & \cellcolor{highlight!80}2.965 & \cellcolor{highlight!0}0.127 & \cellcolor{highlight!80}52.96 & \cellcolor{highlight!80}83.58 & \cellcolor{highlight!80}0.037 & \cellcolor{highlight!80}71.17 & \cellcolor{highlight!80}92.84 & \cellcolor{highlight!80}0.103 & \cellcolor{highlight!80}86.98 & \cellcolor{highlight!80}62.51 & \cellcolor{highlight!0}0.144 & \cellcolor{highlight!80}67.70 & \cellcolor{highlight!0}0.106 & \cellcolor{highlight!80}68.12 & \cellcolor{highlight!0}0.126 & \cellcolor{highlight!80}74.66 & \cellcolor{highlight!0}0.093 \\
    w/o bridge pretrain & \cellcolor{highlight!0}94.58 & \cellcolor{highlight!0}0.073 & \cellcolor{highlight!0}93.39 & \cellcolor{highlight!0}0.084 & \cellcolor{highlight!0}0.079 & \cellcolor{highlight!0}9.651 & \cellcolor{highlight!0}0.167 & \cellcolor{highlight!0}43.89 & \cellcolor{highlight!0}73.86 & \cellcolor{highlight!0}0.059 & \cellcolor{highlight!0}62.65 & \cellcolor{highlight!0}86.50 & \cellcolor{highlight!0}0.139 & \cellcolor{highlight!0}85.78 & \cellcolor{highlight!0}60.27 & \cellcolor{highlight!0}0.153 & \cellcolor{highlight!0}65.02 & \cellcolor{highlight!0}0.118 & \cellcolor{highlight!0}66.64 & \cellcolor{highlight!0}0.130 & \cellcolor{highlight!0}72.33 & \cellcolor{highlight!0}0.102 \\
    w/o self KV & \cellcolor{highlight!80}95.41 & \cellcolor{highlight!0}0.068 & \cellcolor{highlight!0}94.29 & \cellcolor{highlight!0}0.076 & \cellcolor{highlight!0}0.062 & \cellcolor{highlight!0}2.997 & \cellcolor{highlight!80}0.126 & \cellcolor{highlight!0}52.14 & \cellcolor{highlight!0}82.22 & \cellcolor{highlight!0}0.039 & \cellcolor{highlight!0}69.28 & \cellcolor{highlight!0}92.26 & \cellcolor{highlight!0}0.107 & \cellcolor{highlight!0}86.88 & \cellcolor{highlight!0}62.46 & \cellcolor{highlight!80}0.143 & \cellcolor{highlight!0}67.64 & \cellcolor{highlight!80}0.106 & \cellcolor{highlight!0}68.07 & \cellcolor{highlight!80}0.125 & \cellcolor{highlight!0}74.58 & \cellcolor{highlight!80}0.092 \\
    camera token & \cellcolor{highlight!0}95.35 & \cellcolor{highlight!0}0.068 & \cellcolor{highlight!0}94.04 & \cellcolor{highlight!0}0.078 & \cellcolor{highlight!0}0.125 & \cellcolor{highlight!0}9.803 & \cellcolor{highlight!0}0.221 & \cellcolor{highlight!0}21.24 & \cellcolor{highlight!0}45.19 & \cellcolor{highlight!0}0.125 & \cellcolor{highlight!0}54.11 & \cellcolor{highlight!0}81.38 & \cellcolor{highlight!0}0.180 & - & - & - & - & - & - & - & - & - \\
    \hline
\end{tabular}%
}
\endgroup

    \end{minipage}
\end{figure*}

\paragraph{Long-horizon trajectory evaluation.}
To probe full feed-forward scalability without sliding-window stitching, we evaluate 500-frame sequences from Oxford-Spires~\citep{tao2025spires}, which features complex wide-area trajectories across large architectural sites.
Despite training on clips of at most 32 frames, our DINOv2 variant exhibits remarkable trajectory stability across these out-of-distribution horizons: it maintains smooth, locally faithful camera paths without the severe drift or jagged path oscillations seen in competing baselines (Figure~\ref{fig:trajectories_selected}), while running $1.7$--$4.3\times$ faster.
This zero-shot robustness reflects its permutation-invariant alternating attention without privileged frames, pairwise relative poses that extrapolate reliably across long sequences, and uncertainty-weighted pose-graph synchronization.
Appendix~\ref{sec:supp_long_horizon} (Figure~\ref{fig:supp_trajectories_all}) reports complete trajectories across all 13 sequences and variants; Section~\ref{sec:limitations} further discusses length extrapolation.

\subsection{Qualitative Comparisons}
\label{sec:qualitative_results}

Figure~\ref{fig:qualitative_results} compares MoGe-4D with 4RC, V-DPM, and Open-D4RT on dynamic videos from DAVIS~\citep{pont2017davis} and the iPhone subset of DyCheck~\citep{gao2022dycheck} (additional examples in Appendix~\ref{sec:supp_qualitative_comparisons}).
Prior methods often exhibit ghosted surfaces and drifting trajectories under simultaneous camera and foreground motion, whereas MoGe-4D recovers coherent scene geometry and smooth 3D trajectories across indoor and outdoor scenes.

\subsection{Ablation Study}
\label{sec:ablation_study}

\begin{figure*}[!t]
    \captionsetup{hypcap=false}
    \centering
    \begin{minipage}[t]{0.255\textwidth}
        \vspace{0pt}
        \centering
        \includegraphics[width=\linewidth]{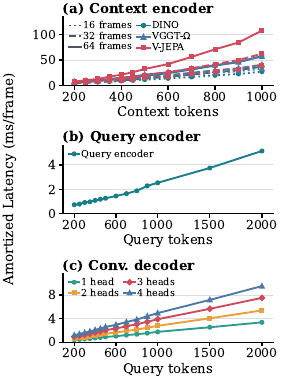}
        \vspace{-1.6em}
        \captionof{figure}{\textbf{Stage-wise runtime breakdown}: amortized latency across per-frame token budgets.}
        \label{fig:runtime_scaling}
    \end{minipage}\hfill
    \begin{minipage}[t]{0.725\textwidth}
        \vspace{0pt}
        \centering
        \includegraphics[width=\linewidth]{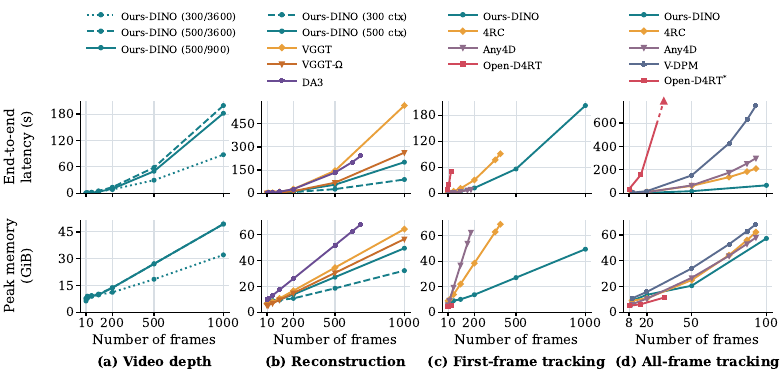}
        \vspace{-1.64em}
        \captionof{figure}{\textbf{End-to-end latency and peak memory versus sequence length.}
        Curves terminate at out-of-memory (OOM) or checkpoint sequence limits; dashed segments with upward triangles exceed the plotted vertical range ($^{*}$without query deduplication; see text).}
        \label{fig:runtime_e2e}
    \end{minipage}
\end{figure*}

\paragraph{Dual-encoder capacity allocation.}
In Figure~\ref{fig:ablation_tradeoffs}(a), enlarging the context encoder substantially reduces both reconstruction and tracking errors, whereas enlarging the query encoder from ViT-B to ViT-L brings only minor gains at ${\sim}2.5\times$ the per-query latency.
Thus, global spatiotemporal reasoning in the context encoder governs overall representational capacity, allowing a lightweight ViT-B query encoder to decode dense predictions accurately at low per-query cost.

\paragraph{Feature-bridge alignment and query attention.}
In Table~\ref{tab:camera_recovery_ablation}, randomly initializing the feature bridges instead of pre-aligning them degrades all metrics, dropping global-point AUC from 71.17\% to 62.65\%.
Pre-aligning the bridge projections helps the query encoder leverage pretrained context features effectively during end-to-end training.
Removing the query stream's own keys and values in conditioned blocks (\emph{w/o self KV}, leaving pure cross-attention) also causes a modest drop across both tasks, supporting the retention of intra-frame self-attention alongside context conditioning.

\paragraph{Point-map pose recovery vs.\ direct regression.}
Replacing closed-form point-map pose recovery with direct regression from a context camera token (trained with the camera loss of $\pi^3$~\citep{pi3}) substantially degrades pose accuracy: ATE increases from 0.057 to 0.125, and global-point AUC drops from 71.17\% to 54.11\% (Table~\ref{tab:camera_recovery_ablation}).
The gap is especially pronounced under wide baselines, indicating that solving poses from confidence-weighted 3D correspondences can be more reliable than direct token regression in challenging, long-range scenarios.

\paragraph{Inference token budgets.}
As per-frame token budgets vary in Figures~\ref{fig:ablation_tradeoffs}(b--d), error reductions saturate around 500 context tokens, 900 map-query tokens, and 250 pose-query tokens; we adopt these as our default settings for a favorable accuracy--speed trade-off.
Increasing map-query tokens beyond 900 barely changes aggregate metrics but progressively sharpens fine geometric boundaries (Figure~\ref{fig:tokens_for_depth}), enabling test-time adjustment between inference speed and spatial detail without re-running the heavy encoder.

\subsection{Runtime Analysis}
\label{sec:runtime_analysis}

\paragraph{Stage-wise breakdown.}
In Figure~\ref{fig:runtime_scaling}, context encoding dominates per-frame latency at high token budgets due to quadratic global attention, motivating our coarse context grid.
At comparable model sizes, V-JEPA 2.1 is slower on long sequences because it applies global attention in every layer, whereas DINOv2 and VGGT-$\Omega$ interleave frame-wise and global attention.
In contrast, the query encoder attends only to a single source frame and cached pair context, keeping per-query latency low and nearly linear in query tokens.
While the convolutional heads are memory-bandwidth bound on high-resolution maps, unused auxiliary heads can be disabled at inference.

\paragraph{End-to-end efficiency.}
Figure~\ref{fig:runtime_e2e} compares end-to-end latency and peak memory across sequence lengths at $518\times 336$ resolution.
For video depth and multiview reconstruction (Figure~\ref{fig:runtime_e2e}(a,b)), our compact context grid shortens global-attention sequences without reducing image resolution, substantially accelerating long-sequence inference.
For dense tracking (Figure~\ref{fig:runtime_e2e}(c,d)), especially all-frame tracking where all methods exhaustively evaluate all $T^2$ frame pairs without support-frame selection, query decoding dominates runtime.
By reusing cached context through a lightweight query encoder, MoGe-4D evaluates all $50^2$ dense pairs in a 50-frame clip about $4\times$ faster than 4RC.
Open-D4RT omits D4RT's query deduplication ($5$--$15\times$ speedup~\citep{d4rt}), yet even assuming a $10\times$ speedup on 32 frames, MoGe-4D remains ${\sim}10\times$ faster ($8.5\,\text{s}$ vs.\ $81.7\,\text{s}$).

\section{Limitations and Discussions}
\label{sec:limitations}

\paragraph{Object boundaries and small structures.}
Sharp depth and motion discontinuities at object boundaries, as well as very thin structures, remain challenging.
Finite query-token budgets and convolutional decoding can smooth abrupt 3D transitions across neighboring pixels, causing flying-point artifacts near edges or losing geometry on few-pixel structures.
Confidence and edge filtering suppress outliers; higher-resolution or adaptive decoding could preserve detail.

\paragraph{Domain-specific adaptation.}
MoGe-4D targets general scene-level geometry and motion perception, whereas specialized domains such as 3D facial capture and articulated hand tracking demand finer domain-specific precision beyond our current scope.
Adapting the model with targeted real-world supervision and synthetic data via agentic 3D modeling (Appendix~\ref{sec:supp_dynamic_scene}) is a promising direction.

\paragraph{Sequence-length extrapolation.}
Although trained on clips of at most 32 frames, our three variants extrapolate differently to longer sequences (Appendix~\ref{sec:supp_long_horizon}).
The DINOv2 variant generalizes most robustly beyond 500 frames, benefiting from strong intra-frame features and permutation-invariant alternating attention without temporal positional encodings.
VGGT-$\Omega$ shares this alternating attention but extrapolates slightly less well, likely due to its privileged reference-frame embedding, whereas V-JEPA 2.1 degrades beyond 100 frames even with YaRN extrapolation~\citep{peng2024yarn} due to its reliance on temporal RoPE.
More broadly, because full global attention scales quadratically with sequence length, feed-forward inference over hundreds of frames serves primarily as a generalization probe rather than a practical deployment mode.
Pairing our query interface with streamable or recurrent context encoders is a natural direction for long-horizon 4D perception.
\section{Conclusion}
\label{sec:conclusion}

We presented \textbf{MoGe-4D}, a feed-forward framework that unifies dense 3D geometry and motion prediction by querying paired allocentric and egocentric point maps.
By decoupling sequence-wide spatiotemporal reasoning in a coarse context encoder from on-demand decoding in a lightweight query encoder, our model reuses global scene representations across arbitrary query combinations and output resolutions.
To make training at this dense spatiotemporal granularity practical, we developed \textbf{RigIDTrack}, a synthetic data pipeline and compact representation that compress surface dynamics into rigidity-regularized affine bodies for arbitrary-view trajectory supervision.
MoGe-4D achieves state-of-the-art dense 3D tracking, matches dedicated 3D models in reconstruction accuracy, and substantially speeds up 4D inference.

\bibliography{citations}
\bibliographystyle{plainnat}

\clearpage
\appendix
\section*{Appendix}
\begin{itemize}\setlength{\itemsep}{4pt}
    \item \textbf{Appendix~\ref{sec:supp_d4d}: MoGe-4D Implementation Details}
    \begin{itemize}\setlength{\itemsep}{1pt}
        \item \ref{sec:supp_d4d_architecture} Architecture and Initialization \dotfill \pageref{sec:supp_d4d_architecture}
        \item \ref{sec:supp_d4d_training} Training Losses \dotfill \pageref{sec:supp_d4d_training}
        \item \ref{sec:supp_d4d_training_setup} Training Procedure \dotfill \pageref{sec:supp_d4d_training_setup}
        \item \ref{sec:supp_training_data_preprocessing} Training Data Preprocessing \dotfill \pageref{sec:supp_training_data_preprocessing}
        \item \ref{sec:supp_d4d_inference} Inference Procedure \dotfill \pageref{sec:supp_d4d_inference}
    \end{itemize}
    \item \textbf{Appendix~\ref{sec:supp_data}: RigIDTrack Data Pipeline Details}
    \begin{itemize}\setlength{\itemsep}{1pt}
        \item \ref{sec:supp_dynamic_scene} Procedural Dynamic Scene Generation \dotfill \pageref{sec:supp_dynamic_scene}
        \item \ref{sec:supp_body_clustering} Affine-Body Motion Compression and Rendering \dotfill \pageref{sec:supp_body_clustering}
        \item \ref{sec:supp_dataset_statistics} Synthetic Dataset Statistics \dotfill \pageref{sec:supp_dataset_statistics}
        \item \ref{sec:supp_rgbd_conversion} RigIDTrack Labeling from RGB-D Videos \dotfill \pageref{sec:supp_rgbd_conversion}
        \item \ref{sec:supp_related_dataset} Related Datasets for Dynamic 3D Tracking \dotfill \pageref{sec:supp_related_dataset}
    \end{itemize}
    \item \textbf{Appendix~\ref{sec:supp_evaluation}: Additional Experiment Details and Results}
    \begin{itemize}\setlength{\itemsep}{1pt}
        \item \ref{sec:supp_eval_data} Benchmark Preparation \dotfill \pageref{sec:supp_eval_data}
        \item \ref{sec:supp_recon_metrics} Multiview Reconstruction Metrics \dotfill \pageref{sec:supp_recon_metrics}
        \item \ref{sec:supp_tracking_metrics} 3D Tracking Metrics \dotfill \pageref{sec:supp_tracking_metrics}
        \item \ref{sec:supp_per_benchmark} Per-Benchmark Results \dotfill \pageref{sec:supp_per_benchmark}
        \item \ref{sec:supp_long_horizon} Long-Horizon Trajectory Evaluation on Oxford-Spires \dotfill \pageref{sec:supp_long_horizon}
        \item \ref{sec:supp_qualitative_comparisons} Additional Qualitative Comparisons \dotfill \pageref{sec:supp_qualitative_comparisons}
    \end{itemize}
\end{itemize}

\clearpage
\section{MoGe-4D Implementation Details}
\label{sec:supp_d4d}

\subsection{Architecture and Initialization}
\label{sec:supp_d4d_architecture}

\paragraph{Context backbone variants.}
We instantiate the context encoder with DINOv2 ViT-g/14, V-JEPA 2.1 ViT-G/16, or the VGGT-$\Omega$ backbone, containing approximately 1.14B, 1.01B, and 0.91B parameters, respectively (excluding the bridges, query encoder, and prediction heads; for VGGT-$\Omega$, we retain its image encoder and multiview aggregator without its task heads).
Each variant outputs multi-layer per-frame spatial features after cross-frame interaction, which are cached and reused across all source--target queries.
The DINOv2 variant omits frame positional encoding to support unordered image sets, whereas the V-JEPA 2.1 variant applies 3D rotary positional embeddings~\citep{su2024roformer} for temporally ordered video.

\paragraph{Query backbone and conditioning.}
All three variants share a DINOv2 ViT-Base query encoder with 12 transformer blocks, a feature width of 768, and 12 attention heads.
Eight of the 12 blocks receive bridged context keys and values alongside the query stream's own keys and values (six blocks take both source and target context; two take source context only), while the remaining four blocks apply standard self-attention.

\paragraph{Cross-backbone bridges.}
Because the context and query encoders have different feature dimensions and intermediate representations, each bridge applies LayerNorm followed by a linear projection from a selected context layer to the corresponding query-backbone layer.
We pretrain the bridges via standalone feature alignment on in-the-wild images and videos for 2,000 steps with a batch size of 8, minimizing $L_2$ and cosine distance between projected context features and target query-backbone features.
At each conditioned query block, the bridged context features are then projected into the additional keys and values described above.

\paragraph{Dense prediction heads.}
A shared convolutional neck decodes the spatial query features into the paired point maps $(\widehat{\boldsymbol{P}}_{i\rightarrow j,i}, \widehat{\boldsymbol{P}}_{i\rightarrow j,j})$ and auxiliary maps.
For each coordinate system, a dedicated tangent head predicts scale-invariant horizontal and vertical surface tangents $\widehat{\boldsymbol{T}}_{i\rightarrow j,k}\in\mathbb{R}^{H_i\times W_i\times 2\times 3}$, supervised against
\begin{equation}
    \boldsymbol{T}^{\mathrm{gt}}_{i\rightarrow j,k}(\boldsymbol{u})
    = \frac{1}{d_i^{\mathrm{gt}}(\boldsymbol{u})}
    \begin{bmatrix}
        \partial_u\boldsymbol{P}^{\mathrm{gt}}_{i\rightarrow j,k}(\boldsymbol{u}) \\
        \partial_v\boldsymbol{P}^{\mathrm{gt}}_{i\rightarrow j,k}(\boldsymbol{u})
    \end{bmatrix},
    \quad k\in\{i,j\},\qquad d_i^{\mathrm{gt}}(\boldsymbol{u})=[\boldsymbol{P}^{\mathrm{gt}}_{i\rightarrow i,i}(\boldsymbol{u})]_z,
    \label{eq:tangent_map}
\end{equation}
where normalizing by source depth removes global scale.
The neck also branches into heads predicting an uncertainty map $\boldsymbol{U}_{i\rightarrow j,k}\in\mathbb{R}_{+}^{H_i\times W_i}$ for each point map, along with source-aligned probability maps $\boldsymbol{M}^{\mathrm{sky}}_{i\rightarrow j}$, $\boldsymbol{M}^{\mathrm{vis}}_{i\rightarrow j}$, and $\boldsymbol{M}^{\mathrm{dyn}}_{i\rightarrow j}\in[0,1]^{H_i\times W_i}$ for sky, target-frame visibility, and dynamic regions.

\paragraph{Scalar prediction heads.}
In the DINOv2 and V-JEPA 2.1 variants, lightweight MLP heads operate on context class tokens to predict per-frame metric scale $s_i\in\mathbb{R}_{+}$ and pairwise covisibility $c_{ij}\in[0,1]$.

\subsection{Training Losses}
\label{sec:supp_d4d_training}

\paragraph{Query sampling.}
Given $N$ frames selected from a clip, the training query set $\mathcal{Q}$ comprises all $N$ self-queries and $O(N)$ cross-frame queries sampled using many-to-one, one-to-many, or random frame-pair patterns.
Both point maps of each query are supervised in their respective camera coordinates $k\in\{i,j\}$ (for self-queries, both outputs contribute to alignment and supervision against $\boldsymbol{P}^{\mathrm{gt}}_{k\rightarrow k,k}$).
All losses are evaluated over pixels with valid ground-truth annotations; validity masks and averaging constants are omitted below for brevity.

\paragraph{Scale and depth-shift alignment.}
We use the ROE solver of MoGe~\citep{moge} to fit a scale $s_k$ and z-shift $t_k$ to the self-query predictions:
\begin{equation}
    \min_{\{s_k,t_k\}}\sum_{k,\boldsymbol{u}}\left\lVert s_k\widehat{\boldsymbol{P}}_{k\rightarrow k,k}(\boldsymbol{u})+t_k\boldsymbol{e}_z-\boldsymbol{P}^{\mathrm{gt}}_{k\rightarrow k,k}(\boldsymbol{u})\right\rVert_1\big/\left\lVert\boldsymbol{P}^{\mathrm{gt}}_{k\rightarrow k,k}(\boldsymbol{u})\right\rVert.
    \label{eq:training_alignment}
\end{equation}
During training, we randomly alternate between shared-scale alignment ($s_k=s, t_k=0$) and per-camera scale--shift alignment (independent $(s_k,t_k)$ per frame).
The fitted $(s_k,t_k)$ are then applied to all predictions expressed in camera $k$.

\paragraph{Geometry losses.}
Point errors are normalized by ground-truth distance to camera $k$, with pixel weights $\omega_{ijk\boldsymbol{u}}$ scaling invisible points by $0.5$ and dynamic points by $2$.
The global geometry loss is
\begin{equation}
    \mathcal{L}^{\mathrm{G}}_{\mathrm{geo}}=\sum_{(i,j)\in\mathcal{Q}}\sum_{k\in\{i,j\}}\sum_{\boldsymbol{u}}\omega_{ijk\boldsymbol{u}}\left\lVert s_k\widehat{\boldsymbol{P}}_{i\rightarrow j,k}(\boldsymbol{u})+t_k\boldsymbol{e}_z-\boldsymbol{P}^{\mathrm{gt}}_{i\rightarrow j,k}(\boldsymbol{u})\right\rVert_1\big/\left\lVert\boldsymbol{P}^{\mathrm{gt}}_{i\rightarrow j,k}(\boldsymbol{u})\right\rVert.
    \label{eq:global_geometry_loss}
\end{equation}
For local geometry, we adopt the stochastic radial partitioning of MoGe-3~\citep{moge3}, which groups points by ray direction and log-distance at coarse ($\alpha=4$) and fine ($\alpha=32$) granularities after pooling predictions expressed in the same camera.
Within each region $R\in\mathcal{R}_{\alpha}$, we fit a local translation while retaining the camera scale $s_k$:
\begin{equation}
    \mathcal{L}^{\mathrm{L},\alpha}_{\mathrm{geo}}=\sum_{R\in\mathcal{R}_{\alpha}}\min_{\boldsymbol{t}_R}\sum_{(i,j,k,\boldsymbol{u})\in R}\omega_{ijk\boldsymbol{u}}\left\lVert s_k\widehat{\boldsymbol{P}}_{i\rightarrow j,k}(\boldsymbol{u})+\boldsymbol{t}_R-\boldsymbol{P}^{\mathrm{gt}}_{i\rightarrow j,k}(\boldsymbol{u})\right\rVert_1\big/\left\lVert\boldsymbol{P}^{\mathrm{gt}}_{i\rightarrow j,k}(\boldsymbol{u})\right\rVert.
    \label{eq:local_geometry_loss}
\end{equation}

An edge loss further penalizes angular differences between adjacent-pixel spatial gradients along the image axes:
\begin{equation}
    \mathcal{L}_{\mathrm{edge}}=\sum_{(i,j),\boldsymbol{u}}\sum_{k\in\{i,j\}}\sum_{d\in\{u,v\}}\angle\!\left(\partial_d\widehat{\boldsymbol{P}}_{i\rightarrow j,k}(\boldsymbol{u}),\partial_d\boldsymbol{P}^{\mathrm{gt}}_{i\rightarrow j,k}(\boldsymbol{u})\right)\big/\min(H_i,W_i).
\end{equation}

\paragraph{Motion losses.}
For each source frame $i$, we sort its queried target times and take temporal differences across adjacent pairs $(j_a,j_b)$ in both source-camera and target-camera coordinates:
\begin{equation}
    \begin{aligned}
        \Delta\widehat{\boldsymbol{P}}^{\mathrm{src}}&=s_i\left(\widehat{\boldsymbol{P}}_{i\rightarrow j_a,i}-\widehat{\boldsymbol{P}}_{i\rightarrow j_b,i}\right),\\
        \Delta\widehat{\boldsymbol{P}}^{\mathrm{tgt}}&=s_{j_a}\widehat{\boldsymbol{P}}_{i\rightarrow j_a,j_a}-s_{j_b}\widehat{\boldsymbol{P}}_{i\rightarrow j_b,j_b}.
    \end{aligned}
\end{equation}
Both coordinate choices use the same translation-aligned objective:
\begin{equation}
    \ell^{\mathrm{G/L}}_{\mathrm{mot}}=\sum_{R\in\mathcal{R}_{\mathrm{G/L}}}\min_{\boldsymbol{t}_R}\sum_{\boldsymbol{u}\in R}\omega^{\mathrm{G/L}}_{\boldsymbol{u}}\left\lVert\Delta\widehat{\boldsymbol{P}}(\boldsymbol{u})+\boldsymbol{t}_R-\Delta\boldsymbol{P}^{\mathrm{gt}}(\boldsymbol{u})\right\rVert_1\big/\max\!\left(\lVert\boldsymbol{P}^{\mathrm{gt}}_{j_a}(\boldsymbol{u})\rVert,\lVert\boldsymbol{P}^{\mathrm{gt}}_{j_b}(\boldsymbol{u})\rVert\right),
    \label{eq:motion_loss}
\end{equation}
where $\boldsymbol{P}^{\mathrm{gt}}_{j_a}$ and $\boldsymbol{P}^{\mathrm{gt}}_{j_b}$ are the ground-truth maps at times $j_a$ and $j_b$ in the corresponding camera frame.
Here $\mathcal{R}_{\mathrm{G}}$ is the full valid source map, and $\mathcal{R}_{\mathrm{L}}$ consists of radial regions at granularity $\alpha\in\{4,32\}$ constructed separately for each coordinate choice.
Pixel weights $\omega^{\mathrm{G/L}}_{\boldsymbol{u}}$ combine the visibility and dynamic weights at $j_a$ and $j_b$ via their geometric mean for $\mathcal{R}_{\mathrm{G}}$ and their maximum for $\mathcal{R}_{\mathrm{L}}$.

\paragraph{Uncertainty loss.}
Let $e_{ijk\boldsymbol{u}}=\lVert s_k\widehat{\boldsymbol{P}}_{i\rightarrow j,k}(\boldsymbol{u})+t_k\boldsymbol{e}_z-\boldsymbol{P}^{\mathrm{gt}}_{i\rightarrow j,k}(\boldsymbol{u})\rVert / s_k$ denote the unnormalized point residual in the predicted coordinate scale (removing the metric alignment scale $s_k$ without dividing by point distance).
Defining the log error $\Delta_{ijk\boldsymbol{u}}=\log U_{i\rightarrow j,k}(\boldsymbol{u})-\log\operatorname{stopgrad}(e_{ijk\boldsymbol{u}})$ (with both arguments clamped at $10^{-3}$) and its $\omega_{ijk\boldsymbol{u}}$-weighted mean $\bar{\Delta}$, we supervise $U_{i\rightarrow j,k}$ with a scale-invariant log (SILog) loss ($\gamma=0.5$):
\begin{equation}
    \mathcal{L}_{U}=\sum_{(i,j),\boldsymbol{u}}\sum_{k\in\{i,j\}}\omega_{ijk\boldsymbol{u}}\,\operatorname{Huber}_{1}\!\left(\Delta_{ijk\boldsymbol{u}}-\gamma\bar{\Delta}\right),
    \label{eq:uncertainty_loss}
\end{equation}
where $\operatorname{Huber}_{\beta}$ denotes the Huber loss with threshold $\beta$.

\paragraph{Surface tangent loss.}
Predicted surface tangents are supervised against Eq.~\ref{eq:tangent_map} in their respective camera frames:
\begin{equation}
    \mathcal{L}_{\mathrm{tan}}=\sum_{(i,j),\boldsymbol{u}}\sum_{k\in\{i,j\}}\operatorname{Huber}_{1}\!\left(\widehat{\boldsymbol{T}}_{i\rightarrow j,k}(\boldsymbol{u})-\boldsymbol{T}^{\mathrm{gt}}_{i\rightarrow j,k}(\boldsymbol{u})\right).
\end{equation}

\paragraph{Mask, metric-scale, and covisibility losses.}
Sky, visibility, and dynamic masks are supervised with binary cross-entropy:
\begin{equation}
    \mathcal{L}_{\mathrm{mask}}^{m}=-\sum_{(i,j),\boldsymbol{u}}\left[M^{m,\mathrm{gt}}\log\widehat{M}^{m}+(1-M^{m,\mathrm{gt}})\log(1-\widehat{M}^{m})\right],\quad m\in\{\mathrm{sky},\mathrm{vis},\mathrm{dyn}\}.
\end{equation}
On datasets with metric ground truth, the scalar metric-scale prediction $\widehat{s}_j$ is supervised against the detached alignment scale $s_j$:
\begin{equation}
    \mathcal{L}_{\mathrm{scale}}=\sum_j\operatorname{Huber}_{1}\!\left(\log\widehat{s}_j-\log\operatorname{stopgrad}(s_j)\right).
\end{equation}
Pairwise covisibility is supervised with cross-entropy against the ground-truth covisibility ratio $c_{ij}^{\mathrm{gt}}$:
\begin{equation}
    \mathcal{L}_{\mathrm{covis}}=-\sum_{i,j}\left[c_{ij}^{\mathrm{gt}}\log\widehat{c}_{ij}+(1-c_{ij}^{\mathrm{gt}})\log(1-\widehat{c}_{ij})\right].
\end{equation}

\paragraph{Loss weights and annotation quality.}
Training datasets are grouped into three annotation tiers: high-quality synthetic data (Tier A), high-precision real scans and reconstructions (Tier B), and noisier real-world reconstructions or LiDAR data (Tier C).
Table~\ref{tab:training_loss_weights} lists the loss weights enabled for each tier.

\begin{table}[htbp]
    \centering
    \caption{Training loss weights by geometry annotation quality (Tiers A--C).}
    \label{tab:training_loss_weights}
    \begin{tabular}{lccc}
        \toprule
        Loss & A & B & C \\
        \midrule
        Global geometry & 1.0 & 1.0 & 1.0 \\
        Local geometry ($\alpha=4$) & 1.5 & 1.5 & -- \\
        Local geometry ($\alpha=32$) & 2.0 & -- & -- \\
        Global motion & 1.0 & 1.0 & 1.0 \\
        Local motion ($\alpha=4$) & 1.0 & 1.0 & -- \\
        Local motion ($\alpha=32$) & 1.0 & -- & -- \\
        Uncertainty & 0.1 & 0.1 & 0.1 \\
        Edge & 1.0 & -- & -- \\
        Tangent & 0.1 & -- & -- \\
        Each mask (sky, visibility, dynamic) & 0.1 & 0.1 & 0.1 \\
        Metric scale & 0.1 & 0.1 & 0.1 \\
        Covisibility & 0.1 & 0.1 & 0.1 \\
        \bottomrule
    \end{tabular}
\end{table}

\subsection{Training Procedure}
\label{sec:supp_d4d_training_setup}

We initialize the encoders from pretrained backbones and the bridges from feature-alignment pretraining, then fine-tune the full model on a mixture of synthetic dynamic scenes and real and synthetic multiview datasets.

\paragraph{Optimization.}
We train for 160k steps on 64 NVIDIA A100-40GB GPUs with a global batch size of 128 clips, using 8-bit AdamW and the base learning rates in Table~\ref{tab:training_learning_rates}.
During the first 1,000 steps, only the newly initialized neck and prediction heads are updated at their base learning rate.
Over the next 1,000 steps, the learning rates of both encoders and the bridges warm up linearly from zero to their base values, after which all parameter groups follow a cosine decay to $10\%$ of their base rates.

\begin{table}[htbp]
    \centering
    \caption{Base learning rates by context backbone. All configurations use a DINOv2 ViT-B query encoder.}
    \label{tab:training_learning_rates}
    \begin{tabular}{lccc}
        \toprule
        Parameters & DINOv2 ViT-g & V-JEPA 2.1 ViT-g & VGGT-$\Omega$ \\
        \midrule
        Context encoder & $5\times10^{-6}$ & $2\times10^{-5}$ & $2\times10^{-5}$ \\
        Query encoder & $10^{-5}$ & $10^{-5}$ & $10^{-5}$ \\
        Bridges & $10^{-5}$ & $5\times10^{-5}$ & $5\times10^{-5}$ \\
        Neck and prediction heads & $10^{-4}$ & $10^{-4}$ & $10^{-4}$ \\
        \bottomrule
    \end{tabular}
\end{table}

\paragraph{Computational curriculum.}
Training starts with short clips and small query-token budgets, then progressively increases the maximum clip length from 4 to 32 frames alongside the query-token budget.

\paragraph{Data curriculum.}
We first train exclusively on RigIDTrack to establish dynamic tracking before introducing static multiview data, then expand to the full dataset mixture (Table~\ref{tab:training_data_mixture}).
In the final stage, clips with and without dynamic tracking annotations are sampled at an approximate 1:1 ratio, with RigIDTrack accounting for over 40\% of all training samples.

\begin{table}[htbp]
    \centering
    \caption{Training datasets and clip counts. A--C denote geometry annotation quality; $^\dagger$dynamic tracking labels generated by our RGB-D pipeline.}
    \label{tab:training_data_mixture}
    \begin{tabular}{llcr}
        \toprule
        Dataset & Quality & \shortstack{Dynamic\\tracking} & Clips \\
        \midrule
        RigIDTrack (multi-object) & A & Yes & 9,598 \\
        RigIDTrack (single-object) & A & Yes & 9,389 \\
        OmniWorldGame~\citep{omniworld} & A & Yes$^{\dagger}$ & 11,424 \\
        DynamicReplica~\citep{dynamicstereo} & A & Yes & 503 \\
        Hypersim~\citep{hypersim} & A & No & 769 \\
        TartanAir V2~\citep{tartanair} & A & No & 1,014 \\
        EDEN~\citep{eden} & A & No & 735 \\
        MatrixCity~\citep{matrixcity} & A & No & 48 \\
        MidAir~\citep{midair} & A & No & 192 \\
        MVS-Synth~\citep{deepmvs} & A & No & 120 \\
        Structured3D~\citep{structured3d} & A & No & 19,760 \\
        \midrule
        ScanNet++~\citep{scannetpp} & B & No & 222 \\
        BlendedMVS~\citep{blendedmvs} & B & No & 502 \\
        MegaDepth~\citep{megadepth} & B & No & 267 \\
        ARKitScenes~\citep{arkitscenes} & B & No & 2,232 \\
        \midrule
        Argoverse 2~\citep{argoverse2} & C & Yes$^{\dagger}$ & 7,000 \\
        Waymo~\citep{waymo} & C & Yes$^{\dagger}$ & 3,723 \\
        DL3DV~\citep{dl3dv} & C & No & 6,378 \\
        \bottomrule
    \end{tabular}
\end{table}

\paragraph{Data augmentation.}
We vary temporal sampling strides, spatial crops, and aspect ratios, allowing frames within a clip to have different resolutions.
Photometric augmentations include color jitter, JPEG compression, blur, and depth-of-field effects, with synthetic datasets additionally receiving shot noise.

\subsection{Training Data Preprocessing}
\label{sec:supp_training_data_preprocessing}

\paragraph{Argoverse 2 and Waymo.}
The both driving datasets provide accurate by sparse LiDAR captured depth. We first complete their sparse depth maps using VideoLDCM~\citep{vigeo,ldcm}, then run the data preprocessing (Sec.~\ref{sec:supp_rgbd_conversion}) to build RigIDTrack annotations from these RGB-D videos.

\paragraph{OmniWorldGame.}
OmniWorldGame provides synthetic depth and camera extrinsics without long-term tracking labels.
We run the data preprocessing (Sec.~\ref{sec:supp_rgbd_conversion}) to build RigIDTrack annotations from it.

\paragraph{DL3DV.}
We refine the MVS depth maps from the CUT3R release~\citep{cut3r} using VideoLDCM~\citep{vigeo,ldcm} to recover fine-scale geometric detail.

\subsection{Inference Procedure}
\label{sec:supp_d4d_inference}

Throughout inference, each raw prediction $\widehat{\boldsymbol{P}}_{i\rightarrow j,k}$ ($k\in\{i,j\}$) is mapped to camera-space coordinates $\boldsymbol{P}_{i\rightarrow j,k}$ and world-space coordinates $\boldsymbol{P}^{\mathrm{world}}_{i\rightarrow j}$ via
\begin{equation}
    \boldsymbol{R}_k\boldsymbol{P}^{\mathrm{world}}_{i\rightarrow j}+\boldsymbol{t}_k
    =\boldsymbol{P}_{i\rightarrow j,k}
    =\alpha_k(\widehat{\boldsymbol{P}}_{i\rightarrow j,k}+\delta_k\boldsymbol{e}_z),\qquad k\in\{i,j\},
    \label{eq:inference_coordinate_spaces}
\end{equation}
where $(\boldsymbol{R}_k,\boldsymbol{t}_k)$ is the world-to-camera pose, $\alpha_k>0$ is a per-camera scale factor ($\alpha_k=1$ in rigid mode, or estimated in similarity mode), and $\delta_k$ is the optical-axis shift.

\paragraph{Camera focal length and z-shift recovery.}
For each self-query $(j,j)$, let $\widehat{\boldsymbol{P}}_{j\rightarrow j,j}(\boldsymbol{u})=(\widehat{x}_{j\boldsymbol{u}},\widehat{y}_{j\boldsymbol{u}},\widehat{z}_{j\boldsymbol{u}})$ and let $\boldsymbol{q}_j(\boldsymbol{u})$ denote centered pixel coordinates normalized by the image half-diagonal.
Assuming a centered principal point, a scalar focal length $f_j$ and shift $\delta_j$ minimize the weighted reprojection error:
\begin{equation}
    E_j(f_j,\delta_j)=\sum_{\boldsymbol{u}}w^{\mathrm{proj}}_{j\boldsymbol{u}}\left\lVert f_j\frac{(\widehat{x}_{j\boldsymbol{u}},\widehat{y}_{j\boldsymbol{u}})}{\widehat{z}_{j\boldsymbol{u}}+\delta_j}-\boldsymbol{q}_j(\boldsymbol{u})\right\rVert^2,
    \label{eq:calibration_reprojection}
\end{equation}
with frame-normalized pixel weights $w^{\mathrm{proj}}_{j\boldsymbol{u}}\propto(1-M^{\mathrm{sky}}_j)/U_{j\rightarrow j}$ over valid finite points.
We support either per-frame (\emph{varying}) or sequence-shared (\emph{fixed}) focal lengths:
\begin{equation}
    \begin{aligned}
        \text{Varying focal:}\quad &\min_{\{f_j,\delta_j\}}\sum_j E_j(f_j,\delta_j),\\
        \text{Fixed focal:}\quad &\min_{f,\{\delta_j\}}\sum_j E_j(f,\delta_j).
    \end{aligned}
    \label{eq:calibration_modes}
\end{equation}
Both modes parameterize $\delta_j=-\min_{\boldsymbol{u}}\widehat{z}_{j\boldsymbol{u}}+\exp(\eta_j)$ to enforce positive shifted depths.
The varying-focal objective is solved per frame by analytically eliminating $f_j$ and optimizing $\eta_j$ via Levenberg--Marquardt~\citep{marquardt1963}; the fixed-focal objective is solved jointly across frames using trust-region reflective least squares~\citep{branch1999} with a sparse analytic Jacobian and LSMR solver~\citep{fong2011lsmr}.

\paragraph{Camera pose recovery.}
We estimate global camera poses by aligning paired point maps over a sparse view graph $\mathcal{E}$ constructed from symmetric predicted covisibility scores (combining a maximum spanning tree with greedy D-optimal edge augmentation, and evaluating both query directions per selected edge).

For each directed edge $(i,j)\in\mathcal{E}$, let $\boldsymbol{p}=\widehat{\boldsymbol{P}}_{i\rightarrow j,i}$ and $\boldsymbol{q}=\widehat{\boldsymbol{P}}_{i\rightarrow j,j}$, with pointwise weights $w_{ij}=(1-M_{i\rightarrow j}^{\mathrm{dyn}})(1-M_{i\rightarrow j}^{\mathrm{sky}})M_{i\rightarrow j}^{\mathrm{vis}}/\max(U_{i\rightarrow j,i},U_{i\rightarrow j,j})^2$.
Writing $\boldsymbol{A}_{ij}=(\alpha_i/\alpha_j)\boldsymbol{R}_j\boldsymbol{R}_i^\top$ and $\boldsymbol{b}_{ij}=\boldsymbol{t}_j/\alpha_j-\delta_j\boldsymbol{e}_z-\boldsymbol{A}_{ij}(\boldsymbol{t}_i/\alpha_i-\delta_i\boldsymbol{e}_z)$, global alignment minimizes
\begin{equation}
    \min_{\{\boldsymbol{R}_j,\alpha_j,\boldsymbol{t}_j\}} E_{\mathrm{pose}}=\sum_{(i,j)\in\mathcal{E}}\sum_{\boldsymbol{u}\in\mathcal{I}_{ij}}w_{ij}(\boldsymbol{u})\left\lVert\boldsymbol{A}_{ij}\boldsymbol{p}(\boldsymbol{u})+\boldsymbol{b}_{ij}-\boldsymbol{q}(\boldsymbol{u})\right\rVert^2,
    \label{eq:registration_point_objective}
\end{equation}
where $\boldsymbol{R}_j\in\mathrm{SO}(3)$ and $\alpha_j>0$.
Each edge is summarized by its total weight, weighted centroids, and centered covariance matrices so that $E_{\mathrm{pose}}$ is evaluated without revisiting dense pixels.
Following similarity synchronization~\citep{yu2024simsync}, we first estimate pairwise relative rotations via SVD~\citep{kabsch1976,umeyama1991} and synchronize them under the weighted chordal objective~\citep{dellaert2020shonan} (spectral initialization followed by Procrustes refinement).
Relative scales are estimated from ratios of weighted RMS distances to centroids and synchronized by log-space linear least squares.
With rotations and scales fixed, the quadratic translation subproblem is solved in closed form via a weighted graph Laplacian.

\paragraph{Support frame selection.}
Given a selected support set $\mathcal{S}$, the visibility coverage map of frame $i$ is $V_i^{\mathcal{S}}(\boldsymbol{u})=\max_{j\in\mathcal{S}}M_{i\rightarrow j}^{\mathrm{vis}}(\boldsymbol{u})$.
A frame satisfies the coverage criterion $\rho$ (default $0.95$) when the $(1-\rho)$ quantile of $V_i^{\mathcal{S}}$ is at least $0.5$.
Starting from $\mathcal{S}=\{1\}$, we iteratively add the uncovered frame with the lowest mean coverage until all frames meet the criterion or the support-frame budget is reached.
During reconstruction at target time $t$, we aggregate $\boldsymbol{P}^{\mathrm{world}}_{i\rightarrow t}$ over $i\in\mathcal{S}$.
\clearpage
\section{RigIDTrack Data Pipeline Details}
\label{sec:supp_data}

Sections~\ref{sec:supp_dynamic_scene}--\ref{sec:supp_dataset_statistics} detail our synthetic RigIDTrack dataset, including procedural dynamic scene generation, affine-body motion compression and rendering, and dataset statistics.
Beyond synthetic rendering, Section~\ref{sec:supp_rgbd_conversion} presents an annotation pipeline that builds RigIDTrack representations from existing dynamic RGB-D videos by lifting 2D tracks and discovering motion-consistent affine bodies.
Section~\ref{sec:supp_related_dataset} compares RigIDTrack with related dynamic 3D tracking datasets.

\subsection{Procedural Dynamic Scene Generation}
\label{sec:supp_dynamic_scene}

Built with Blender, the generator procedurally samples indoor and outdoor environments and populates them with humans, animated objects, static props, and vehicles drawn from Objaverse~\citep{deitke2022objaverse}, Objaverse-XL~\citep{deitke2023objaversexl}, and licensed commercial assets, randomizing their categories, counts, scales, and placements.
Smooth, collision-aware camera trajectories are simulated with varying speeds and spatial ranges, guiding object placement toward visible regions while rejecting intersecting bounding boxes.
Objects can have native skeletal or non-rigid animation and undergo rigid-body physics simulation with collision proxies and turbulence-driven impulses.
Camera orientation alternates between free exploration and gazing at nearby visible objects.

\paragraph{Limitations and agentic scene synthesis.}
While rule-based procedural generation scales effectively, the resulting appearance and motion distributions remain constrained by hand-crafted placement and simulation rules.
Given recent advances in multimodal large language model (MLLM) agents for 3D modeling, agentic workflows offer a more scalable and higher-quality alternative for synthesizing diverse dynamic scenes.
Crucially, our affine-body compression (Sec.~\ref{sec:supp_body_clustering}) operates directly on animated meshes, thus it is agnostic to upstream scene authoring and can directly process agent-generated scenes.

\subsection{Affine-Body Motion Compression and Rendering}
\label{sec:supp_body_clustering}

We compress surface animation into rigidity-regularized affine bodies by grouping mesh faces whose relative affine poses remain consistent over time.
Static and rigid objects can omit this process, and we focus on deforming meshes.
Persistent face IDs and local UV coordinates provide exact surface correspondences across frames.
We first fit per-face canonical-to-world pose sequences $A_{p,t}^{\mathrm{face}}$ and accumulate canonical point statistics, then group motion-consistent faces via Algorithm~\ref{alg:rigidtrack_clustering}, and finally refit a shared trajectory $A_{b,t}$ for each body.

\begin{algorithm}
    \caption{Motion-consistent grouping}
    \label{alg:rigidtrack_clustering}
    \begin{algorithmic}[1]
        \Require Face pose sequences $A_{p,t}^{\mathrm{face}}$; point statistics $(\boldsymbol\mu_p,C_p,W_p,n_p)$
        \Require Camera-space face centers $\boldsymbol y_{pt}$; compression tolerance $\varepsilon=10^{-3}$
        \Ensure Face-to-body labels
        \item[\textbf{Notation:}]
        \Statex $A_{p,t}^{\mathrm{face}}$: face canonical-to-world pose.
        \Statex $M_p$: time-independent face-to-seed offset.
        \Statex $\boldsymbol\mu_p,C_p,n_p$: canonical mean, covariance, and observation count.
        \Statex $W_p,S_p$: total distance weight and canonical moment matrix; $q_p,e_p$: priority and error.
        \Statex $\mathcal U,\mathcal V$: unassigned faces and surviving members of the current body.
        \Statex $T$: frame count; $\mathcal J$: test frames; $s$: selected seed; $h$: test stride.
        \State Compute $S_p$ using Eq.~\ref{eq:rigidtrack_consistency}; set priority $q_p\gets W_p\operatorname{tr}(C_p)$
        \State Initialize $\mathcal U$ to all faces
        \While{$\mathcal U\ne\varnothing$} \Comment{Continue until all faces are assigned}
            \State $s\gets\arg\max_{p\in\mathcal U}q_p$ \Comment{Choose a reliable motion representative}
            \State $\mathcal V\gets\mathcal U$
            \State $h\gets\max(\lfloor T/4\rfloor,1)$
            \Loop \Comment{Coarse-to-fine strides reject inconsistent candidates cheaply}
                \State $\mathcal J\gets\{1,1+h,1+2h,\ldots\}\cap\{1,\ldots,T\}$
                \State Fit $M_p$ on $\mathcal J$ for $p\in\mathcal V$ using Eq.~\ref{eq:rigidtrack_offset_fit}
                \State $\mathcal V\gets\{p\in\mathcal V:e_p(A_p^{\mathrm{face}},A_s^{\mathrm{face}})<\varepsilon\}$ using Eq.~\ref{eq:rigidtrack_consistency}
                \State \textbf{if} $h=1$ \textbf{then break} \Comment{All frames have been tested}
                \State $h\gets\lfloor h/2\rfloor$
            \EndLoop
            \State Give $\mathcal V$ one body label; $\mathcal U\gets\mathcal U\setminus\mathcal V$
        \EndWhile
    \end{algorithmic}
\end{algorithm}

\paragraph{Rigidity-regularized face pose fitting.}
For each face $p$, we sample surface points via UV correspondence across frames to obtain world-space trajectories $\boldsymbol{y}_{tpk}$, and estimate its canonical-to-world pose $A_{p,t}^{\mathrm{face}}$ sequentially against accumulated canonical observations.
At a single step (suppressing face and time indices), write $A^{\mathrm{face}}(\boldsymbol x)=L\boldsymbol x+\boldsymbol h$, and let $(\boldsymbol{x}_a,\boldsymbol{y}_a)$ denote corresponding canonical and world-space samples with weights $\omega_a$, means $\boldsymbol{\mu}_x,\boldsymbol{\mu}_y$, and covariances $C_{xx},C_{yy}$.
With isotropically regularized covariances $\bar C_{xx}=C_{xx}+\lambda\operatorname{tr}(C_{xx})I/3$ and $\bar C_{yy}=C_{yy}+\lambda\operatorname{tr}(C_{yy})I/3$ ($\lambda=10^{-2}$) and Mahalanobis norm $\|\boldsymbol v\|_C^2=\boldsymbol v^\top C^{-1}\boldsymbol v$, we solve
\begin{equation}
    \begin{aligned}
        \min_{A^{\mathrm{face}}}\quad &\frac{1}{2}\sum_a\omega_a\left[\left\lVert A^{\mathrm{face}}(\boldsymbol{x}_a)-\boldsymbol{y}_a\right\rVert_{\bar C_{yy}}^2
        +\left\lVert \bigl(A^{\mathrm{face}}\bigr)^{-1}(\boldsymbol{y}_a)-\boldsymbol{x}_a\right\rVert_{\bar C_{xx}}^2\right],\\
        \text{subject to}\quad &L\bar C_{xx}L^\top=\bar C_{yy}.
    \end{aligned}
    \label{eq:rigidtrack_pose_fit}
\end{equation}
Whitening both coordinate sets reduces Eq.~\ref{eq:rigidtrack_pose_fit} to orthogonal Procrustes~\citep{schonemann1966}, whose SVD yields $L$ and $\boldsymbol h=\boldsymbol\mu_y-L\boldsymbol\mu_x$.
After each step, inverse-transformed observations update the canonical statistics, with pair weights set to the geometric mean of the current and historical observation weights (unit weights for initial face fitting, and inverse-squared camera-distance weights for final body refitting).

\paragraph{Face--body motion consistency.}
Let $\{\widetilde{\boldsymbol z}_{pa}\}_{a=1}^{n_p}$ denote the homogeneous canonical samples of face $p$, with 3D mean $\boldsymbol\mu_p$ and covariance $C_p$.
To test whether face $p$ is compatible with seed $s$ over test frames $\mathcal J$, we first fit a time-independent affine offset $M_p$ between their canonical spaces by linear least squares:
\begin{equation}
    M_p=\arg\min_{M\,\mathrm{affine}}\sum_{t\in\mathcal J}\sum_{a=1}^{n_p}\left\lVert\left(A_{p,t}^{\mathrm{face}}-A_{s,t}^{\mathrm{face}}M\right)\widetilde{\boldsymbol z}_{pa}\right\rVert^2.
    \label{eq:rigidtrack_offset_fit}
\end{equation}
With $\Delta_{tp}=A_{p,t}^{\mathrm{face}}-A_{s,t}^{\mathrm{face}}M_p$ and camera-space face center $\boldsymbol y_{pt}\in\mathbb R^3$, the maximum distance-normalized RMS discrepancy across $\mathcal J$ is evaluated directly from canonical second moments $S_p$:
\begin{equation}
    e_p(A_p^{\mathrm{face}},A_s^{\mathrm{face}})=\max_{t\in\mathcal J}\frac{1}{\|\boldsymbol y_{pt}\|}\sqrt{\operatorname{tr}\!\left(\Delta_{tp}^\top\Delta_{tp}S_p\right)},\qquad
    S_p=\begin{bmatrix}C_p+\boldsymbol\mu_p\boldsymbol\mu_p^\top&\boldsymbol\mu_p\\\boldsymbol\mu_p^\top&1\end{bmatrix}.
    \label{eq:rigidtrack_consistency}
\end{equation}
Face $p$ is accepted if $e_p<\varepsilon$ ($\varepsilon=10^{-3}$).

\paragraph{Motion-consistent grouping and body pose refitting.}
Algorithm~\ref{alg:rigidtrack_clustering} greedily selects the unassigned seed $s$ with maximum weighted spatial spread $q_p=W_p\operatorname{tr}(C_p)$ (where $W_p$ sums inverse-squared camera distances over face $p$'s observations) and filters candidate faces using coarse-to-fine frame strides $h$.
Once a body's member set $\mathcal V$ is finalized, we pool all member surface trajectories and refit a single canonical-to-world trajectory $A_{b,t}$ via Eq.~\ref{eq:rigidtrack_pose_fit} with inverse-squared camera-distance weights.

\paragraph{Time complexity.}
Face pose fitting and body pose refitting each take $O(NT)$ time for $N$ sampled surface points per frame and $T$ frames.
Grouping $K$ faces into $B$ bodies takes $O(KBT)$ time in the worst case ($B\le K$), while coarse-to-fine stride rejection reduces typical cost toward $O(KB+KT)$.

\paragraph{Rendering the compact record.}
An annotation pass (with motion blur and anti-aliasing disabled) maps per-pixel face IDs to global body IDs.
The stored per-frame record consists of RGB, depth, camera parameters, the pixelwise body-ID map, and per-body $4\times4$ transforms $A_{b,t}$; dense 3D trajectories are recovered via Eq.~\ref{eq:rigidtrack_transport}.

\subsection{Synthetic Dataset Statistics}
\label{sec:supp_dataset_statistics}

RigIDTrack provides RGB videos alongside pixel-aligned depth and body-ID maps at $1280\times960$ resolution and 30 FPS.
The training set comprises multi-object scenes with interacting foreground entities and single-object scenes with an orbiting camera, totaling 18.99K sequences, 5.07M frames, and 46.95 hours of video (Table~\ref{tab:rigidtrack_scale}).
The test set contains 88 sequences generated under the same configuration with held-out scene combinations and motion trajectories.

\begin{table}[htbp]
    \centering
    \caption{RigIDTrack training-set scale.}
    \label{tab:rigidtrack_scale}
    \begin{tabular}{lrrr}
        \toprule
        & Multi-object & Single-object & Total \\
        \midrule
        Sequences & 9.60K & 9.39K & 18.99K \\
        Frames & 4.08M & 0.995M & 5.07M \\
        Video duration (hours) & 37.74 & 9.21 & 46.95 \\
        Mean sequence length (seconds) & 14.15 & 3.53 & 8.90 \\
        \bottomrule
    \end{tabular}
\end{table}

\begin{figure}[htbp]
    \centering
    \includegraphics[width=\linewidth]{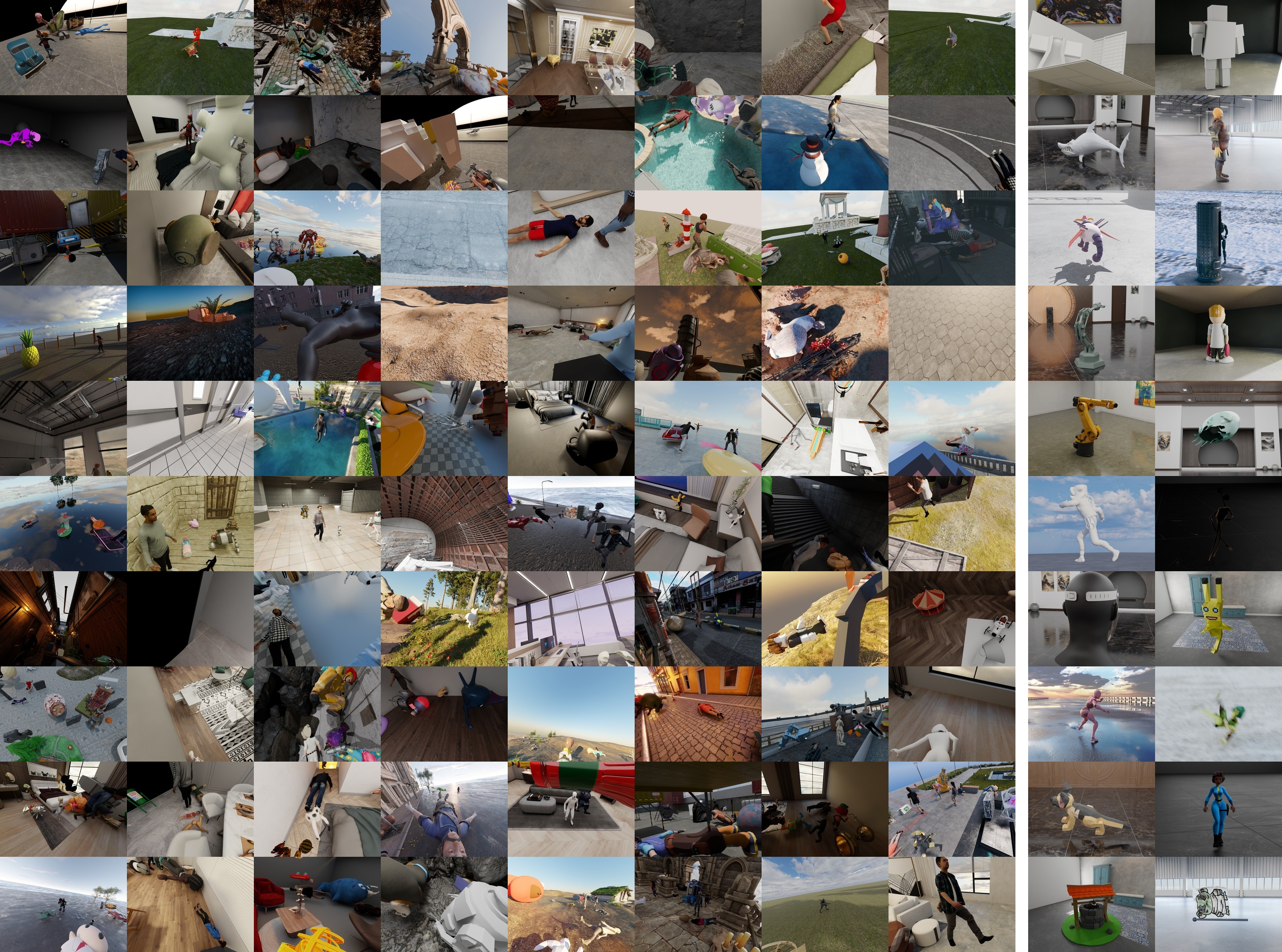}
    \setlength{\abovecaptionskip}{3pt}
    \caption{Appearance diversity across 100 randomly selected training sequences: 80 multi-object scenes on the left and 20 single-object scenes on the right.}
    \label{fig:rigidtrack_appearance}
\end{figure}

\begin{figure}[htbp]
    \centering
    \includegraphics[width=\linewidth]{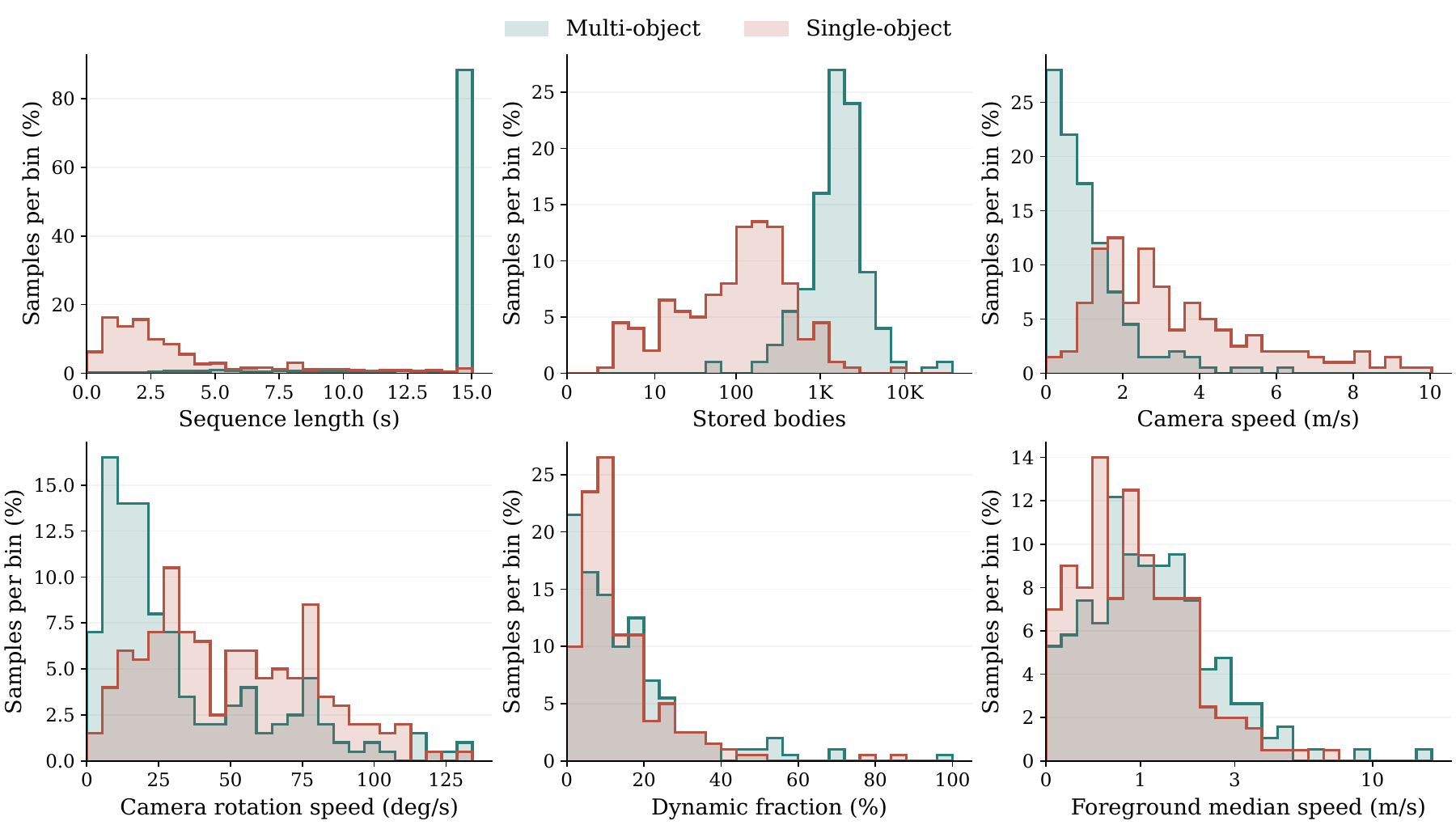}
    \caption{Training-set distributions of sequence length, body count, camera motion, moving-pixel fraction, and median foreground speed.}
    \label{fig:rigidtrack_distributions}
\end{figure}

\paragraph{Diversity and distribution.}
The asset pools comprise 391 environments (277 indoor, 114 outdoor), approximately 9.9K animated objects, 10K static objects, 439 human assets, and 194 vehicles.
Multi-object scenes span all 391 environments (44.4\% indoor, 55.6\% outdoor), populating 5--19 entities indoors and 9--29 entities (including vehicles) outdoors.
Single-object scenes pair 12 background templates with approximately 9.17K distinct animated objects.
Figure~\ref{fig:rigidtrack_appearance} shows sampled RGB frames from both subsets, and Figure~\ref{fig:rigidtrack_distributions} reports the distributions of sequence length, body count, camera motion, moving-pixel fraction, and foreground speed.

\paragraph{Storage statistics.}
Table~\ref{tab:rigidtrack_storage} reports per-frame storage and body counts across 200 sampled frames per subset.
The motion record (body-ID map plus body transforms) averages 120.3 KiB per frame for multi-object scenes (2.27K stored bodies, 237 visible per frame) and 33.6 KiB for single-object scenes (259 stored bodies, 115 visible per frame), remaining smaller than the compressed depth map in both subsets.

\begin{table}[htbp]
    \centering
    \caption{Per-frame storage (KiB) and body counts of RigIDTrack (P95: 95th percentile).}
    \label{tab:rigidtrack_storage}
    \begin{tabular}{lrrrr}
        \toprule
        & \multicolumn{2}{c}{Multi-object} & \multicolumn{2}{c}{Single-object} \\
        \cmidrule(lr){2-3}\cmidrule(lr){4-5}
        & Mean & P95 & Mean & P95 \\
        \midrule
        RGB image & 323.1 & 559.3 & 267.8 & 477.0 \\
        Depth image & 453.3 & 831.5 & 152.1 & 247.4 \\
        Motion record & 120.3 & 242.6 & 33.6 & 72.5 \\
        \quad {\small Body-ID image} & {\small 29.3} & {\small 57.5} & {\small 22.5} & {\small 44.7} \\
        \quad {\small Body transforms} & {\small 91.0} & {\small 205.1} & {\small 11.1} & {\small 38.3} \\
        \midrule
        Stored bodies & 2.27K & 4.82K & 259 & 904 \\
        Visible bodies & 237 & 710 & 115 & 374 \\
        \bottomrule
    \end{tabular}
\end{table}

\subsection{RigIDTrack Labeling from RGB-D Videos}
\label{sec:supp_rgbd_conversion}

To broaden the distribution of dynamic tracking supervision, we build RigIDTrack annotations from existing posed dynamic RGB-D videos using dense 2D tracking priors.
Extending the affine-body motion grouping in Sec.~\ref{sec:supp_body_clustering}, the pipeline lifts 2D tracks into 3D and discovers motion-consistent affine bodies to recover long-range trajectories.

\paragraph{Annotation pipeline.}
The pipeline runs overlapping sliding windows to incrementally discover affine bodies from lifted point trajectories, maintaining registered bodies and an active subset for motion grouping.
\begin{enumerate}
\item[1)] \textbf{Trajectory lifting.}
We use AllTracker~\citep{alltracker}, a 2D Dense Tracking model, to track pixels from a reference frame within the window, and the corresponding per-frame depths and camera parameters lift these 2D tracks into world-space 3D trajectories.
\item[2)] \textbf{Adaptive patch extraction.}
The trajectory map is subdivided into image-space patches until each patch's trajectories can be fitted by a shared affine motion within the error tolerance, discarding patches that are too small or insufficiently supported.
These patches replace mesh faces as the motion primitives in the subsequent grouping step.
\item[3)] \textbf{Motion grouping and body association.}
The patches and active bodies are jointly merged using the motion-grouping algorithm (Algorithm~\ref{alg:rigidtrack_clustering}), extended to accommodate missing or invalid trajectory observations.
Motion consistency is evaluated over valid shared frames with sufficient temporal overlap.
Patches sharing motion with a registered body inherit its identity and extend its lifetime; otherwise, they form new bodies, grouping mutually consistent patches together.
A body that receives no new patches is retired when its remaining overlap expires in the next window.
\item[4)] \textbf{Transformation fitting and ID-map rendering.}
We incrementally refit body transformations across overlapping windows and generate body-ID maps by projecting and rasterizing the valid, body-labeled surface geometry.
\end{enumerate}

\paragraph{Streaming and long-range tracking.}
The sliding-window pipeline processes videos incrementally, with total processing time growing linearly with video length.
Body identities can persist across multiple tracker windows, extending tracking beyond the lifespan of individual 2D tracks.
Tracking at the body level also supports partial occlusion: when a point's 2D track or depth observation is lost, its position can still be inferred from the shared body transformation estimated from sufficiently supported visible parts.
By discovering shared motion among patches, the algorithm can thus maintain correspondences beyond those directly provided by the 2D tracker and depth.

\paragraph{Input requirements and conservative labeling.}
The algorithm depends on the quality of RGB-D observations, camera parameters, and 2D tracking.
In particular, high-quality, temporally stable depth is important, as depth jitter can prevent reliable motion fitting and association.
We therefore adopt a conservative tracking principle: rather than seeking to recover every body or connect every long-range correspondence, we require accepted patch fits and body associations to satisfy the prescribed error tolerance on valid observations.
Unreliable regions are left unmerged or discarded, prioritizing the reliability of retained annotations over coverage.

\subsection{Related Datasets for Dynamic 3D Tracking}
\label{sec:supp_related_dataset}

\begin{table*}[!htbp]
\centering
\caption{
Comparison of representative synthetic datasets for 3D tracking. \emph{Compressed representation} stores motion as compact body transformations rather than per-vertex mesh trajectories ($^\dagger$for rigid objects via object coordinates and poses).
}
\label{tab:dataset_comparison}

\providecommand{\cmark}{\textcolor{green!60!black}{$\checkmark$}}
\providecommand{\xmark}{\textcolor{red!75!black}{$\times$}}
\providecommand{\dmark}{\textcolor{yellow!70!black}{--}}

\resizebox{\textwidth}{!}{%
\renewcommand{\arraystretch}{1.3}
\begin{tabular}{lccccc}
\hline
& \textbf{MOVi-E (Kubric)}
& \textbf{PointOdyssey}
& \textbf{Dynamic Replica}
& \textbf{Syn4D}
& \textbf{Ours} \\
\hline

Resolution ($W\times H$)
& $256\times256$
& $960\times540$
& $1280\times720$
& \dmark
& \textbf{\boldmath$1280\times960$} \\

Frame rate (FPS)
& 12
& 30
& 30
& 30
& \textbf{30} \\

Total clips
& 11.00K
& 104
& 524
& 4.7K
& \textbf{19.08K} \\

Avg. frames per clip
& 24
& 2,035
& 323
& $\sim$300
& \begin{tabular}{@{}c@{}}\textbf{425 (multi-object)} \\ \textbf{106 (single-object)}\end{tabular} \\

Total frames
& 264K
& 216K
& 169.2K
& 1.4M
& \textbf{5.07M} \\

\hline

Dense 3D tracking
& $\checkmark^\dagger$
& \xmark
& \xmark
& \cmark
& \textbf{\cmark} \\

Non-rigid motion
& \xmark
& \cmark
& \cmark
& \cmark
& \textbf{\cmark} \\

Compressed representation
& $\checkmark^\dagger$
& \xmark
& \xmark
& \xmark
& \textbf{\cmark} \\

\hline
\end{tabular}%
}
\end{table*}

\paragraph{Real-world 3D tracking datasets.}
TAPVid-3D~\citep{tapvid3d} curates an evaluation benchmark with sparse metric 3D trajectories across three real-world domains: egocentric indoor activities from Aria Digital Twin (ADT)~\citep{pan2023adt} using motion capture and scanned meshes, autonomous driving from DriveTrack~\citep{drivetrack} using LiDAR and tracked 3D boxes, and multiview human motion from Panoptic Studio (PStudio)~\citep{joo2015panoptic} reconstructed with dynamic 3D Gaussians~\citep{dynamic3dgaussians}.
At a much larger scale for training, Stereo4D~\citep{stereo4d} reconstructs over 100K real-world 4D clips from Internet stereoscopic videos by combining stereo depth, camera pose estimation, and lifted 2D point tracks.
While these datasets offer realistic appearance and motion, their 3D trajectories rely on specialized capture setups or upstream estimators, limiting density, completeness under occlusion, and geometric accuracy compared with synthetic ground truth.

\paragraph{Synthetic 3D tracking datasets.}
Kubric's MOVi-E benchmark~\cite{kubric} provides depth, segmentation, optical flow, and scene flow for rigid-body simulations.
PointOdyssey~\cite{pointodyssey} introduces longer sequences with articulated human, animal, and object motion, providing depth, camera parameters, and sampled 2D/3D point tracks.
Dynamic Replica~\cite{dynamicstereo} combines animated humans and animals with scanned indoor environments, providing stereo depth, optical flow, segmentation, and sparse long-range trajectories.
Syn4D~\cite{syn4d} provides dense multiview 3D tracking by associating each pixel with a mesh triangle, barycentric coordinates, and per-frame animated-mesh vertex positions (Table~\ref{tab:dataset_comparison}).
Rather than storing uncompressed mesh vertex trajectories, RigIDTrack groups surface elements into rigidity-regularized affine bodies and stores a pixelwise body-ID map alongside per-body transformation trajectories, recovering dense 3D correspondences via Eq.~\ref{eq:rigidtrack_transport}.

\section{Additional Experiment Details and Results}
\label{sec:supp_evaluation}

We detail benchmark preparation, alignment protocols, and evaluation metrics for multiview reconstruction and 3D tracking, followed by per-benchmark quantitative results and additional qualitative comparisons.
All random sampling used to construct evaluation inputs uses fixed seeds so that every method receives identical inputs.
Metrics are averaged across clips within each benchmark, and the main-paper summary tables report the unweighted mean across the six benchmarks.

\subsection{Benchmark Preparation}
\label{sec:supp_eval_data}

\paragraph{Multiview reconstruction.}
We evaluate on the six benchmarks in Section~\ref{sec:multiview_reconstruction} (clip counts in Table~\ref{tab:supp_eval_clips}), with up to 32 unposed RGB frames per clip.
For long Bonn, KITTI and ScanNet sequences, we sample temporal intervals of at most 30 original frames while maintaining at least 50\% view overlap between consecutive selected frames.
\begin{itemize}\setlength{\itemsep}{2pt}
	\item \textbf{7-Scenes:} The released Kinect-v1 depth is unregistered to the RGB sensor. We forward-warp each depth map into the RGB camera frame using a calibrated transformation.
	\item \textbf{ETH3D:} Following the Depth Anything 3 evaluation protocol~\citep{depthanything3}, we discard images with a $90^\circ$ roll relative to the rest of the scene before sampling input views.
	\item \textbf{Sintel:} We exclude sequence \texttt{temple\_3} where the content is a moving foreground against a sky background, as camera motion is ambiguous without a static background reference.
	\item \textbf{ScanNet:} We evaluate all 100 test clips using the 32 preselected frames per clip.
\end{itemize}

\paragraph{3D tracking.}
We evaluate on the test splits of DynamicReplica, PointOdyssey, Syn4D, RigIDTrack, ADT, and PStudio (Table~\ref{tab:supp_eval_clips}), using 32 frames per clip sampled at 5 FPS and querying points visible in the first frame.
For datasets with explicit trajectory annotations, all valid first-frame query points are evaluated; for RigIDTrack, we sample up to 1,024 point trajectories per clip (weighting dynamic pixels $5\times$ higher than static pixels).

\begin{table}[!tbp]
	\centering
	\caption{Evaluation clip counts per benchmark.}
	\label{tab:supp_eval_clips}
	\small
	\begin{tabular}{@{}lr@{\hspace{2em}}lr@{}}
		\toprule
		\multicolumn{2}{c}{Multiview reconstruction} & \multicolumn{2}{c}{3D tracking} \\
		\cmidrule(lr){1-2}\cmidrule(lr){3-4}
		Benchmark & Clips & Benchmark & Clips \\
		\midrule
		7-Scenes & 46 & DynamicReplica & 20 \\
		Bonn & 26 & PointOdyssey & 13 \\
		ETH3D & 13 & Syn4D & 512 \\
		KITTI & 27 & ADT & 50 \\
		Sintel & 22 & PStudio & 50 \\
				ScanNet & 100 & RigIDTrack & 88 \\
		\midrule
				Total & 234 & Total & 733 \\
		\bottomrule
	\end{tabular}
\end{table}

\subsection{Multiview Reconstruction Metrics}
\label{sec:supp_recon_metrics}

\paragraph{Notation and accuracy summaries.}
Let $d_{tu}$ and $K_t$ denote ground-truth depth and intrinsics for pixel $u$ in frame $t$, and let $E_t=(R_t,\boldsymbol{t}_t)$ denote the world-to-camera extrinsics; hats denote predictions.
Back-projection yields camera-space points $\boldsymbol{x}_{tu}=d_{tu}K_t^{-1}\bar{\boldsymbol{u}}$ and world-space points $\boldsymbol{p}_{tu}=R_t^\top(\boldsymbol{x}_{tu}-\boldsymbol{t}_t)$, where $\bar{\boldsymbol{u}}$ is the homogeneous pixel coordinate and $r_{tu}=\|\boldsymbol{x}_{tu}\|$ is the distance to camera $t$.
For nonnegative relative errors $e$ over an evaluation set $\mathcal{V}$, threshold accuracy and area under the curve are defined as
\begin{equation}
	\delta_\tau=\operatorname{mean}_{\mathcal{V}}\boldsymbol{1}[e<\tau],
	\qquad
	\mathrm{AUC}_\tau=\frac{1}{\tau}\int_0^\tau\delta_\alpha\,d\alpha
	=\operatorname{mean}_{\mathcal{V}}\max(1-e/\tau,0),
	\label{eq:eval_auc}
\end{equation}
where subscripts 5 and 25 denote $\tau=0.05$ and $0.25$.
We report $\delta$, AUC, and OA as percentages; clip-level means pool all valid pixels or point-time pairs across the clip.

\paragraph{Depth alignment and error.}
We fit a single sequence-level scale $s_d$ or affine transformation $(s_a,b_a)$ over a sampled subset $\mathcal{F}$ of valid ground-truth pixels across the clip:
\begin{equation}
	s_d=\arg\min_s\sum_{(t,u)\in\mathcal{F}}\frac{|s\hat d_{tu}-d_{tu}|}{d_{tu}},
	\qquad
	(s_a,b_a)=\arg\min_{s,b}\sum_{(t,u)\in\mathcal{F}}\frac{|s\hat d_{tu}+b-d_{tu}|}{d_{tu}}.
	\label{eq:eval_depth_alignment}
\end{equation}
Setting $\tilde d=s_d\hat d$ (scale-aligned) or $\tilde d=s_a\hat d+b_a$ (affine-aligned), we evaluate relative depth error over all valid pixels $\mathcal{V}$:
\begin{equation}
	e^{\mathrm{depth}}_{tu}=\frac{|\tilde d_{tu}-d_{tu}|}{d_{tu}},
	\qquad
	\mathrm{Rel}=\operatorname{mean}_{\mathcal{V}}e^{\mathrm{depth}}_{tu},
\end{equation}
and report Rel and $\delta_{25}$ under both alignments.

\paragraph{Camera poses.}
Translation errors are normalized by the ground-truth scene scale $\ell$, defined as the harmonic mean of valid depths.
For each unordered frame pair $i<j$ with relative pose $E_{ij}=E_jE_i^{-1}=(R_{ij},\boldsymbol{t}_{ij})$, relative rotation error (in degrees) and relative translation error (scaled by $s_d$ from Eq.~\ref{eq:eval_depth_alignment}) are
\begin{equation}
	\mathrm{RPE\text{-}R}=\operatorname{mean}_{ij}\operatorname{angle}(R_{ij}\hat R_{ij}^{\top}),
	\qquad
	\mathrm{RPE\text{-}t}=\operatorname{mean}_{ij}\frac{\|s_d\hat{\boldsymbol{t}}_{ij}-\boldsymbol{t}_{ij}\|}{\ell}.
\end{equation}
For absolute trajectory error (ATE), let $\boldsymbol{c}_t=-R_t^\top\boldsymbol{t}_t$ denote the camera center and let $S_c\in\mathrm{Sim}(3)$ be the least-squares similarity alignment between predicted and ground-truth trajectories:
\begin{equation}
	\mathrm{ATE}=\operatorname{mean}_t\frac{\|S_c(\hat{\boldsymbol{c}}_t)-\boldsymbol{c}_t\|}{\ell}.
\end{equation}

\paragraph{Cross-view matching.}
To evaluate cross-view geometric consistency without per-view alignment, we sample up to $10^6$ candidate pixel pairs $(i,u)$ and $(j,v)$ per clip using ground-truth depth and poses, retaining pairs whose reprojected depth matches the target depth within 0.3\% relative error.
We measure the depth-normalized world-space discrepancy:
\begin{equation}
	e^{\mathrm{match}}_{iu,jv}
	=\frac{s_d\|\hat{\boldsymbol{p}}_{iu}-\hat{\boldsymbol{p}}_{jv}\|}{\max(d_{iu},d_{jv})},
\end{equation}
reporting mean relative distance (RD), $\delta_5$, and $\mathrm{AUC}_5$.

\paragraph{Global points.}
We align predicted and ground-truth world points via a sequence-level similarity transform $S_g\in\mathrm{Sim}(3)$:
\begin{equation}
	e^{\mathrm{global}}_{tu}=\frac{\|S_g(\hat{\boldsymbol{p}}_{tu})-\boldsymbol{p}_{tu}\|}{r_{tu}},
	\label{eq:eval_global_error}
\end{equation}
where $S_g$ is estimated from sampled correspondences via RANSAC (normalizing residuals by $r_{tu}$), followed by weighted least-squares refitting ($w\propto r_{tu}^{-2}$) on the lowest-error 80\% of inlier weight.
We report mean RD, $\delta_{25}$, and $\mathrm{AUC}_{25}$.

\subsection{3D Tracking Metrics}
\label{sec:supp_tracking_metrics}

\paragraph{Evaluation sets and alignment.}
For first-frame query point $q$ and target frame $t$, let $\boldsymbol{x}_{tq}$ and $\boldsymbol{p}_{tq}$ denote its ground-truth camera- and world-space coordinates, with $r_{tq}=\|\boldsymbol{x}_{tq}\|$.
We evaluate on two sets: \emph{all} (all valid point-time pairs, including occluded points) and \emph{visible} (point-time pairs visible in frame $t$), both including $t=0$.
Camera-space predictions are aligned by a single scale fitted on first-frame geometry:
\begin{equation}
	s_{\mathrm{track}}=\arg\min_s\sum_{q:(0,q)\in\mathcal{V}}\frac{\|s\hat{\boldsymbol{x}}_{0q}-\boldsymbol{x}_{0q}\|_1}{r_{0q}},
\end{equation}
whereas world-space predictions are aligned by a sequence-level similarity transform $S_{\mathrm{track}}\in\mathrm{Sim}(3)$ fitted over all selected point-time pairs using the same robust estimator as global-point reconstruction.

\paragraph{Relative endpoint error and AUC.}
With $\tilde{\boldsymbol{x}}_{tq}=s_{\mathrm{track}}\hat{\boldsymbol{x}}_{tq}$ and $\tilde{\boldsymbol{p}}_{tq}=S_{\mathrm{track}}(\hat{\boldsymbol{p}}_{tq})$, relative endpoint errors are
\begin{equation}
	e^{\mathrm{cam}}_{tq}=\frac{\|\tilde{\boldsymbol{x}}_{tq}-\boldsymbol{x}_{tq}\|}{r_{tq}},
	\qquad
	e^{\mathrm{world}}_{tq}=\frac{\|\tilde{\boldsymbol{p}}_{tq}-\boldsymbol{p}_{tq}\|}{r_{tq}}.
\end{equation}
We report the mean error (RelEPE) and $\mathrm{AUC}_{25}$ (Eq.~\ref{eq:eval_auc} with $\tau=0.25$) for each coordinate space and evaluation set.

\paragraph{Occlusion accuracy.}
Let $v_{tq}\in\{0,1\}$ denote ground-truth visibility and $\hat v_{tq}$ the predicted visibility probability (for RigIDTrack, $v_{tq}=1$ when the transported point projects inside frame $t$ within a 3\% depth tolerance of $D_t$).
Thresholding at $0.5$ over all valid point-time pairs $\mathcal{D}$ gives
\begin{equation}
	\mathrm{OA}=\operatorname{mean}_{(t,q)\in\mathcal{D}}\boldsymbol{1}\!\left[\boldsymbol{1}[\hat v_{tq}\geq0.5]=v_{tq}\right],
\end{equation}
reported for methods that predict visibility (Open-D4RT, Track4World, and MoGe-4D).

\subsection{Per-Benchmark Results}
\label{sec:supp_per_benchmark}

Tables~\ref{tab:supp_depth}--\ref{tab:supp_global_points} report depth, camera, matching, and global-point results for each reconstruction benchmark.
Tables~\ref{tab:supp_tracking_camera} and~\ref{tab:supp_tracking_world} separate camera- and world-space tracking, and Table~\ref{tab:supp_occlusion} reports OA.
Methods and model variants follow Table~\ref{tab:multiview_reconstruction}; Open-D4RT denotes the unofficial D4RT implementation throughout.

\begin{table*}[!tbp]
\centering
\caption{Per-benchmark depth accuracy ($\delta_{25}\uparrow$, $\mathrm{Rel}\downarrow$).}
\label{tab:supp_depth}
\begingroup
\fontsize{7pt}{8.4pt}\selectfont
\newcommand{\supplementheaderfont}{\fontsize{6pt}{7.2pt}\selectfont}
\renewcommand{\scriptsize}{\supplementheaderfont}
\definecolor{supplementhighlight}{RGB}{255,215,0}
\setlength{\tabcolsep}{1.5pt}
\setlength{\aboverulesep}{0pt}
\setlength{\belowrulesep}{0pt}
\renewcommand{\arraystretch}{1.15}
\resizebox{\textwidth}{!}{%
\begin{tabular}{@{}lrrrr@{\hspace{4pt}}rrrr@{\hspace{4pt}}rrrr@{\hspace{4pt}}rrrr@{\hspace{4pt}}rrrr@{\hspace{4pt}}rrrr@{}}
\toprule
\multirow{3}{*}{\supplementheaderfont Method} & \multicolumn{4}{c}{\supplementheaderfont 7-Scenes} & \multicolumn{4}{c}{\supplementheaderfont Bonn} & \multicolumn{4}{c}{\supplementheaderfont ETH3D} & \multicolumn{4}{c}{\supplementheaderfont KITTI} & \multicolumn{4}{c}{\supplementheaderfont Sintel} & \multicolumn{4}{c}{\supplementheaderfont ScanNet} \\
\cmidrule(lr){2-5}\cmidrule(lr){6-9}\cmidrule(lr){10-13}\cmidrule(lr){14-17}\cmidrule(lr){18-21}\cmidrule(lr){22-25}
 & \multicolumn{2}{c}{\supplementheaderfont Affine} & \multicolumn{2}{c}{\supplementheaderfont Scale} & \multicolumn{2}{c}{\supplementheaderfont Affine} & \multicolumn{2}{c}{\supplementheaderfont Scale} & \multicolumn{2}{c}{\supplementheaderfont Affine} & \multicolumn{2}{c}{\supplementheaderfont Scale} & \multicolumn{2}{c}{\supplementheaderfont Affine} & \multicolumn{2}{c}{\supplementheaderfont Scale} & \multicolumn{2}{c}{\supplementheaderfont Affine} & \multicolumn{2}{c}{\supplementheaderfont Scale} & \multicolumn{2}{c}{\supplementheaderfont Affine} & \multicolumn{2}{c}{\supplementheaderfont Scale} \\
\cmidrule(lr){2-3}\cmidrule(lr){4-5}\cmidrule(lr){6-7}\cmidrule(lr){8-9}\cmidrule(lr){10-11}\cmidrule(lr){12-13}\cmidrule(lr){14-15}\cmidrule(lr){16-17}\cmidrule(lr){18-19}\cmidrule(lr){20-21}\cmidrule(lr){22-23}\cmidrule(lr){24-25}
 & {\supplementheaderfont $\delta_{25}\uparrow$} & {\supplementheaderfont Rel$\downarrow$} & {\supplementheaderfont $\delta_{25}\uparrow$} & {\supplementheaderfont Rel$\downarrow$} & {\supplementheaderfont $\delta_{25}\uparrow$} & {\supplementheaderfont Rel$\downarrow$} & {\supplementheaderfont $\delta_{25}\uparrow$} & {\supplementheaderfont Rel$\downarrow$} & {\supplementheaderfont $\delta_{25}\uparrow$} & {\supplementheaderfont Rel$\downarrow$} & {\supplementheaderfont $\delta_{25}\uparrow$} & {\supplementheaderfont Rel$\downarrow$} & {\supplementheaderfont $\delta_{25}\uparrow$} & {\supplementheaderfont Rel$\downarrow$} & {\supplementheaderfont $\delta_{25}\uparrow$} & {\supplementheaderfont Rel$\downarrow$} & {\supplementheaderfont $\delta_{25}\uparrow$} & {\supplementheaderfont Rel$\downarrow$} & {\supplementheaderfont $\delta_{25}\uparrow$} & {\supplementheaderfont Rel$\downarrow$} & {\supplementheaderfont $\delta_{25}\uparrow$} & {\supplementheaderfont Rel$\downarrow$} & {\supplementheaderfont $\delta_{25}\uparrow$} & {\supplementheaderfont Rel$\downarrow$} \\
\midrule
MonST3R & \cellcolor{supplementhighlight!0}96.40 & \cellcolor{supplementhighlight!0}0.066 & \cellcolor{supplementhighlight!0}96.48 & \cellcolor{supplementhighlight!0}0.069 & \cellcolor{supplementhighlight!0}96.69 & \cellcolor{supplementhighlight!0}0.051 & \cellcolor{supplementhighlight!0}95.80 & \cellcolor{supplementhighlight!1}0.054 & \cellcolor{supplementhighlight!0}91.31 & \cellcolor{supplementhighlight!0}0.098 & \cellcolor{supplementhighlight!0}90.64 & \cellcolor{supplementhighlight!0}0.101 & \cellcolor{supplementhighlight!0}68.52 & \cellcolor{supplementhighlight!0}0.202 & \cellcolor{supplementhighlight!0}63.03 & \cellcolor{supplementhighlight!0}0.230 & \cellcolor{supplementhighlight!0}66.58 & \cellcolor{supplementhighlight!0}0.271 & \cellcolor{supplementhighlight!0}51.54 & \cellcolor{supplementhighlight!0}0.349 & \cellcolor{supplementhighlight!0}97.21 & \cellcolor{supplementhighlight!0}0.056 & \cellcolor{supplementhighlight!0}97.04 & \cellcolor{supplementhighlight!0}0.059 \\
MegaSAM & \cellcolor{supplementhighlight!0}95.51 & \cellcolor{supplementhighlight!0}0.059 & \cellcolor{supplementhighlight!0}94.61 & \cellcolor{supplementhighlight!0}0.064 & \cellcolor{supplementhighlight!0}96.77 & \cellcolor{supplementhighlight!0}0.057 & \cellcolor{supplementhighlight!0}93.72 & \cellcolor{supplementhighlight!0}0.073 & \cellcolor{supplementhighlight!0}66.86 & \cellcolor{supplementhighlight!0}0.207 & \cellcolor{supplementhighlight!0}58.56 & \cellcolor{supplementhighlight!0}0.262 & \cellcolor{supplementhighlight!0}84.92 & \cellcolor{supplementhighlight!0}0.122 & \cellcolor{supplementhighlight!0}83.59 & \cellcolor{supplementhighlight!0}0.130 & \cellcolor{supplementhighlight!2}79.24 & \cellcolor{supplementhighlight!1}0.197 & \cellcolor{supplementhighlight!1}72.00 & \cellcolor{supplementhighlight!1}0.232 & \cellcolor{supplementhighlight!0}91.69 & \cellcolor{supplementhighlight!0}0.088 & \cellcolor{supplementhighlight!0}89.69 & \cellcolor{supplementhighlight!0}0.096 \\
VGGT & \cellcolor{supplementhighlight!0}96.65 & \cellcolor{supplementhighlight!7}0.040 & \cellcolor{supplementhighlight!0}96.66 & \cellcolor{supplementhighlight!7}0.041 & \cellcolor{supplementhighlight!2}98.58 & \cellcolor{supplementhighlight!1}0.037 & \cellcolor{supplementhighlight!1}96.01 & \cellcolor{supplementhighlight!4}0.047 & \cellcolor{supplementhighlight!4}98.92 & \cellcolor{supplementhighlight!10}0.036 & \cellcolor{supplementhighlight!4}99.15 & \cellcolor{supplementhighlight!7}0.038 & \cellcolor{supplementhighlight!14}98.01 & \cellcolor{supplementhighlight!7}0.054 & \cellcolor{supplementhighlight!10}97.57 & \cellcolor{supplementhighlight!4}0.060 & \cellcolor{supplementhighlight!7}80.14 & \cellcolor{supplementhighlight!10}0.169 & \cellcolor{supplementhighlight!14}76.25 & \cellcolor{supplementhighlight!7}0.195 & \cellcolor{supplementhighlight!66}99.14 & \cellcolor{supplementhighlight!43}0.024 & \cellcolor{supplementhighlight!66}99.14 & \cellcolor{supplementhighlight!54}0.027 \\
MapAnything & \cellcolor{supplementhighlight!10}97.57 & \cellcolor{supplementhighlight!4}0.042 & \cellcolor{supplementhighlight!10}97.57 & \cellcolor{supplementhighlight!4}0.043 & \cellcolor{supplementhighlight!20}99.07 & \cellcolor{supplementhighlight!2}0.036 & \cellcolor{supplementhighlight!43}98.31 & \cellcolor{supplementhighlight!20}0.042 & \cellcolor{supplementhighlight!10}99.04 & \cellcolor{supplementhighlight!7}0.037 & \cellcolor{supplementhighlight!7}99.25 & \cellcolor{supplementhighlight!10}0.038 & \cellcolor{supplementhighlight!1}95.46 & \cellcolor{supplementhighlight!1}0.079 & \cellcolor{supplementhighlight!1}95.31 & \cellcolor{supplementhighlight!1}0.080 & \cellcolor{supplementhighlight!0}70.44 & \cellcolor{supplementhighlight!0}0.264 & \cellcolor{supplementhighlight!0}62.64 & \cellcolor{supplementhighlight!0}0.300 & \cellcolor{supplementhighlight!7}98.57 & \cellcolor{supplementhighlight!4}0.041 & \cellcolor{supplementhighlight!7}98.47 & \cellcolor{supplementhighlight!2}0.046 \\
$\pi^3$ & \cellcolor{supplementhighlight!80}98.65 & \cellcolor{supplementhighlight!80}0.026 & \cellcolor{supplementhighlight!80}98.68 & \cellcolor{supplementhighlight!66}0.031 & \cellcolor{supplementhighlight!43}99.15 & \cellcolor{supplementhighlight!80}0.026 & \cellcolor{supplementhighlight!26}97.20 & \cellcolor{supplementhighlight!80}0.032 & \cellcolor{supplementhighlight!43}99.80 & \cellcolor{supplementhighlight!43}0.021 & \cellcolor{supplementhighlight!43}99.81 & \cellcolor{supplementhighlight!34}0.022 & \cellcolor{supplementhighlight!66}99.11 & \cellcolor{supplementhighlight!54}0.037 & \cellcolor{supplementhighlight!66}99.08 & \cellcolor{supplementhighlight!54}0.038 & \cellcolor{supplementhighlight!34}83.85 & \cellcolor{supplementhighlight!34}0.132 & \cellcolor{supplementhighlight!10}75.93 & \cellcolor{supplementhighlight!20}0.189 & \cellcolor{supplementhighlight!80}99.27 & \cellcolor{supplementhighlight!80}0.020 & \cellcolor{supplementhighlight!80}99.27 & \cellcolor{supplementhighlight!80}0.021 \\
Depth Anything 3 & \cellcolor{supplementhighlight!14}97.62 & \cellcolor{supplementhighlight!20}0.033 & \cellcolor{supplementhighlight!14}97.61 & \cellcolor{supplementhighlight!20}0.034 & \cellcolor{supplementhighlight!1}98.28 & \cellcolor{supplementhighlight!7}0.036 & \cellcolor{supplementhighlight!2}96.10 & \cellcolor{supplementhighlight!7}0.047 & \cellcolor{supplementhighlight!1}98.04 & \cellcolor{supplementhighlight!34}0.027 & \cellcolor{supplementhighlight!20}99.60 & \cellcolor{supplementhighlight!43}0.020 & \cellcolor{supplementhighlight!20}98.08 & \cellcolor{supplementhighlight!14}0.052 & \cellcolor{supplementhighlight!20}97.83 & \cellcolor{supplementhighlight!10}0.057 & \cellcolor{supplementhighlight!20}82.55 & \cellcolor{supplementhighlight!20}0.151 & \cellcolor{supplementhighlight!26}76.49 & \cellcolor{supplementhighlight!26}0.181 & \cellcolor{supplementhighlight!4}98.29 & \cellcolor{supplementhighlight!10}0.033 & \cellcolor{supplementhighlight!4}98.24 & \cellcolor{supplementhighlight!14}0.035 \\
LingBot-Map & \cellcolor{supplementhighlight!1}97.40 & \cellcolor{supplementhighlight!1}0.044 & \cellcolor{supplementhighlight!2}97.39 & \cellcolor{supplementhighlight!1}0.045 & \cellcolor{supplementhighlight!0}98.07 & \cellcolor{supplementhighlight!0}0.046 & \cellcolor{supplementhighlight!0}95.46 & \cellcolor{supplementhighlight!0}0.056 & \cellcolor{supplementhighlight!1}97.91 & \cellcolor{supplementhighlight!1}0.047 & \cellcolor{supplementhighlight!1}97.79 & \cellcolor{supplementhighlight!1}0.048 & \cellcolor{supplementhighlight!4}97.03 & \cellcolor{supplementhighlight!2}0.063 & \cellcolor{supplementhighlight!2}96.77 & \cellcolor{supplementhighlight!2}0.065 & \cellcolor{supplementhighlight!1}78.82 & \cellcolor{supplementhighlight!2}0.184 & \cellcolor{supplementhighlight!4}73.20 & \cellcolor{supplementhighlight!2}0.210 & \cellcolor{supplementhighlight!1}98.24 & \cellcolor{supplementhighlight!2}0.043 & \cellcolor{supplementhighlight!1}98.08 & \cellcolor{supplementhighlight!4}0.045 \\
VGGT-Omega & \cellcolor{supplementhighlight!26}97.82 & \cellcolor{supplementhighlight!43}0.032 & \cellcolor{supplementhighlight!26}97.82 & \cellcolor{supplementhighlight!34}0.033 & \cellcolor{supplementhighlight!4}98.60 & \cellcolor{supplementhighlight!20}0.034 & \cellcolor{supplementhighlight!7}96.22 & \cellcolor{supplementhighlight!14}0.042 & \cellcolor{supplementhighlight!80}99.88 & \cellcolor{supplementhighlight!80}0.010 & \cellcolor{supplementhighlight!80}99.88 & \cellcolor{supplementhighlight!80}0.010 & \cellcolor{supplementhighlight!80}99.19 & \cellcolor{supplementhighlight!80}0.032 & \cellcolor{supplementhighlight!80}99.17 & \cellcolor{supplementhighlight!80}0.033 & \cellcolor{supplementhighlight!80}96.67 & \cellcolor{supplementhighlight!66}0.088 & \cellcolor{supplementhighlight!80}96.75 & \cellcolor{supplementhighlight!80}0.091 & \cellcolor{supplementhighlight!10}98.61 & \cellcolor{supplementhighlight!20}0.027 & \cellcolor{supplementhighlight!10}98.58 & \cellcolor{supplementhighlight!20}0.031 \\
$\pi^3$ \scriptsize(DINO, our data) & \cellcolor{supplementhighlight!34}97.92 & \cellcolor{supplementhighlight!2}0.043 & \cellcolor{supplementhighlight!34}97.93 & \cellcolor{supplementhighlight!2}0.044 & \cellcolor{supplementhighlight!66}99.21 & \cellcolor{supplementhighlight!26}0.031 & \cellcolor{supplementhighlight!66}99.21 & \cellcolor{supplementhighlight!54}0.035 & \cellcolor{supplementhighlight!20}99.39 & \cellcolor{supplementhighlight!4}0.037 & \cellcolor{supplementhighlight!10}99.38 & \cellcolor{supplementhighlight!4}0.040 & \cellcolor{supplementhighlight!26}98.40 & \cellcolor{supplementhighlight!26}0.049 & \cellcolor{supplementhighlight!26}98.38 & \cellcolor{supplementhighlight!26}0.050 & \cellcolor{supplementhighlight!26}83.28 & \cellcolor{supplementhighlight!26}0.136 & \cellcolor{supplementhighlight!34}78.30 & \cellcolor{supplementhighlight!43}0.168 & \cellcolor{supplementhighlight!26}98.92 & \cellcolor{supplementhighlight!7}0.037 & \cellcolor{supplementhighlight!20}98.73 & \cellcolor{supplementhighlight!7}0.041 \\
\midrule
Any4D & \cellcolor{supplementhighlight!4}97.47 & \cellcolor{supplementhighlight!0}0.050 & \cellcolor{supplementhighlight!4}97.44 & \cellcolor{supplementhighlight!0}0.053 & \cellcolor{supplementhighlight!14}99.00 & \cellcolor{supplementhighlight!4}0.036 & \cellcolor{supplementhighlight!34}98.03 & \cellcolor{supplementhighlight!10}0.045 & \cellcolor{supplementhighlight!2}98.85 & \cellcolor{supplementhighlight!1}0.053 & \cellcolor{supplementhighlight!1}98.44 & \cellcolor{supplementhighlight!1}0.059 & \cellcolor{supplementhighlight!2}96.99 & \cellcolor{supplementhighlight!1}0.067 & \cellcolor{supplementhighlight!4}96.80 & \cellcolor{supplementhighlight!1}0.069 & \cellcolor{supplementhighlight!1}76.93 & \cellcolor{supplementhighlight!1}0.200 & \cellcolor{supplementhighlight!1}71.82 & \cellcolor{supplementhighlight!1}0.230 & \cellcolor{supplementhighlight!2}98.26 & \cellcolor{supplementhighlight!1}0.050 & \cellcolor{supplementhighlight!2}98.15 & \cellcolor{supplementhighlight!1}0.054 \\
Open-D4RT & \cellcolor{supplementhighlight!7}97.57 & \cellcolor{supplementhighlight!1}0.048 & \cellcolor{supplementhighlight!7}97.51 & \cellcolor{supplementhighlight!1}0.049 & \cellcolor{supplementhighlight!7}98.77 & \cellcolor{supplementhighlight!1}0.040 & \cellcolor{supplementhighlight!1}96.06 & \cellcolor{supplementhighlight!0}0.058 & \cellcolor{supplementhighlight!0}84.78 & \cellcolor{supplementhighlight!0}0.145 & \cellcolor{supplementhighlight!0}78.46 & \cellcolor{supplementhighlight!0}0.177 & \cellcolor{supplementhighlight!0}95.14 & \cellcolor{supplementhighlight!0}0.089 & \cellcolor{supplementhighlight!0}94.76 & \cellcolor{supplementhighlight!0}0.093 & \cellcolor{supplementhighlight!0}72.65 & \cellcolor{supplementhighlight!0}0.220 & \cellcolor{supplementhighlight!0}64.19 & \cellcolor{supplementhighlight!0}0.262 & \cellcolor{supplementhighlight!1}98.19 & \cellcolor{supplementhighlight!0}0.050 & \cellcolor{supplementhighlight!1}97.91 & \cellcolor{supplementhighlight!0}0.057 \\
V-DPM & \cellcolor{supplementhighlight!2}97.41 & \cellcolor{supplementhighlight!10}0.039 & \cellcolor{supplementhighlight!1}97.13 & \cellcolor{supplementhighlight!10}0.040 & \cellcolor{supplementhighlight!54}99.16 & \cellcolor{supplementhighlight!14}0.034 & \cellcolor{supplementhighlight!10}96.72 & \cellcolor{supplementhighlight!1}0.049 & \cellcolor{supplementhighlight!7}98.96 & \cellcolor{supplementhighlight!2}0.042 & \cellcolor{supplementhighlight!2}98.75 & \cellcolor{supplementhighlight!2}0.045 & \cellcolor{supplementhighlight!7}97.81 & \cellcolor{supplementhighlight!10}0.052 & \cellcolor{supplementhighlight!14}97.80 & \cellcolor{supplementhighlight!14}0.054 & \cellcolor{supplementhighlight!4}79.94 & \cellcolor{supplementhighlight!4}0.170 & \cellcolor{supplementhighlight!2}73.16 & \cellcolor{supplementhighlight!4}0.209 & \cellcolor{supplementhighlight!34}99.07 & \cellcolor{supplementhighlight!26}0.027 & \cellcolor{supplementhighlight!34}99.07 & \cellcolor{supplementhighlight!26}0.031 \\
Track4World & \cellcolor{supplementhighlight!20}97.72 & \cellcolor{supplementhighlight!14}0.033 & \cellcolor{supplementhighlight!20}97.72 & \cellcolor{supplementhighlight!14}0.035 & \cellcolor{supplementhighlight!1}98.49 & \cellcolor{supplementhighlight!10}0.035 & \cellcolor{supplementhighlight!4}96.14 & \cellcolor{supplementhighlight!2}0.048 & \cellcolor{supplementhighlight!14}99.15 & \cellcolor{supplementhighlight!26}0.028 & \cellcolor{supplementhighlight!14}99.44 & \cellcolor{supplementhighlight!26}0.026 & \cellcolor{supplementhighlight!10}97.92 & \cellcolor{supplementhighlight!4}0.055 & \cellcolor{supplementhighlight!7}97.54 & \cellcolor{supplementhighlight!7}0.058 & \cellcolor{supplementhighlight!10}80.83 & \cellcolor{supplementhighlight!14}0.168 & \cellcolor{supplementhighlight!20}76.48 & \cellcolor{supplementhighlight!10}0.193 & \cellcolor{supplementhighlight!0}98.00 & \cellcolor{supplementhighlight!1}0.044 & \cellcolor{supplementhighlight!0}97.81 & \cellcolor{supplementhighlight!1}0.048 \\
4RC & \cellcolor{supplementhighlight!1}97.38 & \cellcolor{supplementhighlight!26}0.033 & \cellcolor{supplementhighlight!1}97.37 & \cellcolor{supplementhighlight!26}0.034 & \cellcolor{supplementhighlight!10}98.93 & \cellcolor{supplementhighlight!34}0.030 & \cellcolor{supplementhighlight!14}96.94 & \cellcolor{supplementhighlight!26}0.039 & \cellcolor{supplementhighlight!34}99.71 & \cellcolor{supplementhighlight!54}0.017 & \cellcolor{supplementhighlight!34}99.71 & \cellcolor{supplementhighlight!54}0.017 & \cellcolor{supplementhighlight!1}96.60 & \cellcolor{supplementhighlight!20}0.051 & \cellcolor{supplementhighlight!1}96.43 & \cellcolor{supplementhighlight!20}0.052 & \cellcolor{supplementhighlight!14}81.26 & \cellcolor{supplementhighlight!7}0.170 & \cellcolor{supplementhighlight!43}78.91 & \cellcolor{supplementhighlight!14}0.191 & \cellcolor{supplementhighlight!20}98.80 & \cellcolor{supplementhighlight!54}0.023 & \cellcolor{supplementhighlight!26}98.78 & \cellcolor{supplementhighlight!43}0.027 \\
\midrule
\textbf{Ours} \scriptsize(DINO) & \cellcolor{supplementhighlight!43}98.06 & \cellcolor{supplementhighlight!34}0.032 & \cellcolor{supplementhighlight!43}98.06 & \cellcolor{supplementhighlight!43}0.033 & \cellcolor{supplementhighlight!80}99.26 & \cellcolor{supplementhighlight!43}0.027 & \cellcolor{supplementhighlight!80}99.23 & \cellcolor{supplementhighlight!43}0.036 & \cellcolor{supplementhighlight!54}99.82 & \cellcolor{supplementhighlight!20}0.029 & \cellcolor{supplementhighlight!54}99.83 & \cellcolor{supplementhighlight!20}0.031 & \cellcolor{supplementhighlight!34}98.72 & \cellcolor{supplementhighlight!34}0.041 & \cellcolor{supplementhighlight!34}98.71 & \cellcolor{supplementhighlight!34}0.042 & \cellcolor{supplementhighlight!43}84.29 & \cellcolor{supplementhighlight!43}0.130 & \cellcolor{supplementhighlight!7}75.60 & \cellcolor{supplementhighlight!34}0.180 & \cellcolor{supplementhighlight!14}98.79 & \cellcolor{supplementhighlight!14}0.032 & \cellcolor{supplementhighlight!14}98.67 & \cellcolor{supplementhighlight!10}0.035 \\
\textbf{Ours} \scriptsize(VJEPA) & \cellcolor{supplementhighlight!54}98.11 & \cellcolor{supplementhighlight!54}0.031 & \cellcolor{supplementhighlight!54}98.12 & \cellcolor{supplementhighlight!54}0.032 & \cellcolor{supplementhighlight!26}99.08 & \cellcolor{supplementhighlight!54}0.027 & \cellcolor{supplementhighlight!54}99.05 & \cellcolor{supplementhighlight!66}0.033 & \cellcolor{supplementhighlight!26}99.63 & \cellcolor{supplementhighlight!14}0.031 & \cellcolor{supplementhighlight!26}99.64 & \cellcolor{supplementhighlight!14}0.033 & \cellcolor{supplementhighlight!54}99.02 & \cellcolor{supplementhighlight!66}0.035 & \cellcolor{supplementhighlight!54}99.03 & \cellcolor{supplementhighlight!66}0.036 & \cellcolor{supplementhighlight!54}89.44 & \cellcolor{supplementhighlight!54}0.104 & \cellcolor{supplementhighlight!54}82.58 & \cellcolor{supplementhighlight!54}0.140 & \cellcolor{supplementhighlight!43}99.12 & \cellcolor{supplementhighlight!34}0.026 & \cellcolor{supplementhighlight!43}99.10 & \cellcolor{supplementhighlight!34}0.029 \\
\textbf{Ours} \scriptsize(VGGT-$\Omega$) & \cellcolor{supplementhighlight!66}98.14 & \cellcolor{supplementhighlight!66}0.028 & \cellcolor{supplementhighlight!66}98.13 & \cellcolor{supplementhighlight!80}0.029 & \cellcolor{supplementhighlight!34}99.09 & \cellcolor{supplementhighlight!66}0.027 & \cellcolor{supplementhighlight!20}97.18 & \cellcolor{supplementhighlight!34}0.038 & \cellcolor{supplementhighlight!66}99.87 & \cellcolor{supplementhighlight!66}0.014 & \cellcolor{supplementhighlight!66}99.87 & \cellcolor{supplementhighlight!66}0.015 & \cellcolor{supplementhighlight!43}98.82 & \cellcolor{supplementhighlight!43}0.040 & \cellcolor{supplementhighlight!43}98.82 & \cellcolor{supplementhighlight!43}0.041 & \cellcolor{supplementhighlight!66}92.42 & \cellcolor{supplementhighlight!80}0.081 & \cellcolor{supplementhighlight!66}87.77 & \cellcolor{supplementhighlight!66}0.102 & \cellcolor{supplementhighlight!54}99.13 & \cellcolor{supplementhighlight!66}0.021 & \cellcolor{supplementhighlight!54}99.12 & \cellcolor{supplementhighlight!66}0.025 \\
\bottomrule
\end{tabular}
}
\endgroup
\end{table*}

\begin{table*}[!tbp]
\centering
\caption{Per-benchmark camera pose accuracy ($\mathrm{ATE}\downarrow$, $\text{RPE-R}\downarrow$, $\text{RPE-t}\downarrow$).}
\label{tab:supp_camera}
\begingroup
\fontsize{7pt}{8.4pt}\selectfont
\newcommand{\supplementheaderfont}{\fontsize{6pt}{7.2pt}\selectfont}
\renewcommand{\scriptsize}{\supplementheaderfont}
\definecolor{supplementhighlight}{RGB}{255,215,0}
\setlength{\tabcolsep}{1.5pt}
\setlength{\aboverulesep}{0pt}
\setlength{\belowrulesep}{0pt}
\renewcommand{\arraystretch}{1.15}
\resizebox{\textwidth}{!}{%
\begin{tabular}{@{}lrrr@{\hspace{4pt}}rrr@{\hspace{4pt}}rrr@{\hspace{4pt}}rrr@{\hspace{4pt}}rrr@{\hspace{4pt}}rrr@{}}
\toprule
\multirow{2}{*}{\supplementheaderfont Method} & \multicolumn{3}{c}{\supplementheaderfont 7-Scenes} & \multicolumn{3}{c}{\supplementheaderfont Bonn} & \multicolumn{3}{c}{\supplementheaderfont ETH3D} & \multicolumn{3}{c}{\supplementheaderfont KITTI} & \multicolumn{3}{c}{\supplementheaderfont Sintel} & \multicolumn{3}{c}{\supplementheaderfont ScanNet} \\
\cmidrule(lr){2-4}\cmidrule(lr){5-7}\cmidrule(lr){8-10}\cmidrule(lr){11-13}\cmidrule(lr){14-16}\cmidrule(lr){17-19}
 & {\supplementheaderfont ATE$\downarrow$} & {\supplementheaderfont RPE-R$\downarrow$} & {\supplementheaderfont RPE-t$\downarrow$} & {\supplementheaderfont ATE$\downarrow$} & {\supplementheaderfont RPE-R$\downarrow$} & {\supplementheaderfont RPE-t$\downarrow$} & {\supplementheaderfont ATE$\downarrow$} & {\supplementheaderfont RPE-R$\downarrow$} & {\supplementheaderfont RPE-t$\downarrow$} & {\supplementheaderfont ATE$\downarrow$} & {\supplementheaderfont RPE-R$\downarrow$} & {\supplementheaderfont RPE-t$\downarrow$} & {\supplementheaderfont ATE$\downarrow$} & {\supplementheaderfont RPE-R$\downarrow$} & {\supplementheaderfont RPE-t$\downarrow$} & {\supplementheaderfont ATE$\downarrow$} & {\supplementheaderfont RPE-R$\downarrow$} & {\supplementheaderfont RPE-t$\downarrow$} \\
\midrule
MonST3R & \cellcolor{supplementhighlight!0}0.171 & \cellcolor{supplementhighlight!0}18.87 & \cellcolor{supplementhighlight!0}0.278 & \cellcolor{supplementhighlight!0}0.027 & \cellcolor{supplementhighlight!0}4.559 & \cellcolor{supplementhighlight!1}0.052 & \cellcolor{supplementhighlight!0}0.409 & \cellcolor{supplementhighlight!0}48.03 & \cellcolor{supplementhighlight!0}0.708 & \cellcolor{supplementhighlight!0}0.160 & \cellcolor{supplementhighlight!0}1.637 & \cellcolor{supplementhighlight!0}0.566 & \cellcolor{supplementhighlight!1}0.023 & \cellcolor{supplementhighlight!0}5.073 & \cellcolor{supplementhighlight!0}0.120 & \cellcolor{supplementhighlight!0}0.076 & \cellcolor{supplementhighlight!0}11.16 & \cellcolor{supplementhighlight!0}0.164 \\
MegaSAM & \cellcolor{supplementhighlight!0}0.075 & \cellcolor{supplementhighlight!0}8.157 & \cellcolor{supplementhighlight!0}0.130 & \cellcolor{supplementhighlight!1}0.023 & \cellcolor{supplementhighlight!0}5.175 & \cellcolor{supplementhighlight!0}0.067 & \cellcolor{supplementhighlight!0}0.519 & \cellcolor{supplementhighlight!0}77.71 & \cellcolor{supplementhighlight!0}0.763 & \cellcolor{supplementhighlight!80}0.005 & \cellcolor{supplementhighlight!80}0.181 & \cellcolor{supplementhighlight!0}0.075 & \cellcolor{supplementhighlight!43}0.009 & \cellcolor{supplementhighlight!34}0.678 & \cellcolor{supplementhighlight!1}0.075 & \cellcolor{supplementhighlight!0}0.117 & \cellcolor{supplementhighlight!0}18.10 & \cellcolor{supplementhighlight!0}0.217 \\
VGGT & \cellcolor{supplementhighlight!20}0.033 & \cellcolor{supplementhighlight!14}3.680 & \cellcolor{supplementhighlight!14}0.069 & \cellcolor{supplementhighlight!14}0.014 & \cellcolor{supplementhighlight!80}1.244 & \cellcolor{supplementhighlight!14}0.030 & \cellcolor{supplementhighlight!2}0.048 & \cellcolor{supplementhighlight!7}1.544 & \cellcolor{supplementhighlight!2}0.086 & \cellcolor{supplementhighlight!2}0.017 & \cellcolor{supplementhighlight!14}0.469 & \cellcolor{supplementhighlight!10}0.046 & \cellcolor{supplementhighlight!2}0.020 & \cellcolor{supplementhighlight!10}0.928 & \cellcolor{supplementhighlight!4}0.061 & \cellcolor{supplementhighlight!26}0.024 & \cellcolor{supplementhighlight!80}1.790 & \cellcolor{supplementhighlight!26}0.046 \\
MapAnything & \cellcolor{supplementhighlight!4}0.042 & \cellcolor{supplementhighlight!7}3.814 & \cellcolor{supplementhighlight!4}0.079 & \cellcolor{supplementhighlight!1}0.024 & \cellcolor{supplementhighlight!1}2.153 & \cellcolor{supplementhighlight!1}0.050 & \cellcolor{supplementhighlight!14}0.040 & \cellcolor{supplementhighlight!14}1.476 & \cellcolor{supplementhighlight!10}0.067 & \cellcolor{supplementhighlight!4}0.016 & \cellcolor{supplementhighlight!1}0.948 & \cellcolor{supplementhighlight!20}0.043 & \cellcolor{supplementhighlight!0}0.042 & \cellcolor{supplementhighlight!1}2.038 & \cellcolor{supplementhighlight!1}0.109 & \cellcolor{supplementhighlight!2}0.039 & \cellcolor{supplementhighlight!1}4.352 & \cellcolor{supplementhighlight!1}0.086 \\
$\pi^3$ & \cellcolor{supplementhighlight!66}0.029 & \cellcolor{supplementhighlight!2}3.864 & \cellcolor{supplementhighlight!43}0.059 & \cellcolor{supplementhighlight!26}0.011 & \cellcolor{supplementhighlight!14}1.463 & \cellcolor{supplementhighlight!34}0.023 & \cellcolor{supplementhighlight!26}0.032 & \cellcolor{supplementhighlight!43}0.798 & \cellcolor{supplementhighlight!26}0.047 & \cellcolor{supplementhighlight!43}0.010 & \cellcolor{supplementhighlight!34}0.404 & \cellcolor{supplementhighlight!43}0.035 & \cellcolor{supplementhighlight!34}0.009 & \cellcolor{supplementhighlight!54}0.458 & \cellcolor{supplementhighlight!54}0.043 & \cellcolor{supplementhighlight!54}0.022 & \cellcolor{supplementhighlight!66}1.828 & \cellcolor{supplementhighlight!54}0.043 \\
Depth Anything 3 & \cellcolor{supplementhighlight!54}0.029 & \cellcolor{supplementhighlight!43}3.421 & \cellcolor{supplementhighlight!66}0.058 & \cellcolor{supplementhighlight!80}0.008 & \cellcolor{supplementhighlight!10}1.482 & \cellcolor{supplementhighlight!66}0.019 & \cellcolor{supplementhighlight!66}0.009 & \cellcolor{supplementhighlight!66}0.650 & \cellcolor{supplementhighlight!66}0.016 & \cellcolor{supplementhighlight!10}0.015 & \cellcolor{supplementhighlight!0}1.624 & \cellcolor{supplementhighlight!1}0.058 & \cellcolor{supplementhighlight!26}0.010 & \cellcolor{supplementhighlight!20}0.804 & \cellcolor{supplementhighlight!43}0.049 & \cellcolor{supplementhighlight!43}0.022 & \cellcolor{supplementhighlight!34}1.925 & \cellcolor{supplementhighlight!34}0.045 \\
LingBot-Map & \cellcolor{supplementhighlight!10}0.039 & \cellcolor{supplementhighlight!10}3.800 & \cellcolor{supplementhighlight!7}0.074 & \cellcolor{supplementhighlight!10}0.015 & \cellcolor{supplementhighlight!43}1.353 & \cellcolor{supplementhighlight!7}0.031 & \cellcolor{supplementhighlight!10}0.043 & \cellcolor{supplementhighlight!20}1.054 & \cellcolor{supplementhighlight!14}0.063 & \cellcolor{supplementhighlight!1}0.022 & \cellcolor{supplementhighlight!20}0.433 & \cellcolor{supplementhighlight!4}0.050 & \cellcolor{supplementhighlight!1}0.020 & \cellcolor{supplementhighlight!2}1.130 & \cellcolor{supplementhighlight!2}0.075 & \cellcolor{supplementhighlight!7}0.033 & \cellcolor{supplementhighlight!10}2.993 & \cellcolor{supplementhighlight!7}0.065 \\
VGGT-Omega & \cellcolor{supplementhighlight!80}0.019 & \cellcolor{supplementhighlight!80}1.991 & \cellcolor{supplementhighlight!80}0.042 & \cellcolor{supplementhighlight!43}0.010 & \cellcolor{supplementhighlight!54}1.344 & \cellcolor{supplementhighlight!80}0.017 & \cellcolor{supplementhighlight!80}0.007 & \cellcolor{supplementhighlight!80}0.420 & \cellcolor{supplementhighlight!80}0.012 & \cellcolor{supplementhighlight!54}0.009 & \cellcolor{supplementhighlight!54}0.347 & \cellcolor{supplementhighlight!26}0.043 & \cellcolor{supplementhighlight!80}0.004 & \cellcolor{supplementhighlight!80}0.348 & \cellcolor{supplementhighlight!80}0.021 & \cellcolor{supplementhighlight!34}0.022 & \cellcolor{supplementhighlight!26}1.956 & \cellcolor{supplementhighlight!43}0.045 \\
$\pi^3$ \scriptsize(DINO, our data) & \cellcolor{supplementhighlight!1}0.043 & \cellcolor{supplementhighlight!4}3.819 & \cellcolor{supplementhighlight!1}0.082 & \cellcolor{supplementhighlight!4}0.017 & \cellcolor{supplementhighlight!34}1.406 & \cellcolor{supplementhighlight!4}0.034 & \cellcolor{supplementhighlight!1}0.105 & \cellcolor{supplementhighlight!1}3.935 & \cellcolor{supplementhighlight!1}0.161 & \cellcolor{supplementhighlight!1}0.020 & \cellcolor{supplementhighlight!10}0.511 & \cellcolor{supplementhighlight!14}0.045 & \cellcolor{supplementhighlight!7}0.015 & \cellcolor{supplementhighlight!1}1.187 & \cellcolor{supplementhighlight!10}0.057 & \cellcolor{supplementhighlight!1}0.041 & \cellcolor{supplementhighlight!2}3.469 & \cellcolor{supplementhighlight!2}0.075 \\
\midrule
Any4D & \cellcolor{supplementhighlight!1}0.052 & \cellcolor{supplementhighlight!0}4.987 & \cellcolor{supplementhighlight!1}0.105 & \cellcolor{supplementhighlight!0}0.044 & \cellcolor{supplementhighlight!1}2.713 & \cellcolor{supplementhighlight!0}0.083 & \cellcolor{supplementhighlight!1}0.221 & \cellcolor{supplementhighlight!1}4.705 & \cellcolor{supplementhighlight!1}0.377 & \cellcolor{supplementhighlight!0}0.028 & \cellcolor{supplementhighlight!2}0.717 & \cellcolor{supplementhighlight!2}0.054 & \cellcolor{supplementhighlight!0}0.043 & \cellcolor{supplementhighlight!0}4.021 & \cellcolor{supplementhighlight!0}0.202 & \cellcolor{supplementhighlight!1}0.060 & \cellcolor{supplementhighlight!1}3.849 & \cellcolor{supplementhighlight!1}0.111 \\
Open-D4RT & \cellcolor{supplementhighlight!0}0.060 & \cellcolor{supplementhighlight!1}4.395 & \cellcolor{supplementhighlight!0}0.112 & \cellcolor{supplementhighlight!0}0.027 & \cellcolor{supplementhighlight!0}3.068 & \cellcolor{supplementhighlight!0}0.069 & \cellcolor{supplementhighlight!0}0.449 & \cellcolor{supplementhighlight!0}54.84 & \cellcolor{supplementhighlight!0}0.752 & \cellcolor{supplementhighlight!0}0.032 & \cellcolor{supplementhighlight!1}0.847 & \cellcolor{supplementhighlight!0}0.103 & \cellcolor{supplementhighlight!0}0.028 & \cellcolor{supplementhighlight!0}2.582 & \cellcolor{supplementhighlight!0}0.121 & \cellcolor{supplementhighlight!0}0.062 & \cellcolor{supplementhighlight!0}5.933 & \cellcolor{supplementhighlight!0}0.126 \\
V-DPM & \cellcolor{supplementhighlight!2}0.042 & \cellcolor{supplementhighlight!1}4.114 & \cellcolor{supplementhighlight!2}0.081 & \cellcolor{supplementhighlight!7}0.016 & \cellcolor{supplementhighlight!2}1.709 & \cellcolor{supplementhighlight!10}0.030 & \cellcolor{supplementhighlight!4}0.048 & \cellcolor{supplementhighlight!4}1.965 & \cellcolor{supplementhighlight!4}0.080 & \cellcolor{supplementhighlight!20}0.012 & \cellcolor{supplementhighlight!43}0.395 & \cellcolor{supplementhighlight!7}0.047 & \cellcolor{supplementhighlight!4}0.017 & \cellcolor{supplementhighlight!14}0.910 & \cellcolor{supplementhighlight!7}0.061 & \cellcolor{supplementhighlight!20}0.026 & \cellcolor{supplementhighlight!20}2.097 & \cellcolor{supplementhighlight!20}0.049 \\
Track4World & \cellcolor{supplementhighlight!43}0.030 & \cellcolor{supplementhighlight!34}3.464 & \cellcolor{supplementhighlight!54}0.059 & \cellcolor{supplementhighlight!66}0.008 & \cellcolor{supplementhighlight!7}1.498 & \cellcolor{supplementhighlight!43}0.021 & \cellcolor{supplementhighlight!43}0.013 & \cellcolor{supplementhighlight!34}0.829 & \cellcolor{supplementhighlight!43}0.022 & \cellcolor{supplementhighlight!34}0.010 & \cellcolor{supplementhighlight!4}0.582 & \cellcolor{supplementhighlight!66}0.030 & \cellcolor{supplementhighlight!54}0.008 & \cellcolor{supplementhighlight!7}0.990 & \cellcolor{supplementhighlight!26}0.051 & \cellcolor{supplementhighlight!4}0.034 & \cellcolor{supplementhighlight!4}3.291 & \cellcolor{supplementhighlight!4}0.067 \\
4RC & \cellcolor{supplementhighlight!26}0.032 & \cellcolor{supplementhighlight!54}3.352 & \cellcolor{supplementhighlight!26}0.062 & \cellcolor{supplementhighlight!54}0.008 & \cellcolor{supplementhighlight!20}1.454 & \cellcolor{supplementhighlight!54}0.020 & \cellcolor{supplementhighlight!54}0.013 & \cellcolor{supplementhighlight!54}0.775 & \cellcolor{supplementhighlight!54}0.022 & \cellcolor{supplementhighlight!7}0.015 & \cellcolor{supplementhighlight!0}1.489 & \cellcolor{supplementhighlight!1}0.054 & \cellcolor{supplementhighlight!10}0.013 & \cellcolor{supplementhighlight!4}1.062 & \cellcolor{supplementhighlight!14}0.054 & \cellcolor{supplementhighlight!66}0.021 & \cellcolor{supplementhighlight!43}1.886 & \cellcolor{supplementhighlight!66}0.043 \\
\midrule
\textbf{Ours} \scriptsize(DINO) & \cellcolor{supplementhighlight!7}0.039 & \cellcolor{supplementhighlight!26}3.503 & \cellcolor{supplementhighlight!10}0.073 & \cellcolor{supplementhighlight!2}0.019 & \cellcolor{supplementhighlight!4}1.653 & \cellcolor{supplementhighlight!2}0.044 & \cellcolor{supplementhighlight!20}0.034 & \cellcolor{supplementhighlight!10}1.495 & \cellcolor{supplementhighlight!20}0.055 & \cellcolor{supplementhighlight!14}0.013 & \cellcolor{supplementhighlight!7}0.544 & \cellcolor{supplementhighlight!34}0.038 & \cellcolor{supplementhighlight!14}0.011 & \cellcolor{supplementhighlight!26}0.769 & \cellcolor{supplementhighlight!20}0.052 & \cellcolor{supplementhighlight!14}0.026 & \cellcolor{supplementhighlight!14}2.357 & \cellcolor{supplementhighlight!14}0.054 \\
\textbf{Ours} \scriptsize(VJEPA) & \cellcolor{supplementhighlight!14}0.034 & \cellcolor{supplementhighlight!20}3.571 & \cellcolor{supplementhighlight!20}0.066 & \cellcolor{supplementhighlight!20}0.012 & \cellcolor{supplementhighlight!66}1.333 & \cellcolor{supplementhighlight!20}0.025 & \cellcolor{supplementhighlight!7}0.047 & \cellcolor{supplementhighlight!2}3.190 & \cellcolor{supplementhighlight!7}0.075 & \cellcolor{supplementhighlight!26}0.011 & \cellcolor{supplementhighlight!26}0.417 & \cellcolor{supplementhighlight!54}0.030 & \cellcolor{supplementhighlight!20}0.010 & \cellcolor{supplementhighlight!43}0.500 & \cellcolor{supplementhighlight!34}0.049 & \cellcolor{supplementhighlight!10}0.029 & \cellcolor{supplementhighlight!7}3.177 & \cellcolor{supplementhighlight!10}0.062 \\
\textbf{Ours} \scriptsize(VGGT-$\Omega$) & \cellcolor{supplementhighlight!34}0.030 & \cellcolor{supplementhighlight!66}3.343 & \cellcolor{supplementhighlight!34}0.060 & \cellcolor{supplementhighlight!34}0.010 & \cellcolor{supplementhighlight!26}1.413 & \cellcolor{supplementhighlight!26}0.024 & \cellcolor{supplementhighlight!34}0.014 & \cellcolor{supplementhighlight!26}0.908 & \cellcolor{supplementhighlight!34}0.024 & \cellcolor{supplementhighlight!66}0.008 & \cellcolor{supplementhighlight!66}0.329 & \cellcolor{supplementhighlight!80}0.023 & \cellcolor{supplementhighlight!66}0.004 & \cellcolor{supplementhighlight!66}0.411 & \cellcolor{supplementhighlight!66}0.036 & \cellcolor{supplementhighlight!80}0.020 & \cellcolor{supplementhighlight!54}1.872 & \cellcolor{supplementhighlight!80}0.042 \\
\bottomrule
\end{tabular}
}
\endgroup
\end{table*}

\begin{table*}[!tbp]
\centering
\caption{Per-benchmark cross-view 3D consistency ($\mathrm{AUC}_{5}\uparrow$, $\delta_{5}\uparrow$, $\mathrm{RD}\downarrow$).}
\label{tab:supp_matching}
\begingroup
\fontsize{7pt}{8.4pt}\selectfont
\newcommand{\supplementheaderfont}{\fontsize{6pt}{7.2pt}\selectfont}
\renewcommand{\scriptsize}{\supplementheaderfont}
\definecolor{supplementhighlight}{RGB}{255,215,0}
\setlength{\tabcolsep}{1.5pt}
\setlength{\aboverulesep}{0pt}
\setlength{\belowrulesep}{0pt}
\renewcommand{\arraystretch}{1.15}
\resizebox{\textwidth}{!}{%
\begin{tabular}{@{}lrrr@{\hspace{4pt}}rrr@{\hspace{4pt}}rrr@{\hspace{4pt}}rrr@{\hspace{4pt}}rrr@{\hspace{4pt}}rrr@{}}
\toprule
\multirow{2}{*}{\supplementheaderfont Method} & \multicolumn{3}{c}{\supplementheaderfont 7-Scenes} & \multicolumn{3}{c}{\supplementheaderfont Bonn} & \multicolumn{3}{c}{\supplementheaderfont ETH3D} & \multicolumn{3}{c}{\supplementheaderfont KITTI} & \multicolumn{3}{c}{\supplementheaderfont Sintel} & \multicolumn{3}{c}{\supplementheaderfont ScanNet} \\
\cmidrule(lr){2-4}\cmidrule(lr){5-7}\cmidrule(lr){8-10}\cmidrule(lr){11-13}\cmidrule(lr){14-16}\cmidrule(lr){17-19}
 & {\supplementheaderfont $\mathrm{AUC}_{5}\uparrow$} & {\supplementheaderfont $\delta_{5}\uparrow$} & {\supplementheaderfont RD$\downarrow$} & {\supplementheaderfont $\mathrm{AUC}_{5}\uparrow$} & {\supplementheaderfont $\delta_{5}\uparrow$} & {\supplementheaderfont RD$\downarrow$} & {\supplementheaderfont $\mathrm{AUC}_{5}\uparrow$} & {\supplementheaderfont $\delta_{5}\uparrow$} & {\supplementheaderfont RD$\downarrow$} & {\supplementheaderfont $\mathrm{AUC}_{5}\uparrow$} & {\supplementheaderfont $\delta_{5}\uparrow$} & {\supplementheaderfont RD$\downarrow$} & {\supplementheaderfont $\mathrm{AUC}_{5}\uparrow$} & {\supplementheaderfont $\delta_{5}\uparrow$} & {\supplementheaderfont RD$\downarrow$} & {\supplementheaderfont $\mathrm{AUC}_{5}\uparrow$} & {\supplementheaderfont $\delta_{5}\uparrow$} & {\supplementheaderfont RD$\downarrow$} \\
\midrule
MonST3R & \cellcolor{supplementhighlight!0}23.08 & \cellcolor{supplementhighlight!0}48.92 & \cellcolor{supplementhighlight!0}0.077 & \cellcolor{supplementhighlight!1}57.83 & \cellcolor{supplementhighlight!1}90.17 & \cellcolor{supplementhighlight!1}0.031 & \cellcolor{supplementhighlight!0}16.76 & \cellcolor{supplementhighlight!0}30.98 & \cellcolor{supplementhighlight!0}0.314 & \cellcolor{supplementhighlight!0}21.05 & \cellcolor{supplementhighlight!0}40.35 & \cellcolor{supplementhighlight!0}0.122 & \cellcolor{supplementhighlight!0}34.93 & \cellcolor{supplementhighlight!0}51.98 & \cellcolor{supplementhighlight!0}0.161 & \cellcolor{supplementhighlight!0}48.01 & \cellcolor{supplementhighlight!0}76.45 & \cellcolor{supplementhighlight!0}0.047 \\
MegaSAM & \cellcolor{supplementhighlight!4}52.25 & \cellcolor{supplementhighlight!0}79.18 & \cellcolor{supplementhighlight!0}0.088 & \cellcolor{supplementhighlight!1}60.94 & \cellcolor{supplementhighlight!0}86.35 & \cellcolor{supplementhighlight!0}0.056 & \cellcolor{supplementhighlight!1}32.45 & \cellcolor{supplementhighlight!0}43.67 & \cellcolor{supplementhighlight!0}0.591 & \cellcolor{supplementhighlight!7}49.26 & \cellcolor{supplementhighlight!1}71.54 & \cellcolor{supplementhighlight!0}0.076 & \cellcolor{supplementhighlight!26}68.26 & \cellcolor{supplementhighlight!7}85.59 & \cellcolor{supplementhighlight!4}0.047 & \cellcolor{supplementhighlight!2}56.67 & \cellcolor{supplementhighlight!1}77.34 & \cellcolor{supplementhighlight!0}0.099 \\
VGGT & \cellcolor{supplementhighlight!10}60.72 & \cellcolor{supplementhighlight!10}89.33 & \cellcolor{supplementhighlight!14}0.030 & \cellcolor{supplementhighlight!14}68.47 & \cellcolor{supplementhighlight!4}94.15 & \cellcolor{supplementhighlight!7}0.025 & \cellcolor{supplementhighlight!7}51.57 & \cellcolor{supplementhighlight!4}79.27 & \cellcolor{supplementhighlight!7}0.036 & \cellcolor{supplementhighlight!34}58.19 & \cellcolor{supplementhighlight!34}85.13 & \cellcolor{supplementhighlight!34}0.033 & \cellcolor{supplementhighlight!14}65.74 & \cellcolor{supplementhighlight!10}85.98 & \cellcolor{supplementhighlight!26}0.035 & \cellcolor{supplementhighlight!43}74.21 & \cellcolor{supplementhighlight!34}96.71 & \cellcolor{supplementhighlight!54}0.016 \\
MapAnything & \cellcolor{supplementhighlight!7}56.46 & \cellcolor{supplementhighlight!14}90.13 & \cellcolor{supplementhighlight!10}0.031 & \cellcolor{supplementhighlight!2}62.28 & \cellcolor{supplementhighlight!7}94.28 & \cellcolor{supplementhighlight!4}0.025 & \cellcolor{supplementhighlight!14}54.27 & \cellcolor{supplementhighlight!14}86.43 & \cellcolor{supplementhighlight!10}0.033 & \cellcolor{supplementhighlight!1}46.22 & \cellcolor{supplementhighlight!2}76.42 & \cellcolor{supplementhighlight!2}0.055 & \cellcolor{supplementhighlight!1}52.15 & \cellcolor{supplementhighlight!1}76.42 & \cellcolor{supplementhighlight!0}0.087 & \cellcolor{supplementhighlight!7}58.38 & \cellcolor{supplementhighlight!7}90.21 & \cellcolor{supplementhighlight!2}0.030 \\
$\pi^3$ & \cellcolor{supplementhighlight!43}65.29 & \cellcolor{supplementhighlight!43}93.61 & \cellcolor{supplementhighlight!54}0.024 & \cellcolor{supplementhighlight!43}71.15 & \cellcolor{supplementhighlight!66}96.16 & \cellcolor{supplementhighlight!20}0.022 & \cellcolor{supplementhighlight!34}68.88 & \cellcolor{supplementhighlight!43}95.50 & \cellcolor{supplementhighlight!54}0.019 & \cellcolor{supplementhighlight!80}63.68 & \cellcolor{supplementhighlight!66}91.76 & \cellcolor{supplementhighlight!66}0.025 & \cellcolor{supplementhighlight!34}72.45 & \cellcolor{supplementhighlight!34}92.33 & \cellcolor{supplementhighlight!34}0.029 & \cellcolor{supplementhighlight!66}75.99 & \cellcolor{supplementhighlight!66}97.31 & \cellcolor{supplementhighlight!66}0.015 \\
Depth Anything 3 & \cellcolor{supplementhighlight!66}65.54 & \cellcolor{supplementhighlight!26}93.14 & \cellcolor{supplementhighlight!34}0.025 & \cellcolor{supplementhighlight!54}71.30 & \cellcolor{supplementhighlight!54}96.06 & \cellcolor{supplementhighlight!80}0.020 & \cellcolor{supplementhighlight!54}72.12 & \cellcolor{supplementhighlight!54}96.25 & \cellcolor{supplementhighlight!26}0.028 & \cellcolor{supplementhighlight!4}47.06 & \cellcolor{supplementhighlight!4}77.15 & \cellcolor{supplementhighlight!4}0.044 & \cellcolor{supplementhighlight!7}60.97 & \cellcolor{supplementhighlight!14}86.13 & \cellcolor{supplementhighlight!7}0.037 & \cellcolor{supplementhighlight!10}67.22 & \cellcolor{supplementhighlight!10}93.49 & \cellcolor{supplementhighlight!10}0.022 \\
LingBot-Map & \cellcolor{supplementhighlight!1}47.22 & \cellcolor{supplementhighlight!1}85.33 & \cellcolor{supplementhighlight!1}0.036 & \cellcolor{supplementhighlight!0}56.48 & \cellcolor{supplementhighlight!1}92.07 & \cellcolor{supplementhighlight!1}0.029 & \cellcolor{supplementhighlight!2}40.65 & \cellcolor{supplementhighlight!2}75.08 & \cellcolor{supplementhighlight!2}0.046 & \cellcolor{supplementhighlight!0}39.25 & \cellcolor{supplementhighlight!0}68.27 & \cellcolor{supplementhighlight!1}0.060 & \cellcolor{supplementhighlight!0}44.24 & \cellcolor{supplementhighlight!0}71.39 & \cellcolor{supplementhighlight!1}0.058 & \cellcolor{supplementhighlight!1}53.43 & \cellcolor{supplementhighlight!1}87.67 & \cellcolor{supplementhighlight!1}0.031 \\
VGGT-Omega & \cellcolor{supplementhighlight!2}49.72 & \cellcolor{supplementhighlight!4}87.97 & \cellcolor{supplementhighlight!4}0.033 & \cellcolor{supplementhighlight!10}67.57 & \cellcolor{supplementhighlight!14}95.05 & \cellcolor{supplementhighlight!14}0.023 & \cellcolor{supplementhighlight!80}81.83 & \cellcolor{supplementhighlight!80}98.68 & \cellcolor{supplementhighlight!80}0.011 & \cellcolor{supplementhighlight!66}63.23 & \cellcolor{supplementhighlight!80}92.17 & \cellcolor{supplementhighlight!80}0.024 & \cellcolor{supplementhighlight!80}78.23 & \cellcolor{supplementhighlight!80}93.73 & \cellcolor{supplementhighlight!43}0.025 & \cellcolor{supplementhighlight!34}73.65 & \cellcolor{supplementhighlight!14}95.95 & \cellcolor{supplementhighlight!14}0.018 \\
$\pi^3$ \scriptsize(DINO, our data) & \cellcolor{supplementhighlight!1}47.89 & \cellcolor{supplementhighlight!2}85.64 & \cellcolor{supplementhighlight!2}0.035 & \cellcolor{supplementhighlight!4}63.35 & \cellcolor{supplementhighlight!10}94.43 & \cellcolor{supplementhighlight!10}0.024 & \cellcolor{supplementhighlight!0}21.36 & \cellcolor{supplementhighlight!1}47.15 & \cellcolor{supplementhighlight!1}0.085 & \cellcolor{supplementhighlight!10}50.62 & \cellcolor{supplementhighlight!20}83.42 & \cellcolor{supplementhighlight!14}0.035 & \cellcolor{supplementhighlight!4}60.82 & \cellcolor{supplementhighlight!4}84.90 & \cellcolor{supplementhighlight!10}0.037 & \cellcolor{supplementhighlight!1}53.25 & \cellcolor{supplementhighlight!2}88.50 & \cellcolor{supplementhighlight!4}0.029 \\
\midrule
Any4D & \cellcolor{supplementhighlight!0}40.94 & \cellcolor{supplementhighlight!0}77.20 & \cellcolor{supplementhighlight!0}0.047 & \cellcolor{supplementhighlight!0}54.21 & \cellcolor{supplementhighlight!0}86.49 & \cellcolor{supplementhighlight!0}0.040 & \cellcolor{supplementhighlight!1}21.51 & \cellcolor{supplementhighlight!1}50.47 & \cellcolor{supplementhighlight!1}0.083 & \cellcolor{supplementhighlight!1}40.64 & \cellcolor{supplementhighlight!1}71.69 & \cellcolor{supplementhighlight!1}0.056 & \cellcolor{supplementhighlight!0}45.35 & \cellcolor{supplementhighlight!0}70.49 & \cellcolor{supplementhighlight!0}0.067 & \cellcolor{supplementhighlight!0}43.16 & \cellcolor{supplementhighlight!0}77.09 & \cellcolor{supplementhighlight!1}0.043 \\
Open-D4RT & \cellcolor{supplementhighlight!0}41.73 & \cellcolor{supplementhighlight!1}80.14 & \cellcolor{supplementhighlight!1}0.040 & \cellcolor{supplementhighlight!0}46.15 & \cellcolor{supplementhighlight!0}82.32 & \cellcolor{supplementhighlight!0}0.040 & \cellcolor{supplementhighlight!0}7.958 & \cellcolor{supplementhighlight!0}19.90 & \cellcolor{supplementhighlight!0}0.499 & \cellcolor{supplementhighlight!0}35.24 & \cellcolor{supplementhighlight!0}62.48 & \cellcolor{supplementhighlight!0}0.070 & \cellcolor{supplementhighlight!1}57.67 & \cellcolor{supplementhighlight!1}76.41 & \cellcolor{supplementhighlight!1}0.064 & \cellcolor{supplementhighlight!0}35.41 & \cellcolor{supplementhighlight!0}71.65 & \cellcolor{supplementhighlight!0}0.047 \\
V-DPM & \cellcolor{supplementhighlight!14}61.37 & \cellcolor{supplementhighlight!7}88.58 & \cellcolor{supplementhighlight!7}0.031 & \cellcolor{supplementhighlight!7}65.67 & \cellcolor{supplementhighlight!2}92.56 & \cellcolor{supplementhighlight!2}0.027 & \cellcolor{supplementhighlight!10}53.67 & \cellcolor{supplementhighlight!10}84.10 & \cellcolor{supplementhighlight!14}0.033 & \cellcolor{supplementhighlight!26}57.00 & \cellcolor{supplementhighlight!26}84.82 & \cellcolor{supplementhighlight!20}0.034 & \cellcolor{supplementhighlight!20}66.45 & \cellcolor{supplementhighlight!26}86.85 & \cellcolor{supplementhighlight!20}0.035 & \cellcolor{supplementhighlight!26}73.58 & \cellcolor{supplementhighlight!26}96.63 & \cellcolor{supplementhighlight!43}0.017 \\
Track4World & \cellcolor{supplementhighlight!20}62.77 & \cellcolor{supplementhighlight!20}92.29 & \cellcolor{supplementhighlight!20}0.026 & \cellcolor{supplementhighlight!34}69.94 & \cellcolor{supplementhighlight!34}95.65 & \cellcolor{supplementhighlight!43}0.021 & \cellcolor{supplementhighlight!26}65.87 & \cellcolor{supplementhighlight!26}92.98 & \cellcolor{supplementhighlight!20}0.029 & \cellcolor{supplementhighlight!2}46.49 & \cellcolor{supplementhighlight!7}77.64 & \cellcolor{supplementhighlight!7}0.043 & \cellcolor{supplementhighlight!10}65.49 & \cellcolor{supplementhighlight!20}86.81 & \cellcolor{supplementhighlight!14}0.036 & \cellcolor{supplementhighlight!4}58.12 & \cellcolor{supplementhighlight!4}89.51 & \cellcolor{supplementhighlight!7}0.028 \\
4RC & \cellcolor{supplementhighlight!54}65.54 & \cellcolor{supplementhighlight!34}93.44 & \cellcolor{supplementhighlight!26}0.026 & \cellcolor{supplementhighlight!80}71.49 & \cellcolor{supplementhighlight!80}96.27 & \cellcolor{supplementhighlight!54}0.020 & \cellcolor{supplementhighlight!43}70.46 & \cellcolor{supplementhighlight!34}94.24 & \cellcolor{supplementhighlight!43}0.021 & \cellcolor{supplementhighlight!14}51.13 & \cellcolor{supplementhighlight!10}80.13 & \cellcolor{supplementhighlight!10}0.042 & \cellcolor{supplementhighlight!2}59.14 & \cellcolor{supplementhighlight!2}80.04 & \cellcolor{supplementhighlight!2}0.048 & \cellcolor{supplementhighlight!54}74.82 & \cellcolor{supplementhighlight!54}97.15 & \cellcolor{supplementhighlight!34}0.017 \\
\midrule
\textbf{Ours} \scriptsize(DINO) & \cellcolor{supplementhighlight!34}64.99 & \cellcolor{supplementhighlight!66}93.98 & \cellcolor{supplementhighlight!66}0.023 & \cellcolor{supplementhighlight!20}68.78 & \cellcolor{supplementhighlight!20}95.44 & \cellcolor{supplementhighlight!26}0.022 & \cellcolor{supplementhighlight!20}61.36 & \cellcolor{supplementhighlight!20}92.15 & \cellcolor{supplementhighlight!34}0.022 & \cellcolor{supplementhighlight!43}60.09 & \cellcolor{supplementhighlight!54}89.06 & \cellcolor{supplementhighlight!54}0.028 & \cellcolor{supplementhighlight!43}75.12 & \cellcolor{supplementhighlight!54}92.89 & \cellcolor{supplementhighlight!80}0.023 & \cellcolor{supplementhighlight!14}70.03 & \cellcolor{supplementhighlight!20}96.40 & \cellcolor{supplementhighlight!20}0.018 \\
\textbf{Ours} \scriptsize(VJEPA) & \cellcolor{supplementhighlight!26}64.18 & \cellcolor{supplementhighlight!54}93.82 & \cellcolor{supplementhighlight!43}0.024 & \cellcolor{supplementhighlight!26}69.29 & \cellcolor{supplementhighlight!26}95.57 & \cellcolor{supplementhighlight!34}0.021 & \cellcolor{supplementhighlight!4}49.93 & \cellcolor{supplementhighlight!7}80.69 & \cellcolor{supplementhighlight!4}0.037 & \cellcolor{supplementhighlight!54}60.57 & \cellcolor{supplementhighlight!43}88.63 & \cellcolor{supplementhighlight!43}0.028 & \cellcolor{supplementhighlight!54}75.37 & \cellcolor{supplementhighlight!43}92.82 & \cellcolor{supplementhighlight!66}0.024 & \cellcolor{supplementhighlight!20}71.03 & \cellcolor{supplementhighlight!43}96.73 & \cellcolor{supplementhighlight!26}0.018 \\
\textbf{Ours} \scriptsize(VGGT-$\Omega$) & \cellcolor{supplementhighlight!80}67.56 & \cellcolor{supplementhighlight!80}94.47 & \cellcolor{supplementhighlight!80}0.022 & \cellcolor{supplementhighlight!66}71.38 & \cellcolor{supplementhighlight!43}95.70 & \cellcolor{supplementhighlight!66}0.020 & \cellcolor{supplementhighlight!66}76.67 & \cellcolor{supplementhighlight!66}97.76 & \cellcolor{supplementhighlight!66}0.013 & \cellcolor{supplementhighlight!20}53.91 & \cellcolor{supplementhighlight!14}82.73 & \cellcolor{supplementhighlight!26}0.033 & \cellcolor{supplementhighlight!66}78.19 & \cellcolor{supplementhighlight!66}93.16 & \cellcolor{supplementhighlight!54}0.024 & \cellcolor{supplementhighlight!80}77.87 & \cellcolor{supplementhighlight!80}97.65 & \cellcolor{supplementhighlight!80}0.014 \\
\bottomrule
\end{tabular}
}
\endgroup
\end{table*}

\begin{table*}[!tbp]
\centering
\caption{Per-benchmark global-point accuracy ($\mathrm{AUC}_{25}\uparrow$, $\delta_{25}\uparrow$, $\mathrm{RD}\downarrow$).}
\label{tab:supp_global_points}
\begingroup
\fontsize{7pt}{8.4pt}\selectfont
\newcommand{\supplementheaderfont}{\fontsize{6pt}{7.2pt}\selectfont}
\renewcommand{\scriptsize}{\supplementheaderfont}
\definecolor{supplementhighlight}{RGB}{255,215,0}
\setlength{\tabcolsep}{1.5pt}
\setlength{\aboverulesep}{0pt}
\setlength{\belowrulesep}{0pt}
\renewcommand{\arraystretch}{1.15}
\resizebox{\textwidth}{!}{%
\begin{tabular}{@{}lrrr@{\hspace{4pt}}rrr@{\hspace{4pt}}rrr@{\hspace{4pt}}rrr@{\hspace{4pt}}rrr@{\hspace{4pt}}rrr@{}}
\toprule
\multirow{2}{*}{\supplementheaderfont Method} & \multicolumn{3}{c}{\supplementheaderfont 7-Scenes} & \multicolumn{3}{c}{\supplementheaderfont Bonn} & \multicolumn{3}{c}{\supplementheaderfont ETH3D} & \multicolumn{3}{c}{\supplementheaderfont KITTI} & \multicolumn{3}{c}{\supplementheaderfont Sintel} & \multicolumn{3}{c}{\supplementheaderfont ScanNet} \\
\cmidrule(lr){2-4}\cmidrule(lr){5-7}\cmidrule(lr){8-10}\cmidrule(lr){11-13}\cmidrule(lr){14-16}\cmidrule(lr){17-19}
 & {\supplementheaderfont $\mathrm{AUC}_{25}\uparrow$} & {\supplementheaderfont $\delta_{25}\uparrow$} & {\supplementheaderfont RD$\downarrow$} & {\supplementheaderfont $\mathrm{AUC}_{25}\uparrow$} & {\supplementheaderfont $\delta_{25}\uparrow$} & {\supplementheaderfont RD$\downarrow$} & {\supplementheaderfont $\mathrm{AUC}_{25}\uparrow$} & {\supplementheaderfont $\delta_{25}\uparrow$} & {\supplementheaderfont RD$\downarrow$} & {\supplementheaderfont $\mathrm{AUC}_{25}\uparrow$} & {\supplementheaderfont $\delta_{25}\uparrow$} & {\supplementheaderfont RD$\downarrow$} & {\supplementheaderfont $\mathrm{AUC}_{25}\uparrow$} & {\supplementheaderfont $\delta_{25}\uparrow$} & {\supplementheaderfont RD$\downarrow$} & {\supplementheaderfont $\mathrm{AUC}_{25}\uparrow$} & {\supplementheaderfont $\delta_{25}\uparrow$} & {\supplementheaderfont RD$\downarrow$} \\
\midrule
MonST3R & \cellcolor{supplementhighlight!0}63.64 & \cellcolor{supplementhighlight!0}92.65 & \cellcolor{supplementhighlight!0}0.101 & \cellcolor{supplementhighlight!0}74.54 & \cellcolor{supplementhighlight!0}93.97 & \cellcolor{supplementhighlight!0}0.080 & \cellcolor{supplementhighlight!0}34.52 & \cellcolor{supplementhighlight!0}58.89 & \cellcolor{supplementhighlight!0}0.367 & \cellcolor{supplementhighlight!0}28.37 & \cellcolor{supplementhighlight!0}59.42 & \cellcolor{supplementhighlight!0}0.303 & \cellcolor{supplementhighlight!0}33.39 & \cellcolor{supplementhighlight!0}61.52 & \cellcolor{supplementhighlight!0}0.357 & \cellcolor{supplementhighlight!0}71.06 & \cellcolor{supplementhighlight!0}92.88 & \cellcolor{supplementhighlight!0}0.095 \\
MegaSAM & \cellcolor{supplementhighlight!0}69.01 & \cellcolor{supplementhighlight!0}87.27 & \cellcolor{supplementhighlight!0}0.115 & \cellcolor{supplementhighlight!0}73.72 & \cellcolor{supplementhighlight!0}91.57 & \cellcolor{supplementhighlight!0}0.089 & \cellcolor{supplementhighlight!0}18.00 & \cellcolor{supplementhighlight!0}28.82 & \cellcolor{supplementhighlight!0}0.787 & \cellcolor{supplementhighlight!1}55.87 & \cellcolor{supplementhighlight!0}78.64 & \cellcolor{supplementhighlight!0}0.193 & \cellcolor{supplementhighlight!1}44.59 & \cellcolor{supplementhighlight!1}71.85 & \cellcolor{supplementhighlight!1}0.254 & \cellcolor{supplementhighlight!0}49.85 & \cellcolor{supplementhighlight!0}74.22 & \cellcolor{supplementhighlight!0}0.214 \\
VGGT & \cellcolor{supplementhighlight!7}82.79 & \cellcolor{supplementhighlight!1}96.56 & \cellcolor{supplementhighlight!10}0.049 & \cellcolor{supplementhighlight!7}83.48 & \cellcolor{supplementhighlight!7}96.72 & \cellcolor{supplementhighlight!14}0.049 & \cellcolor{supplementhighlight!14}82.99 & \cellcolor{supplementhighlight!10}98.43 & \cellcolor{supplementhighlight!14}0.044 & \cellcolor{supplementhighlight!20}75.69 & \cellcolor{supplementhighlight!14}97.70 & \cellcolor{supplementhighlight!20}0.065 & \cellcolor{supplementhighlight!14}50.95 & \cellcolor{supplementhighlight!4}76.49 & \cellcolor{supplementhighlight!14}0.201 & \cellcolor{supplementhighlight!54}88.77 & \cellcolor{supplementhighlight!43}98.94 & \cellcolor{supplementhighlight!54}0.030 \\
MapAnything & \cellcolor{supplementhighlight!4}81.74 & \cellcolor{supplementhighlight!7}96.96 & \cellcolor{supplementhighlight!4}0.052 & \cellcolor{supplementhighlight!1}80.96 & \cellcolor{supplementhighlight!2}96.71 & \cellcolor{supplementhighlight!2}0.056 & \cellcolor{supplementhighlight!10}82.50 & \cellcolor{supplementhighlight!14}98.57 & \cellcolor{supplementhighlight!10}0.047 & \cellcolor{supplementhighlight!4}69.89 & \cellcolor{supplementhighlight!2}94.95 & \cellcolor{supplementhighlight!2}0.100 & \cellcolor{supplementhighlight!0}39.01 & \cellcolor{supplementhighlight!0}66.21 & \cellcolor{supplementhighlight!0}0.323 & \cellcolor{supplementhighlight!2}81.29 & \cellcolor{supplementhighlight!1}97.13 & \cellcolor{supplementhighlight!2}0.060 \\
$\pi^3$ & \cellcolor{supplementhighlight!66}85.46 & \cellcolor{supplementhighlight!80}98.40 & \cellcolor{supplementhighlight!66}0.039 & \cellcolor{supplementhighlight!43}86.02 & \cellcolor{supplementhighlight!26}97.59 & \cellcolor{supplementhighlight!54}0.039 & \cellcolor{supplementhighlight!34}91.15 & \cellcolor{supplementhighlight!43}99.64 & \cellcolor{supplementhighlight!43}0.023 & \cellcolor{supplementhighlight!34}76.73 & \cellcolor{supplementhighlight!43}98.75 & \cellcolor{supplementhighlight!34}0.061 & \cellcolor{supplementhighlight!43}57.17 & \cellcolor{supplementhighlight!34}80.61 & \cellcolor{supplementhighlight!43}0.159 & \cellcolor{supplementhighlight!80}89.99 & \cellcolor{supplementhighlight!80}99.18 & \cellcolor{supplementhighlight!80}0.027 \\
Depth Anything 3 & \cellcolor{supplementhighlight!34}84.46 & \cellcolor{supplementhighlight!26}97.59 & \cellcolor{supplementhighlight!34}0.044 & \cellcolor{supplementhighlight!20}84.79 & \cellcolor{supplementhighlight!4}96.71 & \cellcolor{supplementhighlight!10}0.049 & \cellcolor{supplementhighlight!66}93.57 & \cellcolor{supplementhighlight!34}99.60 & \cellcolor{supplementhighlight!34}0.024 & \cellcolor{supplementhighlight!0}45.35 & \cellcolor{supplementhighlight!1}88.57 & \cellcolor{supplementhighlight!0}0.147 & \cellcolor{supplementhighlight!4}49.54 & \cellcolor{supplementhighlight!20}79.98 & \cellcolor{supplementhighlight!20}0.197 & \cellcolor{supplementhighlight!14}86.28 & \cellcolor{supplementhighlight!10}98.07 & \cellcolor{supplementhighlight!14}0.038 \\
LingBot-Map & \cellcolor{supplementhighlight!1}79.77 & \cellcolor{supplementhighlight!10}97.05 & \cellcolor{supplementhighlight!1}0.055 & \cellcolor{supplementhighlight!2}81.69 & \cellcolor{supplementhighlight!1}96.36 & \cellcolor{supplementhighlight!1}0.059 & \cellcolor{supplementhighlight!4}81.36 & \cellcolor{supplementhighlight!2}97.82 & \cellcolor{supplementhighlight!4}0.051 & \cellcolor{supplementhighlight!2}69.67 & \cellcolor{supplementhighlight!4}95.20 & \cellcolor{supplementhighlight!4}0.087 & \cellcolor{supplementhighlight!2}47.77 & \cellcolor{supplementhighlight!2}75.41 & \cellcolor{supplementhighlight!4}0.218 & \cellcolor{supplementhighlight!7}82.66 & \cellcolor{supplementhighlight!2}97.46 & \cellcolor{supplementhighlight!4}0.052 \\
VGGT-Omega & \cellcolor{supplementhighlight!14}83.51 & \cellcolor{supplementhighlight!34}97.75 & \cellcolor{supplementhighlight!26}0.045 & \cellcolor{supplementhighlight!10}83.69 & \cellcolor{supplementhighlight!20}97.03 & \cellcolor{supplementhighlight!20}0.048 & \cellcolor{supplementhighlight!80}95.65 & \cellcolor{supplementhighlight!80}99.88 & \cellcolor{supplementhighlight!80}0.011 & \cellcolor{supplementhighlight!80}84.27 & \cellcolor{supplementhighlight!80}99.17 & \cellcolor{supplementhighlight!66}0.042 & \cellcolor{supplementhighlight!80}79.81 & \cellcolor{supplementhighlight!80}96.07 & \cellcolor{supplementhighlight!66}0.107 & \cellcolor{supplementhighlight!26}87.91 & \cellcolor{supplementhighlight!20}98.48 & \cellcolor{supplementhighlight!26}0.035 \\
$\pi^3$ \scriptsize(DINO, our data) & \cellcolor{supplementhighlight!2}80.88 & \cellcolor{supplementhighlight!20}97.58 & \cellcolor{supplementhighlight!2}0.052 & \cellcolor{supplementhighlight!26}85.12 & \cellcolor{supplementhighlight!66}98.43 & \cellcolor{supplementhighlight!26}0.042 & \cellcolor{supplementhighlight!1}65.49 & \cellcolor{supplementhighlight!1}92.83 & \cellcolor{supplementhighlight!1}0.096 & \cellcolor{supplementhighlight!10}73.77 & \cellcolor{supplementhighlight!26}98.03 & \cellcolor{supplementhighlight!14}0.069 & \cellcolor{supplementhighlight!26}53.25 & \cellcolor{supplementhighlight!43}80.76 & \cellcolor{supplementhighlight!26}0.166 & \cellcolor{supplementhighlight!4}81.96 & \cellcolor{supplementhighlight!7}97.99 & \cellcolor{supplementhighlight!7}0.051 \\
\midrule
Any4D & \cellcolor{supplementhighlight!1}76.99 & \cellcolor{supplementhighlight!0}95.96 & \cellcolor{supplementhighlight!0}0.068 & \cellcolor{supplementhighlight!1}76.89 & \cellcolor{supplementhighlight!1}96.35 & \cellcolor{supplementhighlight!1}0.066 & \cellcolor{supplementhighlight!1}65.72 & \cellcolor{supplementhighlight!1}94.60 & \cellcolor{supplementhighlight!1}0.122 & \cellcolor{supplementhighlight!7}72.54 & \cellcolor{supplementhighlight!7}96.52 & \cellcolor{supplementhighlight!7}0.078 & \cellcolor{supplementhighlight!1}42.62 & \cellcolor{supplementhighlight!1}69.50 & \cellcolor{supplementhighlight!1}0.283 & \cellcolor{supplementhighlight!1}76.05 & \cellcolor{supplementhighlight!1}96.87 & \cellcolor{supplementhighlight!1}0.067 \\
Open-D4RT & \cellcolor{supplementhighlight!0}76.24 & \cellcolor{supplementhighlight!2}96.84 & \cellcolor{supplementhighlight!1}0.065 & \cellcolor{supplementhighlight!0}75.71 & \cellcolor{supplementhighlight!0}95.45 & \cellcolor{supplementhighlight!0}0.071 & \cellcolor{supplementhighlight!0}20.05 & \cellcolor{supplementhighlight!0}42.50 & \cellcolor{supplementhighlight!0}0.603 & \cellcolor{supplementhighlight!1}52.93 & \cellcolor{supplementhighlight!0}86.87 & \cellcolor{supplementhighlight!1}0.146 & \cellcolor{supplementhighlight!0}40.80 & \cellcolor{supplementhighlight!0}66.97 & \cellcolor{supplementhighlight!0}0.284 & \cellcolor{supplementhighlight!0}71.92 & \cellcolor{supplementhighlight!0}95.46 & \cellcolor{supplementhighlight!0}0.085 \\
V-DPM & \cellcolor{supplementhighlight!10}82.84 & \cellcolor{supplementhighlight!1}96.35 & \cellcolor{supplementhighlight!7}0.051 & \cellcolor{supplementhighlight!4}82.73 & \cellcolor{supplementhighlight!14}96.80 & \cellcolor{supplementhighlight!4}0.051 & \cellcolor{supplementhighlight!7}81.78 & \cellcolor{supplementhighlight!7}98.25 & \cellcolor{supplementhighlight!7}0.048 & \cellcolor{supplementhighlight!26}76.19 & \cellcolor{supplementhighlight!10}97.53 & \cellcolor{supplementhighlight!26}0.064 & \cellcolor{supplementhighlight!20}51.26 & \cellcolor{supplementhighlight!7}76.74 & \cellcolor{supplementhighlight!10}0.206 & \cellcolor{supplementhighlight!34}88.64 & \cellcolor{supplementhighlight!54}99.00 & \cellcolor{supplementhighlight!43}0.031 \\
Track4World & \cellcolor{supplementhighlight!20}83.58 & \cellcolor{supplementhighlight!14}97.57 & \cellcolor{supplementhighlight!20}0.046 & \cellcolor{supplementhighlight!14}84.48 & \cellcolor{supplementhighlight!10}96.78 & \cellcolor{supplementhighlight!7}0.049 & \cellcolor{supplementhighlight!26}88.01 & \cellcolor{supplementhighlight!20}99.39 & \cellcolor{supplementhighlight!20}0.036 & \cellcolor{supplementhighlight!14}74.28 & \cellcolor{supplementhighlight!20}97.71 & \cellcolor{supplementhighlight!10}0.070 & \cellcolor{supplementhighlight!7}50.03 & \cellcolor{supplementhighlight!14}77.53 & \cellcolor{supplementhighlight!7}0.216 & \cellcolor{supplementhighlight!1}76.73 & \cellcolor{supplementhighlight!4}97.62 & \cellcolor{supplementhighlight!1}0.062 \\
4RC & \cellcolor{supplementhighlight!26}84.27 & \cellcolor{supplementhighlight!4}96.95 & \cellcolor{supplementhighlight!14}0.046 & \cellcolor{supplementhighlight!66}86.17 & \cellcolor{supplementhighlight!34}97.97 & \cellcolor{supplementhighlight!34}0.042 & \cellcolor{supplementhighlight!43}92.12 & \cellcolor{supplementhighlight!26}99.47 & \cellcolor{supplementhighlight!54}0.022 & \cellcolor{supplementhighlight!0}49.72 & \cellcolor{supplementhighlight!1}89.31 & \cellcolor{supplementhighlight!1}0.136 & \cellcolor{supplementhighlight!10}50.88 & \cellcolor{supplementhighlight!10}77.06 & \cellcolor{supplementhighlight!2}0.220 & \cellcolor{supplementhighlight!43}88.72 & \cellcolor{supplementhighlight!26}98.68 & \cellcolor{supplementhighlight!34}0.031 \\
\midrule
\textbf{Ours} \scriptsize(DINO) & \cellcolor{supplementhighlight!43}84.86 & \cellcolor{supplementhighlight!43}98.04 & \cellcolor{supplementhighlight!43}0.042 & \cellcolor{supplementhighlight!34}85.75 & \cellcolor{supplementhighlight!43}98.01 & \cellcolor{supplementhighlight!43}0.040 & \cellcolor{supplementhighlight!20}85.87 & \cellcolor{supplementhighlight!54}99.67 & \cellcolor{supplementhighlight!26}0.036 & \cellcolor{supplementhighlight!43}77.79 & \cellcolor{supplementhighlight!34}98.66 & \cellcolor{supplementhighlight!43}0.058 & \cellcolor{supplementhighlight!34}55.09 & \cellcolor{supplementhighlight!26}80.43 & \cellcolor{supplementhighlight!34}0.165 & \cellcolor{supplementhighlight!10}86.21 & \cellcolor{supplementhighlight!34}98.72 & \cellcolor{supplementhighlight!20}0.037 \\
\textbf{Ours} \scriptsize(VJEPA) & \cellcolor{supplementhighlight!54}85.09 & \cellcolor{supplementhighlight!54}98.08 & \cellcolor{supplementhighlight!54}0.041 & \cellcolor{supplementhighlight!80}87.24 & \cellcolor{supplementhighlight!80}98.79 & \cellcolor{supplementhighlight!80}0.036 & \cellcolor{supplementhighlight!2}79.94 & \cellcolor{supplementhighlight!4}98.22 & \cellcolor{supplementhighlight!2}0.051 & \cellcolor{supplementhighlight!66}84.20 & \cellcolor{supplementhighlight!66}99.01 & \cellcolor{supplementhighlight!80}0.042 & \cellcolor{supplementhighlight!54}62.03 & \cellcolor{supplementhighlight!54}85.72 & \cellcolor{supplementhighlight!54}0.130 & \cellcolor{supplementhighlight!20}86.69 & \cellcolor{supplementhighlight!14}98.18 & \cellcolor{supplementhighlight!10}0.039 \\
\textbf{Ours} \scriptsize(VGGT-$\Omega$) & \cellcolor{supplementhighlight!80}86.25 & \cellcolor{supplementhighlight!66}98.09 & \cellcolor{supplementhighlight!80}0.038 & \cellcolor{supplementhighlight!54}86.17 & \cellcolor{supplementhighlight!54}98.26 & \cellcolor{supplementhighlight!66}0.039 & \cellcolor{supplementhighlight!54}93.01 & \cellcolor{supplementhighlight!66}99.87 & \cellcolor{supplementhighlight!66}0.018 & \cellcolor{supplementhighlight!54}82.87 & \cellcolor{supplementhighlight!54}98.80 & \cellcolor{supplementhighlight!54}0.045 & \cellcolor{supplementhighlight!66}66.95 & \cellcolor{supplementhighlight!66}89.99 & \cellcolor{supplementhighlight!80}0.105 & \cellcolor{supplementhighlight!66}89.53 & \cellcolor{supplementhighlight!66}99.04 & \cellcolor{supplementhighlight!66}0.028 \\
\bottomrule
\end{tabular}
}
\endgroup
\end{table*}

\begin{table*}[!tbp]
\centering
\caption{Per-benchmark 3D tracking accuracy in camera coordinates ($\mathrm{AUC}_{25}\uparrow$, $\mathrm{RelEPE}\downarrow$).}
\label{tab:supp_tracking_camera}
\begingroup
\fontsize{7pt}{8.4pt}\selectfont
\newcommand{\supplementheaderfont}{\fontsize{5pt}{6pt}\selectfont}
\renewcommand{\scriptsize}{\supplementheaderfont}
\definecolor{supplementhighlight}{RGB}{255,215,0}
\setlength{\tabcolsep}{0.75pt}
\setlength{\aboverulesep}{0pt}
\setlength{\belowrulesep}{0pt}
\renewcommand{\arraystretch}{1.15}
\resizebox{\textwidth}{!}{%
\begin{tabular}{@{}lrrrr@{\hspace{4pt}}rrrr@{\hspace{4pt}}rrrr@{\hspace{4pt}}rrrr@{\hspace{4pt}}rrrr@{\hspace{4pt}}rrrr@{}}
\toprule
\multirow{3}{*}{\supplementheaderfont Method} & \multicolumn{4}{c}{\supplementheaderfont DynamicReplica} & \multicolumn{4}{c}{\supplementheaderfont PointOdyssey} & \multicolumn{4}{c}{\supplementheaderfont Syn4D} & \multicolumn{4}{c}{\supplementheaderfont ADT} & \multicolumn{4}{c}{\supplementheaderfont PStudio} & \multicolumn{4}{c}{\supplementheaderfont RigIDTrack} \\
\cmidrule(lr){2-5}\cmidrule(lr){6-9}\cmidrule(lr){10-13}\cmidrule(lr){14-17}\cmidrule(lr){18-21}\cmidrule(lr){22-25}
 & \multicolumn{2}{c}{\supplementheaderfont All} & \multicolumn{2}{c}{\supplementheaderfont Visible} & \multicolumn{2}{c}{\supplementheaderfont All} & \multicolumn{2}{c}{\supplementheaderfont Visible} & \multicolumn{2}{c}{\supplementheaderfont All} & \multicolumn{2}{c}{\supplementheaderfont Visible} & \multicolumn{2}{c}{\supplementheaderfont All} & \multicolumn{2}{c}{\supplementheaderfont Visible} & \multicolumn{2}{c}{\supplementheaderfont All} & \multicolumn{2}{c}{\supplementheaderfont Visible} & \multicolumn{2}{c}{\supplementheaderfont All} & \multicolumn{2}{c}{\supplementheaderfont Visible} \\
\cmidrule(lr){2-3}\cmidrule(lr){4-5}\cmidrule(lr){6-7}\cmidrule(lr){8-9}\cmidrule(lr){10-11}\cmidrule(lr){12-13}\cmidrule(lr){14-15}\cmidrule(lr){16-17}\cmidrule(lr){18-19}\cmidrule(lr){20-21}\cmidrule(lr){22-23}\cmidrule(lr){24-25}
 & {\supplementheaderfont AUC$\uparrow$} & {\supplementheaderfont \shortstack{RelEPE\\$\downarrow$}} & {\supplementheaderfont AUC$\uparrow$} & {\supplementheaderfont \shortstack{RelEPE\\$\downarrow$}} & {\supplementheaderfont AUC$\uparrow$} & {\supplementheaderfont \shortstack{RelEPE\\$\downarrow$}} & {\supplementheaderfont AUC$\uparrow$} & {\supplementheaderfont \shortstack{RelEPE\\$\downarrow$}} & {\supplementheaderfont AUC$\uparrow$} & {\supplementheaderfont \shortstack{RelEPE\\$\downarrow$}} & {\supplementheaderfont AUC$\uparrow$} & {\supplementheaderfont \shortstack{RelEPE\\$\downarrow$}} & {\supplementheaderfont AUC$\uparrow$} & {\supplementheaderfont \shortstack{RelEPE\\$\downarrow$}} & {\supplementheaderfont AUC$\uparrow$} & {\supplementheaderfont \shortstack{RelEPE\\$\downarrow$}} & {\supplementheaderfont AUC$\uparrow$} & {\supplementheaderfont \shortstack{RelEPE\\$\downarrow$}} & {\supplementheaderfont AUC$\uparrow$} & {\supplementheaderfont \shortstack{RelEPE\\$\downarrow$}} & {\supplementheaderfont AUC$\uparrow$} & {\supplementheaderfont \shortstack{RelEPE\\$\downarrow$}} & {\supplementheaderfont AUC$\uparrow$} & {\supplementheaderfont \shortstack{RelEPE\\$\downarrow$}} \\
\midrule
St4RTrack & \cellcolor{supplementhighlight!0}44.44 & \cellcolor{supplementhighlight!0}0.195 & \cellcolor{supplementhighlight!0}46.22 & \cellcolor{supplementhighlight!0}0.188 & \cellcolor{supplementhighlight!0}14.24 & \cellcolor{supplementhighlight!0}0.693 & \cellcolor{supplementhighlight!0}16.69 & \cellcolor{supplementhighlight!0}0.624 & \cellcolor{supplementhighlight!0}10.97 & \cellcolor{supplementhighlight!0}0.605 & \cellcolor{supplementhighlight!0}17.05 & \cellcolor{supplementhighlight!0}0.482 & \cellcolor{supplementhighlight!0}9.888 & \cellcolor{supplementhighlight!0}0.867 & \cellcolor{supplementhighlight!0}12.17 & \cellcolor{supplementhighlight!0}0.726 & \cellcolor{supplementhighlight!0}21.08 & \cellcolor{supplementhighlight!0}0.563 & \cellcolor{supplementhighlight!0}22.57 & \cellcolor{supplementhighlight!0}0.568 & \cellcolor{supplementhighlight!0}3.669 & \cellcolor{supplementhighlight!0}1.033 & \cellcolor{supplementhighlight!0}8.478 & \cellcolor{supplementhighlight!0}0.757 \\
Open-D4RT & \cellcolor{supplementhighlight!14}80.96 & \cellcolor{supplementhighlight!14}0.055 & \cellcolor{supplementhighlight!14}81.69 & \cellcolor{supplementhighlight!14}0.053 & \cellcolor{supplementhighlight!7}65.42 & \cellcolor{supplementhighlight!14}0.125 & \cellcolor{supplementhighlight!7}67.79 & \cellcolor{supplementhighlight!14}0.114 & \cellcolor{supplementhighlight!1}47.39 & \cellcolor{supplementhighlight!1}0.200 & \cellcolor{supplementhighlight!1}50.64 & \cellcolor{supplementhighlight!1}0.181 & \cellcolor{supplementhighlight!3}47.97 & \cellcolor{supplementhighlight!14}0.271 & \cellcolor{supplementhighlight!3}51.40 & \cellcolor{supplementhighlight!14}0.227 & \cellcolor{supplementhighlight!14}74.30 & \cellcolor{supplementhighlight!14}0.066 & \cellcolor{supplementhighlight!14}75.47 & \cellcolor{supplementhighlight!24}0.063 & \cellcolor{supplementhighlight!3}24.72 & \cellcolor{supplementhighlight!7}0.354 & \cellcolor{supplementhighlight!3}33.43 & \cellcolor{supplementhighlight!3}0.278 \\
Any4D & \cellcolor{supplementhighlight!7}78.80 & \cellcolor{supplementhighlight!7}0.066 & \cellcolor{supplementhighlight!7}79.73 & \cellcolor{supplementhighlight!7}0.063 & \cellcolor{supplementhighlight!3}57.78 & \cellcolor{supplementhighlight!3}0.159 & \cellcolor{supplementhighlight!3}61.11 & \cellcolor{supplementhighlight!3}0.139 & \cellcolor{supplementhighlight!3}48.66 & \cellcolor{supplementhighlight!3}0.194 & \cellcolor{supplementhighlight!3}53.19 & \cellcolor{supplementhighlight!3}0.160 & \cellcolor{supplementhighlight!1}22.90 & \cellcolor{supplementhighlight!3}0.357 & \cellcolor{supplementhighlight!1}27.49 & \cellcolor{supplementhighlight!7}0.306 & \cellcolor{supplementhighlight!0}50.59 & \cellcolor{supplementhighlight!1}0.132 & \cellcolor{supplementhighlight!0}51.97 & \cellcolor{supplementhighlight!1}0.127 & \cellcolor{supplementhighlight!1}16.88 & \cellcolor{supplementhighlight!1}0.399 & \cellcolor{supplementhighlight!1}26.42 & \cellcolor{supplementhighlight!1}0.284 \\
TraceAnything & \cellcolor{supplementhighlight!0}55.77 & \cellcolor{supplementhighlight!0}0.152 & \cellcolor{supplementhighlight!0}56.58 & \cellcolor{supplementhighlight!0}0.150 & \cellcolor{supplementhighlight!0}17.42 & \cellcolor{supplementhighlight!0}0.391 & \cellcolor{supplementhighlight!0}18.79 & \cellcolor{supplementhighlight!0}0.359 & \cellcolor{supplementhighlight!0}21.04 & \cellcolor{supplementhighlight!0}0.390 & \cellcolor{supplementhighlight!0}24.73 & \cellcolor{supplementhighlight!0}0.326 & \cellcolor{supplementhighlight!0}20.21 & \cellcolor{supplementhighlight!0}0.457 & \cellcolor{supplementhighlight!0}23.86 & \cellcolor{supplementhighlight!0}0.368 & \cellcolor{supplementhighlight!1}53.17 & \cellcolor{supplementhighlight!0}0.140 & \cellcolor{supplementhighlight!1}55.04 & \cellcolor{supplementhighlight!0}0.133 & \cellcolor{supplementhighlight!0}4.373 & \cellcolor{supplementhighlight!0}0.692 & \cellcolor{supplementhighlight!0}8.948 & \cellcolor{supplementhighlight!0}0.493 \\
V-DPM & \cellcolor{supplementhighlight!3}76.18 & \cellcolor{supplementhighlight!3}0.068 & \cellcolor{supplementhighlight!3}76.68 & \cellcolor{supplementhighlight!3}0.067 & \cellcolor{supplementhighlight!80}78.41 & \cellcolor{supplementhighlight!80}0.076 & \cellcolor{supplementhighlight!80}81.11 & \cellcolor{supplementhighlight!80}0.062 & \cellcolor{supplementhighlight!14}68.38 & \cellcolor{supplementhighlight!24}0.116 & \cellcolor{supplementhighlight!14}71.99 & \cellcolor{supplementhighlight!14}0.097 & \cellcolor{supplementhighlight!24}64.44 & \cellcolor{supplementhighlight!7}0.345 & \cellcolor{supplementhighlight!24}67.54 & \cellcolor{supplementhighlight!3}0.306 & \cellcolor{supplementhighlight!24}74.68 & \cellcolor{supplementhighlight!24}0.066 & \cellcolor{supplementhighlight!24}75.67 & \cellcolor{supplementhighlight!14}0.063 & \cellcolor{supplementhighlight!14}34.58 & \cellcolor{supplementhighlight!14}0.290 & \cellcolor{supplementhighlight!7}45.58 & \cellcolor{supplementhighlight!14}0.205 \\
Track4World & \cellcolor{supplementhighlight!1}75.08 & \cellcolor{supplementhighlight!1}0.072 & \cellcolor{supplementhighlight!1}76.14 & \cellcolor{supplementhighlight!1}0.069 & \cellcolor{supplementhighlight!1}54.12 & \cellcolor{supplementhighlight!1}0.184 & \cellcolor{supplementhighlight!1}56.85 & \cellcolor{supplementhighlight!1}0.158 & \cellcolor{supplementhighlight!7}53.94 & \cellcolor{supplementhighlight!7}0.178 & \cellcolor{supplementhighlight!7}60.34 & \cellcolor{supplementhighlight!7}0.142 & \cellcolor{supplementhighlight!7}56.59 & \cellcolor{supplementhighlight!38}0.202 & \cellcolor{supplementhighlight!7}63.89 & \cellcolor{supplementhighlight!38}0.154 & \cellcolor{supplementhighlight!3}55.03 & \cellcolor{supplementhighlight!3}0.122 & \cellcolor{supplementhighlight!3}56.33 & \cellcolor{supplementhighlight!3}0.117 & \cellcolor{supplementhighlight!7}28.72 & \cellcolor{supplementhighlight!3}0.394 & \cellcolor{supplementhighlight!14}46.11 & \cellcolor{supplementhighlight!7}0.235 \\
4RC & \cellcolor{supplementhighlight!56}85.74 & \cellcolor{supplementhighlight!24}0.048 & \cellcolor{supplementhighlight!56}86.71 & \cellcolor{supplementhighlight!24}0.045 & \cellcolor{supplementhighlight!56}73.72 & \cellcolor{supplementhighlight!7}0.134 & \cellcolor{supplementhighlight!38}75.38 & \cellcolor{supplementhighlight!7}0.126 & \cellcolor{supplementhighlight!24}68.44 & \cellcolor{supplementhighlight!14}0.116 & \cellcolor{supplementhighlight!24}72.19 & \cellcolor{supplementhighlight!24}0.097 & \cellcolor{supplementhighlight!14}62.74 & \cellcolor{supplementhighlight!1}0.363 & \cellcolor{supplementhighlight!14}66.14 & \cellcolor{supplementhighlight!1}0.317 & \cellcolor{supplementhighlight!7}63.78 & \cellcolor{supplementhighlight!7}0.097 & \cellcolor{supplementhighlight!7}64.69 & \cellcolor{supplementhighlight!7}0.093 & \cellcolor{supplementhighlight!24}48.89 & \cellcolor{supplementhighlight!24}0.273 & \cellcolor{supplementhighlight!24}59.07 & \cellcolor{supplementhighlight!24}0.190 \\
\midrule
\textbf{Ours} \scriptsize(DINO) & \cellcolor{supplementhighlight!24}84.64 & \cellcolor{supplementhighlight!38}0.042 & \cellcolor{supplementhighlight!24}85.61 & \cellcolor{supplementhighlight!38}0.039 & \cellcolor{supplementhighlight!14}68.68 & \cellcolor{supplementhighlight!24}0.111 & \cellcolor{supplementhighlight!14}70.99 & \cellcolor{supplementhighlight!24}0.096 & \cellcolor{supplementhighlight!38}73.09 & \cellcolor{supplementhighlight!38}0.090 & \cellcolor{supplementhighlight!38}77.40 & \cellcolor{supplementhighlight!38}0.070 & \cellcolor{supplementhighlight!38}69.82 & \cellcolor{supplementhighlight!56}0.170 & \cellcolor{supplementhighlight!56}73.91 & \cellcolor{supplementhighlight!56}0.121 & \cellcolor{supplementhighlight!56}83.05 & \cellcolor{supplementhighlight!56}0.045 & \cellcolor{supplementhighlight!56}83.92 & \cellcolor{supplementhighlight!56}0.043 & \cellcolor{supplementhighlight!56}64.26 & \cellcolor{supplementhighlight!38}0.133 & \cellcolor{supplementhighlight!56}78.32 & \cellcolor{supplementhighlight!56}0.064 \\
\textbf{Ours} \scriptsize(VJEPA) & \cellcolor{supplementhighlight!38}84.90 & \cellcolor{supplementhighlight!56}0.041 & \cellcolor{supplementhighlight!38}85.76 & \cellcolor{supplementhighlight!56}0.039 & \cellcolor{supplementhighlight!24}71.13 & \cellcolor{supplementhighlight!38}0.100 & \cellcolor{supplementhighlight!24}72.67 & \cellcolor{supplementhighlight!38}0.092 & \cellcolor{supplementhighlight!56}77.77 & \cellcolor{supplementhighlight!80}0.073 & \cellcolor{supplementhighlight!56}81.08 & \cellcolor{supplementhighlight!56}0.060 & \cellcolor{supplementhighlight!80}73.35 & \cellcolor{supplementhighlight!80}0.130 & \cellcolor{supplementhighlight!80}76.68 & \cellcolor{supplementhighlight!80}0.092 & \cellcolor{supplementhighlight!80}83.64 & \cellcolor{supplementhighlight!80}0.043 & \cellcolor{supplementhighlight!80}84.23 & \cellcolor{supplementhighlight!80}0.042 & \cellcolor{supplementhighlight!38}64.19 & \cellcolor{supplementhighlight!56}0.120 & \cellcolor{supplementhighlight!38}75.94 & \cellcolor{supplementhighlight!38}0.068 \\
\textbf{Ours} \scriptsize(VGGT-$\Omega$) & \cellcolor{supplementhighlight!80}85.84 & \cellcolor{supplementhighlight!80}0.039 & \cellcolor{supplementhighlight!80}86.81 & \cellcolor{supplementhighlight!80}0.036 & \cellcolor{supplementhighlight!38}73.13 & \cellcolor{supplementhighlight!56}0.098 & \cellcolor{supplementhighlight!56}75.50 & \cellcolor{supplementhighlight!56}0.086 & \cellcolor{supplementhighlight!80}77.98 & \cellcolor{supplementhighlight!56}0.073 & \cellcolor{supplementhighlight!80}81.39 & \cellcolor{supplementhighlight!80}0.059 & \cellcolor{supplementhighlight!56}69.97 & \cellcolor{supplementhighlight!24}0.215 & \cellcolor{supplementhighlight!38}73.16 & \cellcolor{supplementhighlight!24}0.165 & \cellcolor{supplementhighlight!38}81.03 & \cellcolor{supplementhighlight!38}0.051 & \cellcolor{supplementhighlight!38}81.99 & \cellcolor{supplementhighlight!38}0.048 & \cellcolor{supplementhighlight!80}68.89 & \cellcolor{supplementhighlight!80}0.111 & \cellcolor{supplementhighlight!80}80.09 & \cellcolor{supplementhighlight!80}0.059 \\
\bottomrule
\end{tabular}
}
\endgroup
\end{table*}

\begin{table*}[!tbp]
\centering
\caption{Per-benchmark 3D tracking accuracy in world coordinates ($\mathrm{AUC}_{25}\uparrow$, $\mathrm{RelEPE}\downarrow$).}
\label{tab:supp_tracking_world}
\begingroup
\fontsize{7pt}{8.4pt}\selectfont
\newcommand{\supplementheaderfont}{\fontsize{5pt}{6pt}\selectfont}
\renewcommand{\scriptsize}{\supplementheaderfont}
\definecolor{supplementhighlight}{RGB}{255,215,0}
\setlength{\tabcolsep}{0.75pt}
\setlength{\aboverulesep}{0pt}
\setlength{\belowrulesep}{0pt}
\renewcommand{\arraystretch}{1.15}
\resizebox{\textwidth}{!}{%
\begin{tabular}{@{}lrrrr@{\hspace{4pt}}rrrr@{\hspace{4pt}}rrrr@{\hspace{4pt}}rrrr@{\hspace{4pt}}rrrr@{\hspace{4pt}}rrrr@{}}
\toprule
\multirow{3}{*}{\supplementheaderfont Method} & \multicolumn{4}{c}{\supplementheaderfont DynamicReplica} & \multicolumn{4}{c}{\supplementheaderfont PointOdyssey} & \multicolumn{4}{c}{\supplementheaderfont Syn4D} & \multicolumn{4}{c}{\supplementheaderfont ADT} & \multicolumn{4}{c}{\supplementheaderfont PStudio} & \multicolumn{4}{c}{\supplementheaderfont RigIDTrack} \\
\cmidrule(lr){2-5}\cmidrule(lr){6-9}\cmidrule(lr){10-13}\cmidrule(lr){14-17}\cmidrule(lr){18-21}\cmidrule(lr){22-25}
 & \multicolumn{2}{c}{\supplementheaderfont All} & \multicolumn{2}{c}{\supplementheaderfont Visible} & \multicolumn{2}{c}{\supplementheaderfont All} & \multicolumn{2}{c}{\supplementheaderfont Visible} & \multicolumn{2}{c}{\supplementheaderfont All} & \multicolumn{2}{c}{\supplementheaderfont Visible} & \multicolumn{2}{c}{\supplementheaderfont All} & \multicolumn{2}{c}{\supplementheaderfont Visible} & \multicolumn{2}{c}{\supplementheaderfont All} & \multicolumn{2}{c}{\supplementheaderfont Visible} & \multicolumn{2}{c}{\supplementheaderfont All} & \multicolumn{2}{c}{\supplementheaderfont Visible} \\
\cmidrule(lr){2-3}\cmidrule(lr){4-5}\cmidrule(lr){6-7}\cmidrule(lr){8-9}\cmidrule(lr){10-11}\cmidrule(lr){12-13}\cmidrule(lr){14-15}\cmidrule(lr){16-17}\cmidrule(lr){18-19}\cmidrule(lr){20-21}\cmidrule(lr){22-23}\cmidrule(lr){24-25}
 & {\supplementheaderfont AUC$\uparrow$} & {\supplementheaderfont \shortstack{RelEPE\\$\downarrow$}} & {\supplementheaderfont AUC$\uparrow$} & {\supplementheaderfont \shortstack{RelEPE\\$\downarrow$}} & {\supplementheaderfont AUC$\uparrow$} & {\supplementheaderfont \shortstack{RelEPE\\$\downarrow$}} & {\supplementheaderfont AUC$\uparrow$} & {\supplementheaderfont \shortstack{RelEPE\\$\downarrow$}} & {\supplementheaderfont AUC$\uparrow$} & {\supplementheaderfont \shortstack{RelEPE\\$\downarrow$}} & {\supplementheaderfont AUC$\uparrow$} & {\supplementheaderfont \shortstack{RelEPE\\$\downarrow$}} & {\supplementheaderfont AUC$\uparrow$} & {\supplementheaderfont \shortstack{RelEPE\\$\downarrow$}} & {\supplementheaderfont AUC$\uparrow$} & {\supplementheaderfont \shortstack{RelEPE\\$\downarrow$}} & {\supplementheaderfont AUC$\uparrow$} & {\supplementheaderfont \shortstack{RelEPE\\$\downarrow$}} & {\supplementheaderfont AUC$\uparrow$} & {\supplementheaderfont \shortstack{RelEPE\\$\downarrow$}} & {\supplementheaderfont AUC$\uparrow$} & {\supplementheaderfont \shortstack{RelEPE\\$\downarrow$}} & {\supplementheaderfont AUC$\uparrow$} & {\supplementheaderfont \shortstack{RelEPE\\$\downarrow$}} \\
\midrule
St4RTrack & \cellcolor{supplementhighlight!0}57.70 & \cellcolor{supplementhighlight!0}0.158 & \cellcolor{supplementhighlight!0}59.82 & \cellcolor{supplementhighlight!0}0.150 & \cellcolor{supplementhighlight!0}23.85 & \cellcolor{supplementhighlight!0}0.380 & \cellcolor{supplementhighlight!0}27.94 & \cellcolor{supplementhighlight!0}0.344 & \cellcolor{supplementhighlight!0}22.47 & \cellcolor{supplementhighlight!0}0.425 & \cellcolor{supplementhighlight!0}30.89 & \cellcolor{supplementhighlight!0}0.354 & \cellcolor{supplementhighlight!0}22.10 & \cellcolor{supplementhighlight!0}0.602 & \cellcolor{supplementhighlight!0}26.83 & \cellcolor{supplementhighlight!0}0.535 & \cellcolor{supplementhighlight!0}44.01 & \cellcolor{supplementhighlight!0}0.185 & \cellcolor{supplementhighlight!0}46.63 & \cellcolor{supplementhighlight!0}0.176 & \cellcolor{supplementhighlight!0}6.961 & \cellcolor{supplementhighlight!0}0.613 & \cellcolor{supplementhighlight!0}16.35 & \cellcolor{supplementhighlight!0}0.498 \\
Open-D4RT & \cellcolor{supplementhighlight!7}81.84 & \cellcolor{supplementhighlight!7}0.053 & \cellcolor{supplementhighlight!7}82.62 & \cellcolor{supplementhighlight!7}0.051 & \cellcolor{supplementhighlight!7}70.33 & \cellcolor{supplementhighlight!14}0.112 & \cellcolor{supplementhighlight!7}72.07 & \cellcolor{supplementhighlight!14}0.102 & \cellcolor{supplementhighlight!1}59.65 & \cellcolor{supplementhighlight!1}0.161 & \cellcolor{supplementhighlight!1}62.53 & \cellcolor{supplementhighlight!1}0.147 & \cellcolor{supplementhighlight!3}54.24 & \cellcolor{supplementhighlight!14}0.293 & \cellcolor{supplementhighlight!3}59.43 & \cellcolor{supplementhighlight!14}0.277 & \cellcolor{supplementhighlight!14}83.20 & \cellcolor{supplementhighlight!14}0.044 & \cellcolor{supplementhighlight!14}84.48 & \cellcolor{supplementhighlight!24}0.040 & \cellcolor{supplementhighlight!7}44.41 & \cellcolor{supplementhighlight!3}0.287 & \cellcolor{supplementhighlight!7}53.05 & \cellcolor{supplementhighlight!1}0.232 \\
Any4D & \cellcolor{supplementhighlight!3}81.54 & \cellcolor{supplementhighlight!3}0.059 & \cellcolor{supplementhighlight!3}82.59 & \cellcolor{supplementhighlight!3}0.056 & \cellcolor{supplementhighlight!3}66.59 & \cellcolor{supplementhighlight!3}0.140 & \cellcolor{supplementhighlight!3}69.41 & \cellcolor{supplementhighlight!7}0.122 & \cellcolor{supplementhighlight!7}61.00 & \cellcolor{supplementhighlight!7}0.151 & \cellcolor{supplementhighlight!3}66.05 & \cellcolor{supplementhighlight!3}0.124 & \cellcolor{supplementhighlight!0}49.69 & \cellcolor{supplementhighlight!3}0.348 & \cellcolor{supplementhighlight!0}56.32 & \cellcolor{supplementhighlight!7}0.312 & \cellcolor{supplementhighlight!1}70.89 & \cellcolor{supplementhighlight!1}0.086 & \cellcolor{supplementhighlight!1}72.97 & \cellcolor{supplementhighlight!1}0.078 & \cellcolor{supplementhighlight!3}42.02 & \cellcolor{supplementhighlight!7}0.282 & \cellcolor{supplementhighlight!3}51.87 & \cellcolor{supplementhighlight!7}0.216 \\
TraceAnything & \cellcolor{supplementhighlight!0}68.95 & \cellcolor{supplementhighlight!0}0.117 & \cellcolor{supplementhighlight!0}70.09 & \cellcolor{supplementhighlight!0}0.115 & \cellcolor{supplementhighlight!0}45.76 & \cellcolor{supplementhighlight!0}0.286 & \cellcolor{supplementhighlight!0}46.96 & \cellcolor{supplementhighlight!0}0.272 & \cellcolor{supplementhighlight!0}54.09 & \cellcolor{supplementhighlight!0}0.219 & \cellcolor{supplementhighlight!0}57.59 & \cellcolor{supplementhighlight!0}0.194 & \cellcolor{supplementhighlight!1}52.18 & \cellcolor{supplementhighlight!0}0.363 & \cellcolor{supplementhighlight!1}58.14 & \cellcolor{supplementhighlight!1}0.336 & \cellcolor{supplementhighlight!0}65.44 & \cellcolor{supplementhighlight!0}0.108 & \cellcolor{supplementhighlight!0}66.85 & \cellcolor{supplementhighlight!0}0.103 & \cellcolor{supplementhighlight!1}30.25 & \cellcolor{supplementhighlight!0}0.372 & \cellcolor{supplementhighlight!0}37.85 & \cellcolor{supplementhighlight!0}0.324 \\
V-DPM & \cellcolor{supplementhighlight!14}82.42 & \cellcolor{supplementhighlight!14}0.052 & \cellcolor{supplementhighlight!14}83.09 & \cellcolor{supplementhighlight!14}0.051 & \cellcolor{supplementhighlight!80}79.79 & \cellcolor{supplementhighlight!80}0.073 & \cellcolor{supplementhighlight!80}82.44 & \cellcolor{supplementhighlight!80}0.059 & \cellcolor{supplementhighlight!24}74.01 & \cellcolor{supplementhighlight!14}0.099 & \cellcolor{supplementhighlight!24}77.96 & \cellcolor{supplementhighlight!24}0.080 & \cellcolor{supplementhighlight!24}63.34 & \cellcolor{supplementhighlight!7}0.340 & \cellcolor{supplementhighlight!24}69.83 & \cellcolor{supplementhighlight!0}0.338 & \cellcolor{supplementhighlight!24}83.52 & \cellcolor{supplementhighlight!24}0.043 & \cellcolor{supplementhighlight!24}84.59 & \cellcolor{supplementhighlight!14}0.041 & \cellcolor{supplementhighlight!14}52.80 & \cellcolor{supplementhighlight!24}0.220 & \cellcolor{supplementhighlight!14}63.64 & \cellcolor{supplementhighlight!24}0.156 \\
Track4World & \cellcolor{supplementhighlight!1}76.94 & \cellcolor{supplementhighlight!1}0.068 & \cellcolor{supplementhighlight!1}78.14 & \cellcolor{supplementhighlight!1}0.064 & \cellcolor{supplementhighlight!1}59.72 & \cellcolor{supplementhighlight!1}0.176 & \cellcolor{supplementhighlight!1}62.44 & \cellcolor{supplementhighlight!1}0.147 & \cellcolor{supplementhighlight!3}60.35 & \cellcolor{supplementhighlight!3}0.152 & \cellcolor{supplementhighlight!7}67.91 & \cellcolor{supplementhighlight!7}0.118 & \cellcolor{supplementhighlight!7}57.66 & \cellcolor{supplementhighlight!56}0.180 & \cellcolor{supplementhighlight!7}67.82 & \cellcolor{supplementhighlight!56}0.135 & \cellcolor{supplementhighlight!3}73.74 & \cellcolor{supplementhighlight!3}0.072 & \cellcolor{supplementhighlight!3}75.24 & \cellcolor{supplementhighlight!3}0.066 & \cellcolor{supplementhighlight!0}28.47 & \cellcolor{supplementhighlight!1}0.366 & \cellcolor{supplementhighlight!1}50.19 & \cellcolor{supplementhighlight!3}0.221 \\
4RC & \cellcolor{supplementhighlight!56}86.12 & \cellcolor{supplementhighlight!24}0.047 & \cellcolor{supplementhighlight!56}87.11 & \cellcolor{supplementhighlight!24}0.044 & \cellcolor{supplementhighlight!56}73.99 & \cellcolor{supplementhighlight!7}0.135 & \cellcolor{supplementhighlight!38}76.16 & \cellcolor{supplementhighlight!3}0.126 & \cellcolor{supplementhighlight!14}73.98 & \cellcolor{supplementhighlight!24}0.099 & \cellcolor{supplementhighlight!14}77.62 & \cellcolor{supplementhighlight!14}0.081 & \cellcolor{supplementhighlight!14}61.37 & \cellcolor{supplementhighlight!1}0.349 & \cellcolor{supplementhighlight!14}68.24 & \cellcolor{supplementhighlight!3}0.324 & \cellcolor{supplementhighlight!7}82.26 & \cellcolor{supplementhighlight!7}0.050 & \cellcolor{supplementhighlight!7}83.56 & \cellcolor{supplementhighlight!7}0.045 & \cellcolor{supplementhighlight!24}53.47 & \cellcolor{supplementhighlight!14}0.232 & \cellcolor{supplementhighlight!24}65.87 & \cellcolor{supplementhighlight!14}0.167 \\
\midrule
\textbf{Ours} \scriptsize(DINO) & \cellcolor{supplementhighlight!24}85.17 & \cellcolor{supplementhighlight!38}0.041 & \cellcolor{supplementhighlight!24}86.20 & \cellcolor{supplementhighlight!38}0.038 & \cellcolor{supplementhighlight!14}72.37 & \cellcolor{supplementhighlight!24}0.099 & \cellcolor{supplementhighlight!14}75.82 & \cellcolor{supplementhighlight!56}0.082 & \cellcolor{supplementhighlight!38}78.44 & \cellcolor{supplementhighlight!38}0.074 & \cellcolor{supplementhighlight!38}82.48 & \cellcolor{supplementhighlight!38}0.056 & \cellcolor{supplementhighlight!56}69.54 & \cellcolor{supplementhighlight!38}0.229 & \cellcolor{supplementhighlight!56}75.68 & \cellcolor{supplementhighlight!38}0.195 & \cellcolor{supplementhighlight!56}88.03 & \cellcolor{supplementhighlight!56}0.033 & \cellcolor{supplementhighlight!56}89.15 & \cellcolor{supplementhighlight!56}0.029 & \cellcolor{supplementhighlight!38}69.79 & \cellcolor{supplementhighlight!38}0.120 & \cellcolor{supplementhighlight!56}82.12 & \cellcolor{supplementhighlight!38}0.061 \\
\textbf{Ours} \scriptsize(VJEPA) & \cellcolor{supplementhighlight!38}85.44 & \cellcolor{supplementhighlight!56}0.040 & \cellcolor{supplementhighlight!38}86.34 & \cellcolor{supplementhighlight!56}0.037 & \cellcolor{supplementhighlight!38}73.83 & \cellcolor{supplementhighlight!56}0.093 & \cellcolor{supplementhighlight!24}75.98 & \cellcolor{supplementhighlight!24}0.083 & \cellcolor{supplementhighlight!56}80.91 & \cellcolor{supplementhighlight!80}0.064 & \cellcolor{supplementhighlight!56}83.92 & \cellcolor{supplementhighlight!56}0.052 & \cellcolor{supplementhighlight!80}73.02 & \cellcolor{supplementhighlight!80}0.138 & \cellcolor{supplementhighlight!80}79.26 & \cellcolor{supplementhighlight!80}0.097 & \cellcolor{supplementhighlight!80}89.00 & \cellcolor{supplementhighlight!80}0.030 & \cellcolor{supplementhighlight!80}89.86 & \cellcolor{supplementhighlight!80}0.027 & \cellcolor{supplementhighlight!56}71.11 & \cellcolor{supplementhighlight!56}0.100 & \cellcolor{supplementhighlight!38}81.65 & \cellcolor{supplementhighlight!56}0.061 \\
\textbf{Ours} \scriptsize(VGGT-$\Omega$) & \cellcolor{supplementhighlight!80}86.23 & \cellcolor{supplementhighlight!80}0.038 & \cellcolor{supplementhighlight!80}87.21 & \cellcolor{supplementhighlight!80}0.035 & \cellcolor{supplementhighlight!24}73.63 & \cellcolor{supplementhighlight!38}0.096 & \cellcolor{supplementhighlight!56}76.42 & \cellcolor{supplementhighlight!38}0.083 & \cellcolor{supplementhighlight!80}81.21 & \cellcolor{supplementhighlight!56}0.065 & \cellcolor{supplementhighlight!80}84.53 & \cellcolor{supplementhighlight!80}0.051 & \cellcolor{supplementhighlight!38}67.33 & \cellcolor{supplementhighlight!24}0.231 & \cellcolor{supplementhighlight!38}74.12 & \cellcolor{supplementhighlight!24}0.196 & \cellcolor{supplementhighlight!38}86.47 & \cellcolor{supplementhighlight!38}0.038 & \cellcolor{supplementhighlight!38}87.45 & \cellcolor{supplementhighlight!38}0.034 & \cellcolor{supplementhighlight!80}72.92 & \cellcolor{supplementhighlight!80}0.099 & \cellcolor{supplementhighlight!80}84.01 & \cellcolor{supplementhighlight!80}0.048 \\
\bottomrule
\end{tabular}
}
\endgroup
\end{table*}

\begin{table*}[!tbp]
\centering
\caption{Per-benchmark occlusion accuracy ($\mathrm{OA}\uparrow$) for 3D tracking.}
\label{tab:supp_occlusion}
\begingroup
\fontsize{7pt}{8.4pt}\selectfont
\newcommand{\supplementheaderfont}{\fontsize{6pt}{7.2pt}\selectfont}
\renewcommand{\scriptsize}{\supplementheaderfont}
\definecolor{supplementhighlight}{RGB}{255,215,0}
\setlength{\tabcolsep}{1.5pt}
\setlength{\aboverulesep}{0pt}
\setlength{\belowrulesep}{0pt}
\renewcommand{\arraystretch}{1.15}
\begin{tabular}{@{}lr@{\hspace{4pt}}r@{\hspace{4pt}}r@{\hspace{4pt}}r@{\hspace{4pt}}r@{\hspace{4pt}}r@{}}
\toprule
\multirow{2}{*}{\supplementheaderfont Method} & {\supplementheaderfont DynamicReplica} & {\supplementheaderfont PointOdyssey} & {\supplementheaderfont Syn4D} & {\supplementheaderfont ADT} & {\supplementheaderfont PStudio} & {\supplementheaderfont RigIDTrack} \\
 & {\supplementheaderfont OA$\uparrow$} & {\supplementheaderfont OA$\uparrow$} & {\supplementheaderfont OA$\uparrow$} & {\supplementheaderfont OA$\uparrow$} & {\supplementheaderfont OA$\uparrow$} & {\supplementheaderfont OA$\uparrow$} \\
\midrule
Open-D4RT & \cellcolor{supplementhighlight!0}88.23 & \cellcolor{supplementhighlight!80}79.81 & \cellcolor{supplementhighlight!0}78.74 & \cellcolor{supplementhighlight!0}86.13 & \cellcolor{supplementhighlight!1}84.36 & \cellcolor{supplementhighlight!0}87.68 \\
Track4World & \cellcolor{supplementhighlight!1}89.20 & \cellcolor{supplementhighlight!0}75.76 & \cellcolor{supplementhighlight!1}83.27 & \cellcolor{supplementhighlight!34}93.91 & \cellcolor{supplementhighlight!0}78.31 & \cellcolor{supplementhighlight!1}89.33 \\
\midrule
\textbf{Ours} \scriptsize(DINO) & \cellcolor{supplementhighlight!34}90.16 & \cellcolor{supplementhighlight!34}78.10 & \cellcolor{supplementhighlight!80}88.79 & \cellcolor{supplementhighlight!10}92.42 & \cellcolor{supplementhighlight!80}88.63 & \cellcolor{supplementhighlight!80}94.80 \\
\textbf{Ours} \scriptsize(VJEPA) & \cellcolor{supplementhighlight!80}90.16 & \cellcolor{supplementhighlight!1}77.75 & \cellcolor{supplementhighlight!34}88.36 & \cellcolor{supplementhighlight!80}94.28 & \cellcolor{supplementhighlight!34}87.65 & \cellcolor{supplementhighlight!10}94.28 \\
\textbf{Ours} \scriptsize(VGGT-$\Omega$) & \cellcolor{supplementhighlight!10}90.10 & \cellcolor{supplementhighlight!10}77.77 & \cellcolor{supplementhighlight!10}88.34 & \cellcolor{supplementhighlight!1}90.58 & \cellcolor{supplementhighlight!10}85.58 & \cellcolor{supplementhighlight!34}94.45 \\
\bottomrule
\end{tabular}
\endgroup
\end{table*}

\subsection{Long-Horizon Trajectory Evaluation on Oxford-Spires}
\label{sec:supp_long_horizon}

\begin{figure*}[p]
	\centering
	\includegraphics[width=\textwidth,height=0.90\textheight,keepaspectratio]{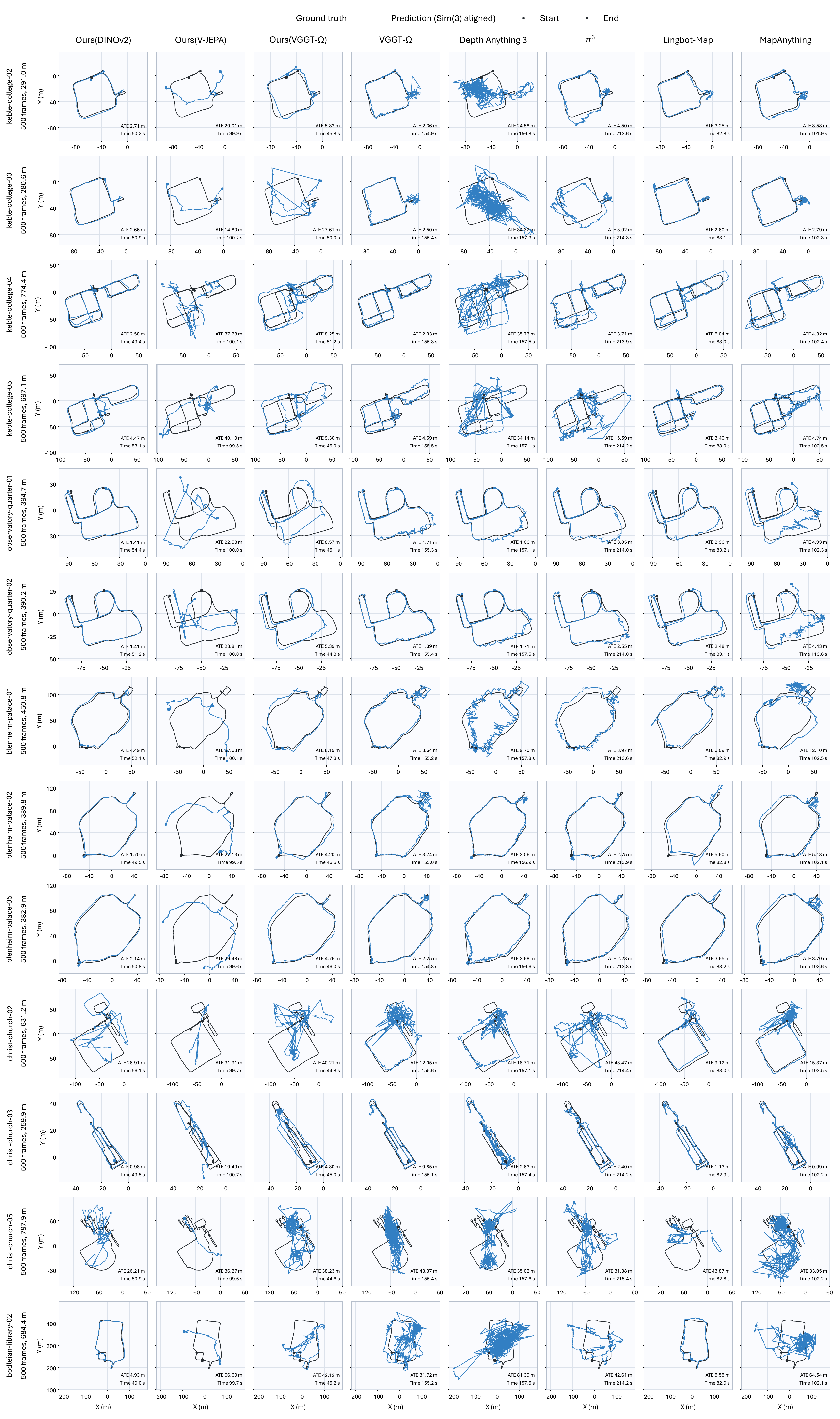}
	\caption{\textbf{Complete 500-frame camera trajectory comparisons on Oxford-Spires.} All methods run in a single feed-forward pass without sliding windows.}
	\label{fig:supp_trajectories_all}
\end{figure*}

Figure~\ref{fig:supp_trajectories_all} reports the complete long-horizon camera trajectory results across all 13 evaluated 500-frame sequences from the Oxford-Spires dataset~\citep{tao2025spires}, comparing all three MoGe-4D variants (DINOv2, V-JEPA 2.1, and VGGT-$\Omega$) alongside VGGT-$\Omega$, Depth Anything 3, $\pi^3$, LingBot-Map, and MapAnything.
Each method processes the full 500-frame sequence in a single feed-forward pass without sliding windows.
Consistent with the analysis in Section~\ref{sec:limitations}, our permutation-invariant DINOv2 variant generalizes most reliably to 500-frame sequences while running substantially faster than competing feed-forward reconstruction baselines.

\subsection{Additional Qualitative Comparisons}
\label{sec:supp_qualitative_comparisons}

\begin{figure*}[p]
    \centering
    \includegraphics[width=\textwidth,height=0.86\textheight,keepaspectratio]{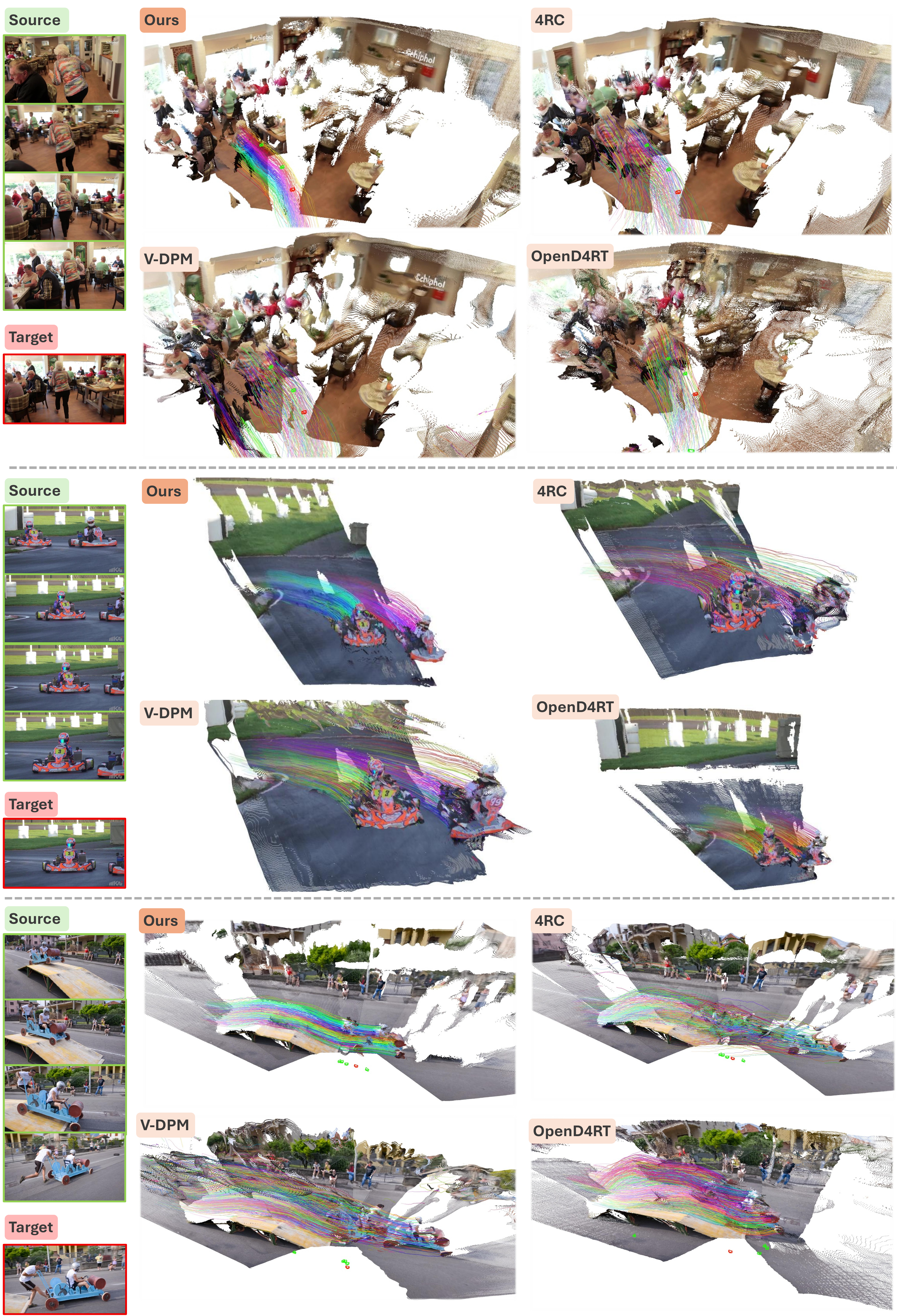}
    \caption{\textbf{Additional qualitative comparisons (part I)}, following the visualization setup of Figure~\ref{fig:qualitative_results}.}
    \label{fig:supp_qualitative_comparisons_1}
\end{figure*}

\begin{figure*}[p]
    \centering
    \includegraphics[width=\textwidth,height=0.86\textheight,keepaspectratio]{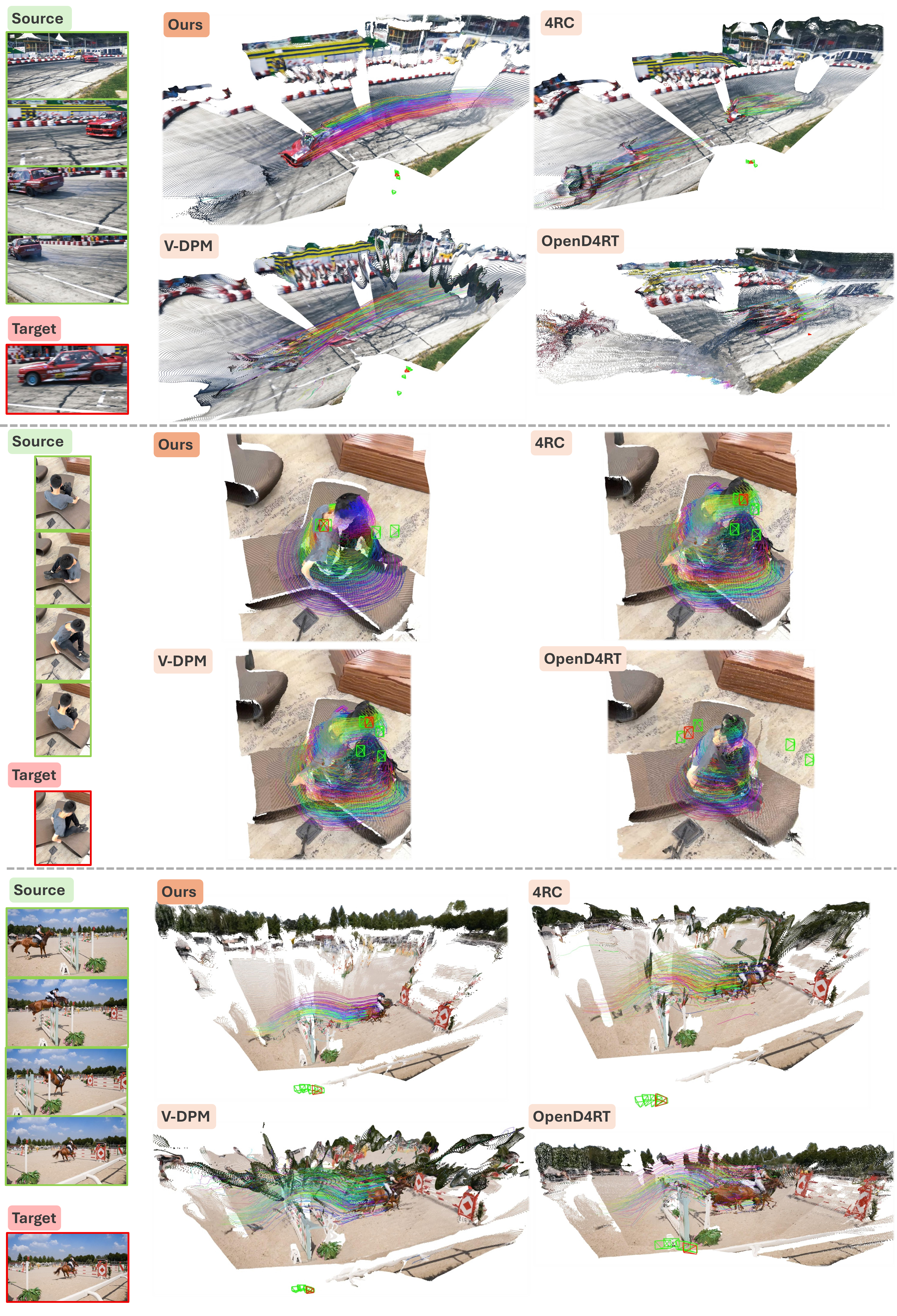}
    \caption{\textbf{Additional qualitative comparisons (part II)}, following the visualization setup of Figure~\ref{fig:qualitative_results}.}
    \label{fig:supp_qualitative_comparisons_2}
\end{figure*}

Figures~\ref{fig:supp_qualitative_comparisons_1} and~\ref{fig:supp_qualitative_comparisons_2} show additional qualitative comparisons with 4RC, V-DPM, and Open-D4RT.
Each example displays four source frames (green), a held-out target frame (red), and the combined 4D reconstruction produced by each method.

\end{document}